\documentclass{article}
\usepackage{ijcai25}

\usepackage{times}
\usepackage{soul}
\usepackage{url}
\usepackage[hidelinks]{hyperref}
\usepackage[utf8]{inputenc}
\usepackage[small,skip=2pt]{caption}
\usepackage{graphicx}
\usepackage{amsmath}
\usepackage{amsfonts}
\usepackage{amsthm}

\usepackage[table,xcdraw]{xcolor}
\usepackage{booktabs}
\usepackage{algorithm}
\usepackage{algpseudocode}
\usepackage[switch]{lineno}
\usepackage{tikz}
\usepackage{adjustbox}
\usepackage{lipsum}
\usepackage{float}
\usepackage{subcaption}
\usepackage{multirow}
\usepackage{wrapfig}
\usepackage{nicematrix}
\usepackage[normalem]{ulem}
\useunder{\uline}{\ul}{}

\usepackage{amssymb}
\usepackage{pifont}
\newcommand{\cmark}{\ding{51}}%
\newcommand{\xmark}{\ding{55}}%

\newtheorem{theorem}{Theorem}

\newtheorem{proposition}{Proposition}

\title{\emph{GatedFWA}: Linear Flash Windowed Attention with Gated Associative Memory}

\author{
Jiaxu Liu$^1$
\and
Yuhe Bai$^{2}$
\and
Xiangyu Yin$^{3}$
\And
Christos-Savvas Bouganis$^1$\\
\affiliations
$^1$Imperial College London,
$^2$Sorbonne University,
$^3$Chalmers University of Technology
\emails
\{j.liu2, christos-savvas.bouganis\}@imperial.ac.uk,
yuhe.bai@sorbonne-universite.fr,
yinxi@chalmers.se
}

\begin{document}

\maketitle

\begin{abstract}
Modern autoregressive models rely on attention, yet the Softmax full attention in Transformers scales quadratically with sequence length. Sliding Window Attention (SWA) achieves linear-time encoding/decoding by constraining the attention pattern, but under an \textit{Associative Memory} interpretation, its difference-style update renders the training objective effectively \emph{unbounded}. In contrast, Softmax attention normalizes updates, leading to \emph{memory shrinkage and gradient vanishing}. We propose GatedFWA: a Memory-\underline{Gated} (\underline{F}lash-)\underline{W}indowed \underline{A}ttention mechanism that preserves SWA’s efficiency while stabilizing memory updates and making gradient flow controllable. In essence, GatedFWA accumulate a per-token/head gate into a decay bias added to the attention logits, acting as a learnable contraction in the memory recurrence. We implement a fused one-pass gate preprocessing and a FlashAttention-compatible kernel that injects the gate under a sliding mask, ensuring I/O efficiency and numerical stability. On language modelling benchmarks, GatedFWA delivers competitive throughput with negligible overhead and better use of global context, and it integrates cleanly with token compression/selection methods such as NSA and generalizes to various autoregressive domains.

\end{abstract}

\section{Introduction}
\label{sec:intro}

Autoregressive modelling, \emph{i.e.}, predicting the next element from past context, powers modern generative systems in language~\cite{minaee2024large}, speech~\cite{chu2024qwen2}, code~\cite{roziere2023code}, multivariate time series \cite{bordes2024introduction}, and event streams~\cite{zhong2025survey}. Transformer~\cite{vaswani2017attention} architectures remain the dominant backbone, but their \emph{full Softmax} attention scales quadratically in sequence length, constraining latency, throughput, and training on long horizons. Sliding Window Attention (SWA) offers a pragmatic compromise: by restricting each query to a local window of width $w$, it preserves parallelism during training and delivers \emph{linear-time decoding} with $\mathcal{O}(Nwd)$ ($\mathcal{O}(wd)$ with KV-cache) arithmetic and $\mathcal{O}(Nd+Nw)$ memory traffic for hidden dimension $d$. 

\begin{figure}[t]
    \centering
    \includegraphics[width=\linewidth]{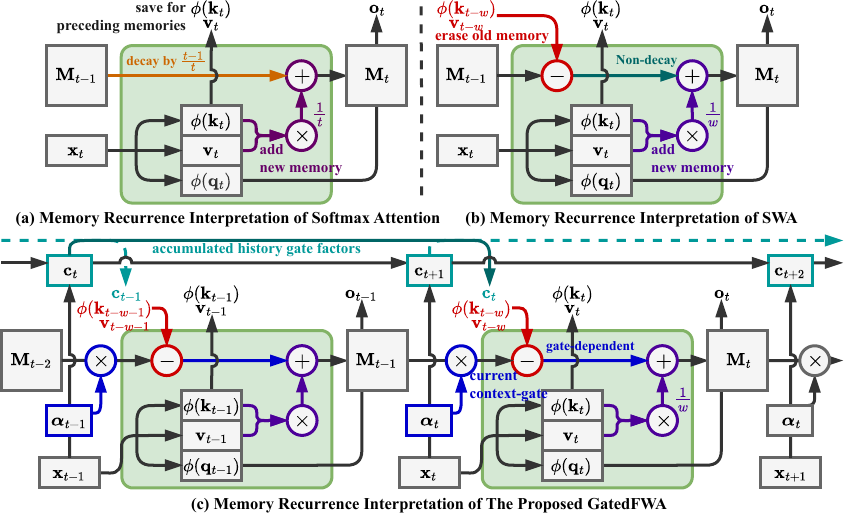}
    \caption{
    Memory recurrence interpretation of \textbf{(a)} Softmax (Eq.~\eqref{eq:recurrent_form_softmax}): the carried memory is scaled by $\frac{t-1}{t}$ and a $\frac{1}{t}$ new term is added, so normalization steadily shrinks per-step updates and drives gradient vanishing through $\mathbf{M}_t$. \textbf{(b)} SWA (Eq.~\eqref{eq:recurrent_form_swa}): within a width $w$ window the state is non-decaying but updated by a difference term $\phi(\mathbf{k}_t)^\top\mathbf{v}_t-\phi(\mathbf{k}_{t-w})^\top\mathbf{v}_{t-w}$; this implicitly optimizes an unbounded linear objective, can over-amplify memory (unstable gradients). \textbf{(c)} GatedFWA (Eq.~\eqref{eq:mem_recurrence_gatedfwa}): with non-negative gate accumulates into a decay bias ($\mathbf{B}_{ti}=\sum_{q=i+1}^t -\boldsymbol{\alpha}_q$), yielding a learnable contraction ($\mathbf{M}_t=\exp(-\boldsymbol{\alpha}_t)\mathbf{M}_{t-1}+\cdots$) that softly erases off-path history, bounds the update, and makes gradient flow controllable while retaining SWA’s linear cost. We draw \textbf{(c)} in two steps because its update depends on multiple prior states for stability. 
    } 
    \label{fig:teaser}
\end{figure}

We treat causal attention as an explicit Associative Memory~\cite{hopfield1982neural,schlag2021linear} that is updated recurrently like an RNN (as depicted in Fig.~\ref{fig:teaser}). This viewpoint enables the investigation of the update expressivity and gradient stability of Transformers. For instance, in \textbf{Softmax attention}, the growing normalization through length $t$ causes the effective per-step update to shrink by $\frac{1}{t}$, which drives \emph{gradient vanishing} through the memory state and weakens credit assignment. For \textbf{SWA}, the update behaves like a local difference between entering and leaving tokens; the induced training objective is effectively \emph{unbounded} with respect to the memory magnitude, which encourages overly large memory updates and yields \emph{gradient instability}. Notably, an orthogonal issue on reachability due to limited window size $w$ can also be interpreted from the SWA memory recurrence since its \emph{dynamically removing} contributions from tokens that fall outside the sliding window. In short, Softmax attention tends to suppress the memory update, while SWA tends to amplify it without a stabilizing counter-term.  This diagnosis motivates our goal: \emph{retaining the linear-time footprint of SWA while introducing a mechanism that keeps the memory update \textbf{stable} and the gradient path \textbf{controllable}}, so that credit can be preserved where needed and suppressed where not.

In this paper, we introduce \textbf{GatedFWA}: a Memory-\underline{Gated} (\underline{F}lash-)\underline{W}indowed \underline{A}ttention that preserves the linear runtime of SWA while addressing both gradient issues. For each token and head, a lightweight non-negative gate is computed and optimized through training, accumulated into a cumulative decay term along the sequence, and added to the attention logits. Interpreted through associative memory, this gate implements a learnable contraction on the carried memory (as Eq.~\eqref{eq:mem_recurrence_gatedfwa}), preventing the unbounded growth encouraged by SWA’s memory update, and it assigns priority to on-path associations so that gradient flow can be maintained across many steps instead of being suppressed by normalization.

For \emph{hardware alignment}, we implement GatedFWA to fit modern accelerator kernels. A one-pass, chunk-wise fused preprocessing computes numerically safe gates and their prefix sums. A minimal FlashAttention-style extension injects the decay bias under a sliding mask. This keeps I/O awareness, SRAM tiling, and numerical stability intact, while adding only lightweight vector loads and bias arithmetic. In practice, GatedFWA retains the $\mathcal{O}(Nwd)$ profile of SWA, and $\mathcal{O}(wd)$ per decoding step with a KV cache, while improving the reliability of gradient assignment in deep autoregressive stacks. Nevertheless, as token compression and selection methods such as NSA can expand effective context beyond a fixed window by retaining salient tokens, GatedFWA can serves as a drop-in replacement of the local sliding module inside such pipelines, addressing reachability independently of our gradient-stability objective.

To summarize, our contributions are: 
\emph{\textbf{(i) Interpretation through Associative Memory.}}
We recast causal attention as an optimized memory and show two gradient pathologies: Softmax causes gradient vanishing via normalization, and SWA causes gradient instability from a difference-style update.
\emph{\textbf{(ii) GatedFWA.}}
A memory-gated variant of sliding attention that adds a learnable contraction and path-selective biasing to the logits, stabilizing the memory update via a controllable memory gate while preserving linear-time complexity.
\emph{\textbf{(iii) Hardware Alignment.}} 
We propose a fused one-pass preprocessing to compute gates and prefix sums, and a FlashAttention-compatible kernel that injects the bias ladder under a sliding mask. This maintains I/O awareness, on-chip tiling, and numerical stability with negligible overhead.
\emph{\textbf{(iv) Empirical Validation.}}
On language modelling benchmarks, GatedFWA (and its NSA compatible variant) attains state-of-the-art efficiency and quality among linear-time attention and state-space baselines, sustaining long-range sensitivity with near-zero preprocessing cost and competitive throughput in both forward and backward passes.

\section{Preliminary}

\subsection{Linear Attention and Sparse Patterns}
\label{sec:pattern-based-transformer}
Sparse-pattern attention is one of the most widely adopted strategies for building efficient Transformers. It mitigates the quadratic cost of full attention by enforcing structured sparsity. Early works such as Sparse Transformers~\cite{child2019generating} used blockwise or strided pattern to achieve $\mathcal{O}(N\sqrt{N})$ complexity. Longformer and ETC~\cite{beltagy2020longformer,ainslie2020etc} improved this by combining local sliding windows with global memory tokens, reducing complexity to $\mathcal{O}(Nw)$, where $w$ is the window size. Other variants include axial attention~\cite{ho2019axial}, hashing-based patterns~\cite{kitaev2020reformer}, and clustering-based sparsity~\cite{roy2021efficient}.  Notably, recent large-scale systems such as NSA~\cite{yuan2025native} and GPT-OSS~\cite{agarwal2025gpt} integrate \textbf{sliding window attention} as their core components, highlighting its status as an industry standard for efficient LLM deployment. 

\subsection{Associative Memory}
Associative memory refers to the ability to learn and recall relationships between entities, even when these entities are not directly related. For example, after visiting \emph{Paris} and seeing the \emph{Eiffel Tower}, one naturally associates the two concepts, such that hearing \emph{Paris} later triggers the recall of the \emph{Eiffel Tower}. Formally, this cognitive mechanism can be modelled by a time-varying memory matrix \( \mathbf{M}_t \in \mathbb{R}^{d_k \times d_v} \), which stores associations of various key-value pairs \((\mathbf{k}_i, \mathbf{v}_i)\), where \( \mathbf{k}_i \in \mathbb{R}^{1\times d_k} \) and \( \mathbf{v}_i \in \mathbb{R}^{1\times d_v} \). A classical representation of such memory is the cumulative outer-product form $\mathbf{M}_t = \sum_{i=1}^{t} \mathbf{k}_i^\top \mathbf{v}_i $,
which encodes all previously observed associations. Retrieval corresponds to a mapping \( f_\mathbf{M} : \mathbb{R}^{d_k} \to \mathbb{R}^{d_v} \), parametrized by \( \mathbf{M} \), such that for any stored pair \((\mathbf{k}_i, \mathbf{v}_i)\), \( f_\mathbf{M}(\mathbf{k}_i) \approx \mathbf{v}_i \). The recall process is robust to small perturbations of the key, \emph{i.e.}, \( \mathbf{q}_i \approx \mathbf{k}_i \Rightarrow f_\mathbf{M}(\mathbf{q}_i) \approx \mathbf{v}_i \). In the simplest case, the associative map is linear: $f_\mathbf{M}(\mathbf{q}) = \mathbf{q}\mathbf{M}$, enabling direct recall of stored values from corresponding keys. In summary, an associative memory model is fully characterized by two components: \emph{\textbf{(i)}} the \emph{update rule (memory recurrence)} governing the evolution of \( \mathbf{M}_t \), and \emph{\textbf{(ii)}} the \emph{associative map} \( f_\mathbf{M} \) defining how stored information is retrieved.

\subsection{Kernel Fusion and FlashAttention}
\label{sec:flashattention-preliminary}
\captionsetup[wrapfigure]{skip=0pt}
\setlength{\columnsep}{10pt}
\begin{wrapfigure}{r}{0.47\linewidth}
  \vspace{-15pt}
  \begin{center}
    \includegraphics[width=0.99\linewidth]{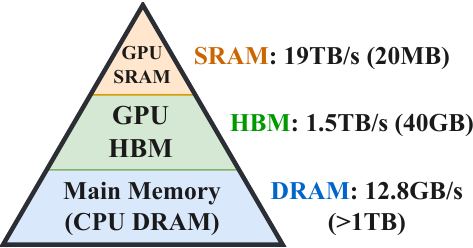}
  \end{center}
  \caption{Memory hierarchy with bandwidth \& memory size.}
  \vspace{-15pt}
  \label{fig:mem_hierarchy}
\end{wrapfigure}
Modern GPUs have steep memory hierarchies: small, fast on-chip SRAM and large, slower HBM (see Fig.~\ref{fig:mem_hierarchy}). In attention, the $N^2$ score and probability matrices are materialized in HBM and touched by multiple passes (\emph{e.g.} scaling, masking, Softmax, dropout). Each pass re-reads/writes these off-chip tensors, making bandwidth and launch overhead the runtime limiter. As sequence length grows, these quadratic intermediates with poor locality dominate cost.

FlashAttention~\cite{dao2022flashattention} (FA) introduces an I/O aware \emph{fused} kernel that tiles $\mathbf{Q}\mathbf{K}^\top$ and performs scaling, causal masking, numerically stable online softmax, dropout, and multiplies by $\mathbf{V}$ all in one streaming pass, keeping tiles in on-chip SRAM. By avoiding materialization of the $N{\times}N$ score/probability tensors in HBM, it removes off-chip round-trips and cuts kernel-launch overhead and bandwidth pressure, yielding substantial speedups. FA remains compatible with patterns such as \emph{sliding-window attention}: masks are applied within tiles so work outside the window is skipped, accelerating both forward and backward. This does not change memory complexity, as FA already achieves $\mathcal{O}(N)$ memory by never materializing the full attention matrix in HBM. In practice, FA is the default attention backend across modern LLM stacks; omitting it risks reduced practical relevance.
\section{Methodology}
\label{sec:method}
We start by defining the general \emph{causal form} of attention mechanism. Let $ \mathbf{X}^{(0)}\in \mathbb{R}^{N\times d}$ be the input embeddings, and for layer $l$, $\{\mathbf{Q}, \mathbf{K}, \mathbf{V}\}^{(l)}= \mathbf{X}^{(l-1)}\{\mathbf{W}_Q,\mathbf{W}_K, \mathbf{W}_V\}^{(l)}$. Conventionally, we denote the $i$-th row vector of matrix a $\mathbf{X}$ by $\mathbf{x}_i\in \mathbb{R}^{1\times d}$. Let $\mathbf{S}^{(l)}\in \mathbb{R}^{N\times N}$ be the attention score. For a causal attention where the visibility of each token $i\in [1, N]$ is determined by mask function $\mathcal{N}(i)$ to prevent foresee, then the $i,j$-th entry of the normalized score is defined by
\begin{equation}
        \resizebox{0.91\linewidth}{!}{\ensuremath{
\mathbf{S}^{(l)}_{ij}=\frac{
    \exp({\boldsymbol{\Phi}^{(l)}_{ij}})  \mathbf{1}\{j\in \mathcal{N}(i)\}
    }{\sum_{k\in\mathcal{N}(i)} \exp({\boldsymbol{\Phi}^{(l)}_{ik}}) }  \text{ where }  \boldsymbol{\Phi}^{(l)}_{ij}=\frac{\mathbf{q}^{(l)}_i (\mathbf{k}^{(l)}_j)^\top}{\sqrt{d_h}}.
}}
\end{equation}
Then, the output of $l$-th attention layer is generally $\mathbf{X}^{(l+1)} \xleftarrow[]{\mathrm{FFN}} \mathbf{O}^{(l)} = \mathbf{S}^{(l)}\mathbf{V}^{(l)}$.
Below, we denote the lower-triangular full \textbf{Softmax attention} score by $\bar{\mathbf{S}}$, so the admissible keys for query $i$ are in $\mathcal{N}(i)=\{j: j\le i\}$. We denote the \textbf{sliding-window attention} score width $w$ by $\hat{\mathbf{S}}$, where admissible keys for query $i$ are in $\mathcal{N}(i)=\{j:i-w < j\le i\}$. 

\subsection{Motivation through Associative Memory}
\label{sec:motivation}

\paragraph{Transformers as Associative Memory.}
Softmax attention can be viewed as an associative memory through an exponential kernel. Let $\phi(\cdot)$ denote a (possibly infinite-dimensional) feature map such that
\(
\exp  (\frac{\mathbf{q} \mathbf{k}^\top}{\sqrt{d_h}} )  =  \langle \phi(\mathbf{q}), \phi(\mathbf{k}) \rangle
\). Define the time-varying memory matrix
\(
\mathbf{M}_t  =  \sum_{j=1}^{t} \phi(\mathbf{k}_j)^\top \mathbf{v}_j    \in  \mathbb{R}^{\mathrm{dim}(\phi) \times d_v}
\). Lets ignore the normalization term for simplicity, the retrieval at time \(t\) is the linear associative map
$
    \tilde{\mathbf{o}}_t  
    =  \phi(\mathbf{q}_t) \mathbf{M}_t 
    =\sum_{j=1}^{t} \exp  ( \mathbf{q}_t \mathbf{k}_j^\top / \sqrt{d_h} ) \mathbf{v}_j .
$
Restoring the Softmax normalizer yields the standard attention output $\mathbf{o}_t  = 
\sum_{j=1}^{t} \frac{ \exp  (\mathbf{q}_t \mathbf{k}_j^\top /\sqrt{d_h} ) \mathbf{v}_j}
{\sum_{k=1}^{t} \exp  (\mathbf{q}_t \mathbf{k}_k^\top /\sqrt{d_h} )}$, 
\emph{i.e.}, full Softmax attention implements associative recall where the \textbf{memory} is the cumulative outer-product
\(
\mathbf{M}_t=\sum_{j\le t} \phi(\mathbf{k}_j)^\top \mathbf{v}_j
\)
and the \textbf{associative map} is \(f_{\mathbf{M}}(\mathbf{q})=\phi(\mathbf{q}) \mathbf{M}\).

With the knowledge above, we can investigate the \emph{associative memory update}. Reformulating the memory $\mathbf{M}_t$ gives
\begin{align}
    \underline{\mathbf{M}_t} = \sum_{i=1}^{t-1} \mathbf{k}_i^\top \mathbf{v}_i + \mathbf{k}_t^\top \mathbf{v}_t = \underline{\mathbf{M}_t + \mathbf{k}_t^\top \mathbf{v}_t}. \label{eq:associative_mem_update}
\end{align}
We refer to Eq.~\eqref{eq:associative_mem_update} to as the \textbf{memory recurrence} of the classical associative memory. Below, we extend this concept to the Softmax attention and SWA.
\begin{theorem}[Memory Recurrence of Exact Attention]\label{thm:recurrent_forms}
    Assume a perfectly defined feature map $\phi(\cdot): \mathbb{R}^{d} \to \mathbb{R}^{\mathrm{dim}(\phi)}$, such that $\langle \phi(\mathbf{q}), \phi(\mathbf{k}) \rangle$ approximates $\exp  (\frac{\mathbf{q} \mathbf{k}^\top}{\sqrt{d_h}})$ arbitrarily well, then the \underline{memory recurrence} of \textbf{Softmax attention} with normalization is formulated by
    \begin{align}
        \mathbf{M}_t = \frac{t-1}{t}\mathbf{M}_{t-1} + \frac{1}{t} \phi(\mathbf{k}_t)^{\top} \mathbf{v}_t ,\label{eq:recurrent_form_softmax}
    \end{align}
    and that of \textbf{SWA} ($t>w$) with normalization is formulated by
    \begin{align}
        \mathbf{M}_t = \mathbf{M}_{t-1} + \frac{1}{w} (\phi(\mathbf{k}_t)^{\top} \mathbf{v}_t  - \phi(\mathbf{k}_{t-w})^{\top} \mathbf{v}_{t-w} ).\label{eq:recurrent_form_swa}
    \end{align}
\end{theorem}
\paragraph{Optimization Objective of Memory Recurrence.}
Regarding the \emph{memory recurrence} as a single-step gradient descent update on $\mathbf{M}$ with step-size $1$, we have
\begin{align}
    \mathbf{M}_t = \mathbf{M}_{t-1} - (- \mathbf{k}_t^\top \mathbf{v}_t) = \mathbf{M}_{t-1} - \frac{\partial \mathcal{L}_t(\mathbf{M}_{t-1})}{\partial \mathbf{M}_{t-1}}, \label{eq:general_opt_obj}
\end{align}
where the \textbf{objective} $\mathcal{L}_t(\mathbf{M}_{t-1}) = -\langle \phi(\mathbf{k}_t) \mathbf{M}_{t-1} , \mathbf{v}_t \rangle$. Essentially, such objective aims to update $\mathbf{M}$ such that recalling $\mathbf{v}_t$ from the updated memory using $\mathbf{k}_t$ is as effective as possible. 
Understanding the objective, we have the following:
\begin{proposition}[Optimization Objective of Exact Attention]\label{prop:optimization_objs}
    With the memory recurrence defined in Thm.~\ref{thm:recurrent_forms}, by Eq.~(\ref{eq:general_opt_obj}), we can solve for the optimization objectives $ \mathcal{L}_t(\mathbf{M}_{t-1})$ respectively for Softmax attention and SWA to fit the form in Eq.~\eqref{eq:recurrent_form_softmax} and Eq.~\eqref{eq:recurrent_form_swa}. For \textbf{Softmax attention}, the \underline{objective} is solved as
    \begin{align}
        \mathcal{L}_t(\mathbf{M}_{t-1}) = \frac{1}{2t} \| \mathbf{M}_{t-1}\|^2_F - \frac{1}{t} \phi(\mathbf{k}_t) \mathbf{M}_{t-1} \mathbf{v}_t^\top,\label{eq:obj_softmax}
    \end{align}
    and for \textbf{SWA} ($t>w$) is
    \begin{equation}
        \resizebox{0.91\linewidth}{!}{\ensuremath{
\mathcal{L}_t(\mathbf{M}_{t-1}) = 
\frac{1}{w}(\phi(\mathbf{k}_{t-w})\mathbf{M}_{t-1}\mathbf{v}_{t-w}^\top - \phi(\mathbf{k}_{t})\mathbf{M}_{t-1}\mathbf{v}_{t}^\top).
}} \label{eq:obj_swa}
    \end{equation}
    One can verify the proposition by plugging Eq.~\eqref{eq:obj_softmax} and Eq.~\eqref{eq:obj_swa} back into Eq.~\eqref{eq:general_opt_obj} to derive the exact recurrences as Thm.~\ref{thm:recurrent_forms}. 
\end{proposition}

\paragraph{Limitations.}
From Prop.~\ref{prop:optimization_objs}, on one hand, we observe in Eq.~\eqref{eq:obj_softmax} that, as $t$ increases, $\frac{1}{t}\to 0$, then $\mathcal{L}_t(\mathbf{M}_{t-1}) \to 0$. This indicates the \emph{\textbf{gradient vanishing}} nature of \emph{Softmax attention} \emph{w.r.t.} $\mathbf{M}$ induced by densified Softmax normalization. On the other hand, by using the identity $\mathbf{a}\mathbf{M} \mathbf{b}^\top = \langle \mathbf{M}, \mathbf{a}^\top \mathbf{b}\rangle_F$ of Frobenius norm, we can reformulate Eq.~\eqref{eq:obj_swa} as
\begin{align}
    \mathcal{L}_t(\mathbf{M}_{t-1}) = 
\frac{1}{w}\langle\mathbf{M}_{t-1}, \phi(\mathbf{k}_{t-w})^\top\mathbf{v}_{t-w} - \phi(\mathbf{k}_{t})^\top\mathbf{v}_{t}\rangle, \label{eq:obj_swa_inner_prod}
\end{align}
let $\Delta_t = \phi(\mathbf{k}_{t-w})^\top\mathbf{v}_{t-w} - \phi(\mathbf{k}_{t})^\top\mathbf{v}_{t}$ we have $\mathcal{L}_t(\mathbf{M}_{t-1}) = \frac{1}{w} \langle \mathbf{M}_{t-1}, \Delta_t \rangle$. Since $\nabla_{\mathbf{M}} \mathcal{L}_t = \frac{1}{w} \Delta_t$, we observe the objective is linearly unbounded optimizing over $\mathbf{M}$ as we can pick arbitrarily small $\mathbf{M} = \alpha \Delta_t$ with $\alpha \to -\infty$ to obtain $\mathcal{L}_t \to -\infty$. This reveals the \emph{\textbf{gradient instability}} nature of \emph{sliding window attention}. Conclusively, as Softmax attention and SWA respectively suffer from gradient vanishing and instability, we are sufficiently motivated to derive a new attention mechanism that has both \emph{bounded and stable gradient} while maintaining the linear complexity of SWA and hardware-friendliness of FlashAttention.

\begin{figure}[t]
    \centering
    \includegraphics[width=0.85\linewidth]{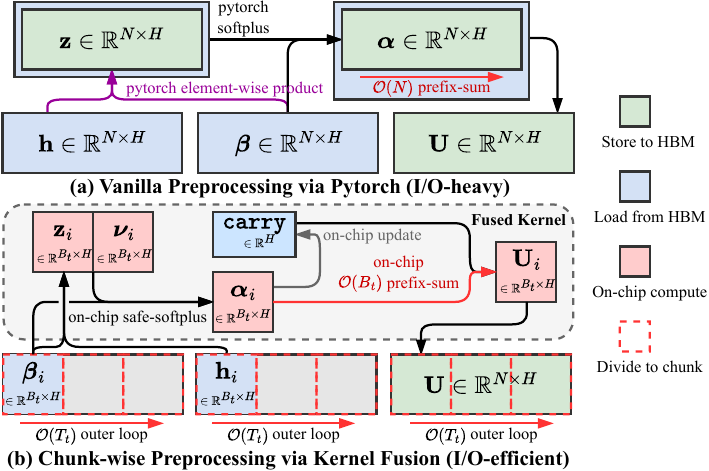}
    \caption{Schematic comparison between \textbf{(upper)} vanilla preprocessing and \textbf{(lower)} our 1-pass fused preprocessing.}
    \label{fig:algorithm_cmp_preprocess}
\end{figure}
\subsection{The GatedFWA}
We build our attention upon SWA. A core difference of GatedFWA is the introduction of memory gate. Per layer $l$, we define the data-dependent memory gate at query position $t$ as
\begingroup
\allowdisplaybreaks
\begin{align}
    &\boldsymbol{\alpha}^{(l)}_t=\frac{1}{\boldsymbol{\beta}^{(l)}_t} \odot \mathrm{softplus} ( \boldsymbol{\beta}^{(l)}_t \odot  \mathbf{h}^{(l)}_t  ) \in \mathbb{R}^{H}> 0, \text{where}\\
    &\mathbf{h}^{(l)}_t=  \underbrace{\mathbf{x}^{(l)}_t \mathbf{W}^{(l)}_g + \mathbf{b}_g^{(l)}}_{\text{(gate pre-activation)}}, \ 
        \boldsymbol{\beta}^{(l)}_t= \underbrace{\mathbf{1}+\mathrm{elu}( \mathbf{x}^{(l)}_t \mathbf{W}^{(l)}_\beta)}_{\text{(amplitude)}} .\label{eq:preactivation_and_amplitude}
\end{align}
\endgroup
Essentially, the $\mathbf{h}^{(l)}_t \in \mathbb{R}^H$ is the gate pre-activation be with $\mathbf{W}_g\in \mathbb{R}^{d \times H}$. We design $\boldsymbol{\beta}^{(l)} \in \mathbb{R}^H >0$ as the amplitude where $\texttt{init}(\mathbf{W}_\beta) = \mathbf{0}_{d\times H}$ so that $\boldsymbol{\beta}_\mathrm{init} = \mathbf{1}_{H}$, giving the attention a mild startup. $\odot$ is the element(head)-wise product. We obtain the preprocessed matrix $\mathbf{U}=\{\mathbf{u}_i\}_{i=1}^N \in \mathbb{R}^{N\times H}$ (materialized) and the gated logits matrix $\mathbf{B}_{ij}$ (for $t \ge j$) by
\begin{align}
&\mathbf{u}^{(l)}_t=\sum\nolimits_{q=1}^{t} - \boldsymbol{\alpha}^{(l)}_q \prec \mathbf{0},\  \mathbf{B}^{(l)}_{tj}=\mathbf{u}^{(l)}_t-\mathbf{u}^{(l)}_j \prec \mathbf{0}.
\end{align}
The GatedFWA then incorporate the gated logits into attention logits as $(\boldsymbol{\Phi}^{(l)}_{ij}+\mathbf{B}^{(l)}_{ij})$, yield element-wise,
\begin{align}
\tilde{\mathbf{S}}^{(l)}_{ij}
=\frac{\exp    (\boldsymbol{\Phi}^{(l)}_{ij} + \mathbf{B}^{(l)}_{ij}  )  \mathbf{1}\{ j:i-w< j\le i\}}
{\sum_{ k=i-w+1 }^{i}\exp    (\boldsymbol{\Phi}^{(l)}_{ik} + \mathbf{B}^{(l)}_{ik}  )}  .
\end{align}
Similar to the analysis in Sec.~\ref{sec:motivation}, below, we illustrate with $H=1$ for brevity and interpret the mechanism of GatedFWA through the framework of associative memory.
\begin{proposition}[Memory Recurrence and Optimization Objective of GatedFWA] \label{prop:recurrence_gatedfwa}
Assume a feature map $\phi$ similar to Thm.~\ref{thm:recurrent_forms} such that $\langle \phi(\mathbf{q}), \phi(\mathbf{k}) \rangle\approx \exp(\frac{\mathbf{q}\mathbf{k}^\top}{\sqrt{d_h}})$, then the memory recurrence of \textbf{GatedFWA} with normalization is formulated by
\begin{align}
    \mathbf{M}_t &= (\exp(-\boldsymbol{\alpha}_t)\mathbf{I}_k)\mathbf{M}_{t-1} \label{eq:mem_recurrence_gatedfwa} \\
    &+ \frac{1}{w} (\phi(\mathbf{k}_t)^\top \mathbf{v}_t 
    - 
    (\mathbf{c}_t\mathbf{I}_k)
    \phi(\mathbf{k}_{t-w})^\top \mathbf{v}_{t-w}), \nonumber
\end{align}
where $\mathbf{c}_t = \prod_{j=t-w+1}^{t-1}\exp(- \boldsymbol{\alpha}_j) \in (0,1)$. And therefore, the optimization objective is formulated by
\begin{align}
    \mathcal{L}_t(\mathbf{M}_{t-1}) &= \frac{1}{2}\| ( \sqrt{1-\exp(\boldsymbol{-\alpha}_t)}  \mathbf{I}_k ) \mathbf{M}_{t-1}\|_F^2  \label{eq:obj_gatedfwa}\\
    &- 
    \frac{1}{w} \langle
    \mathbf{M}_{t-1}
    ,
    \mathbf{c}_t\mathbf{I}_k
    \Delta_t
    + (1-\mathbf{c}_t)\mathbf{I}_k 
    \phi(\mathbf{k}_t)\mathbf{v}_t^\top
    \rangle, \nonumber
\end{align}
where $\Delta_t = \phi(\mathbf{k}_{t-w})^\top\mathbf{v}_{t-w} - \phi(\mathbf{k}_{t})^\top\mathbf{v}_{t}$ (same as Eq.~\eqref{eq:obj_swa_inner_prod}).
\end{proposition}

\begin{algorithm}[t]
\caption{Gate Preprocessing (Fused Tiled Scan) Kernel}
\label{alg:fused-gated-scan}
\begin{adjustbox}{max width=1.0\linewidth}
\begin{minipage}{1.0\linewidth}
\small
\begin{algorithmic}[1]
\Require Matrices $\mathbf{H}\gets \mathbf{X}\mathbf{W}_g\in\mathbb{R}^{N\times H}$,  $\boldsymbol{\beta}\gets \mathbf{1} + \mathrm{elu}(\mathbf{X}\mathbf{W}_{\beta})\in\mathbb{R}^{N\times H}$ and $\mathbf{U}\gets \mathbf{0}_{N\times H}$ in HBM, chunk size $B_t$, small $\varepsilon>0$.
\State Divide $\mathbf{H}$, $\boldsymbol{\beta}$ and $\mathbf{U}$ into $T_t=\left\lceil \frac{N}{B_t}\right\rceil$ blocks $\mathbf{h}_1,\ldots,\mathbf{h}_{T_t}$, $\boldsymbol{\beta}_1,\ldots,\boldsymbol{\beta}_{T_t}$ and $\mathbf{u}_1,\ldots,\mathbf{u}_{T_t}$ of size $B_r\times H$.
  \State On chip register, set $\textsc{carry}\gets \mathbf{0}_H$.
  \For{$1 \le i \le T_t$} 
    \State Load chunk $\mathbf{h}_i$, $\boldsymbol{\beta}_i$ and $\mathbf{u}_i$ from HBM to SRAM.
    \State On chip, compute $\mathbf{z}_i\gets \boldsymbol{\beta}_i \odot \mathbf{h}_i$, $\boldsymbol{\nu}_i\gets \max(\mathbf{z}_i,0)$.
    \State On chip, compute $\mathrm{softplus}(\mathbf{z}_i)\gets \boldsymbol{\nu}_i + \log (e^{\mathbf{z}_i-\boldsymbol{\nu}_i} + e^{-\boldsymbol{\nu}})$.
    \State On chip, compute $\boldsymbol{\alpha}_i\gets \mathrm{softplus}(\mathbf{z}_i) \odot (\boldsymbol{\beta}_i+\varepsilon)^{-1}$. 
    \State On chip, compute $\mathbf{p}_i\gets \mathrm{cumsum}(-\boldsymbol{\alpha}_i) + \textsc{carry}$.
    \State Write $\mathbf{u}_i \gets \mathbf{p}_i$ to HBM.
    \State On chip register, update $\textsc{carry}\gets \textsc{carry} + \sum - \boldsymbol{\alpha}_i$.
\EndFor
\State \Return $\mathbf{U}$.
\end{algorithmic}
\end{minipage}
\end{adjustbox}
\end{algorithm}
\paragraph{Implication from Memory Recurrence.} Inspecting the gradient path of GatedFWA, by Eq.~\eqref{eq:mem_recurrence_gatedfwa} have $\frac{\partial \mathbf{M}_t}{\partial \mathbf{M}_{t-1}} = \exp(-\boldsymbol{\alpha}_t)\mathbf{I}_k$, compared to $\frac{\partial \mathbf{M}_t}{\partial \mathbf{M}_{t-1}}=\mathbf{1}$ for SWA. The sensitivity of any loss $\mathcal{L}_t$ that requires reading out $\mathbf{M}_t$ at time $t$ \emph{w.r.t.} a much earlier memory state $\mathbf{M}_p$ ($p < t$) becomes
\begin{align}
    \frac{\partial \mathcal{L}_t}{\partial \mathbf{M}_p} =  (\prod_{i=p+1}^t \frac{\partial \mathbf{M}_i}{\partial \mathbf{M}_{i-1}}) \frac{\partial \mathcal{L}_t}{\partial \mathbf{M}_t} = (\prod_{i=p+1}^t \exp({-\boldsymbol{\alpha}_i}) \mathbf{I}_k) \frac{\partial \mathcal{L}_t}{\partial \mathbf{M}_t} .\nonumber
\end{align}
Therefore, instead of an uncontrollable gradient path, the GatedFWA has a learnable controllable path, where the model can learn to: \textbf{\emph{(i) preserve gradients}}:  By setting $\boldsymbol{\alpha}_i$ close to $0$, it allows gradients to flow back many steps, capturing long-term dependencies (within the constraints of the window $w$); \textbf{\emph{(ii) block gradients}}: By setting $\boldsymbol{\alpha}_t$ to $+\infty$, it cuts off the gradient flow, preventing irrelevant information from the past from interfering with the current parameter updates. The amplitude $\boldsymbol{\beta}$ (via $\boldsymbol{\alpha}=\boldsymbol{\beta}^{-1}\mathrm{softplus}(\boldsymbol{\beta} \odot \mathbf{h})$) controls how sharply the gate switches between \textit{preserve} (small $\boldsymbol{\alpha}$) and \textit{block} (large $\boldsymbol{\alpha}$), letting the model adapt its effective attention horizon token by token. 

\paragraph{Implication from Optimization Objective.} Recalling the \textbf{\emph{gradient vanishing}} issue in Softmax attention, where $\mathcal{L}_t(\mathbf{M}_{t-1}) \to 0$ as $t\to +\infty$, the Eq.~\eqref{eq:obj_gatedfwa} is invariant to $t$ but dependent to a fixed $w$ by the introduction of sliding window, thus the gradient never vanishes due to marginal utility. For \textbf{\emph{gradient instability}} of SWA, where the norm of $\mathbf{M}_{t-1}$ is encouraged to be as big as possible to facilitate a big $\langle \mathbf{M}_{t-1}, \Delta_t\rangle$, we can observe in Eq.~\eqref{eq:obj_gatedfwa} that the objective of GatedFWA introduced a soft-L2 normalization on $\mathbf{M}_{t-1}$, which discourages a absolutely large $\|\mathbf{M}_{t-1}\|$. Also, the second term in Eq.~\eqref{eq:obj_gatedfwa} introduced a scaling factor $\mathbf{c}_t\in (0,1)$ (typically small), which trades off the associate memory between aligning with history context and current context, making the memory update more expressive and controllable.

\subsection{Hardware-Aligned Design}

We adopt an I/O-aware two-phase design that mirrors efficient attention implementations: \emph{\textbf{(i)} preprocessing phase}, that turns token-local gate features into a cumulative, non-positive gate vector $\mathbf{U}$, \emph{\textbf{(ii)}} a tiled \emph{attention compute phase}, that injects the memory gate on-the-fly inside a FA streaming Softmax under a sliding window, avoiding materialization of any $N\times N$ score matrix and minimizing HBM traffic. $^\star$We implement all our kernels with \textbf{Triton}-lang \cite{tillet2019triton}.

\paragraph{(1) Gate Preprocessing via Fused Tiled Scan.}
As described in Alg.~\ref{alg:fused-gated-scan}, we compute, for each head row, the gated prefix $\mathbf{u}_t  =  -\sum_{i=1}^{t} \frac{\mathrm{softplus}(\boldsymbol{\beta}_i \odot \mathbf{h}_i)}{\boldsymbol{\beta}_i+\varepsilon}$ in a \emph{single streaming pass} over the sequence. The kernel tiles the time axis into chunks of length $B_t$ and keeps a tiny carry vector on-chip so we never materialize intermediates in HBM.

\begin{figure}[t]
    \centering
    \vspace{-5pt}
    \includegraphics[width=\linewidth]{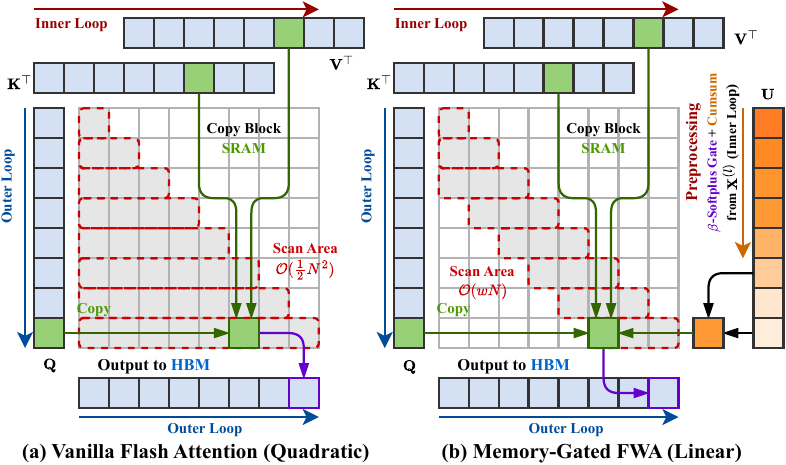}
    \caption{Schematic comparison between \textbf{(left)} Flash Attention (\textit{Dao et.~al}) and \textbf{(right)} our Hardware-efficient GatedFWA.}
    \label{fig:algorithm_cmp_gsfa}
\end{figure}

\paragraph{Efficiency.} We launch one program per head so rows run in parallel. Inside a tile, we perform parallel inclusive scan across the tile’s time positions. The only serial dependency is the single-scalar \emph{carry} that threads results between tiles of the same row. Compared to a two-kernel PyTorch path ($\mathrm{elementwise}$ then $\mathrm{cumsum}$) as illustrated in Fig.~\ref{fig:algorithm_cmp_preprocess}, our design: reads $\mathbf h,\boldsymbol\beta$ once and writes $\mathbf U$ once, without intermediate $\mathbf{z}, \boldsymbol{\alpha}$ tensor in HBM, which avoids an extra kernel launch and global synchronization. We further use the numerically safe $\mathrm{softplus}(u)=\nu+\log(\exp(u-\nu)+\exp(-\nu))$, $\nu=\max(u,0)$, and accumulate in fp32 to avoid overflow.

\paragraph{(2) Attention Computation.}
As described in Alg.~\ref{alg:gated_swfa}, we fuse the gate into FA streaming Softmax with primarily \textbf{three changes} (wrapped in \colorbox{gray!30}{grey}): \emph{\textbf{(i)}} window-aware column pruning so we iterate only over key tiles that can intersect the sliding window; \emph{\textbf{(ii)}} a broadcasted additive bias that injects the gate via $\boldsymbol{\Phi} \leftarrow \mathbf{QK}^\top + \mathbf{U}^q\mathbf{1}^\top - \mathbf{1}(\mathbf{U}^k)^\top$ (realizing $\mathbf{B}_{ij}=\mathbf{u}_i-\mathbf{u}_j$ without materializing $N\times N$); and \emph{\textbf{(iii)}} in-tile SWA masking (keep $q-w+1\le g\le q$, else $-\infty$). Everything else, \textit{e.g.} tiling, online rowwise max/sum, rescaling, and output accumulation remains identical to vanilla FA, so stability and numerics are preserved.

\paragraph{Efficiency.} The overhead from gating is minimal, while the global complexity is significantly improved to linear $\mathcal{O}(N)$ benefiting from the window attention: two extra vector loads ($B_r,B_c$) and $O(B_rB_c)$ adds per tile, negligible \textit{vs.} the $\mathcal{O}(B_rB_c d)$ GEMM. Complexity matches SWA: with time $\mathcal{O} (N w d)$, HBM traffic $\mathcal{O}(Nd+Nw)$ and on-chip working set $\mathcal{O}(B_r d + B_c d + B_r + B_c)$; we parallelize over heads and row tiles unchanged. Backprop reuses the standard FA streaming factors; gradients into $\mathbf{U}$ flow through the same streamed scan used in preprocessing so no $N\times N$ tensors are ever materialized.
\begin{figure*}[t]
\vspace{-10pt}
    \centering
    \begin{subfigure}[t]{0.245\linewidth}
        \centering
        \includegraphics[width=\textwidth]{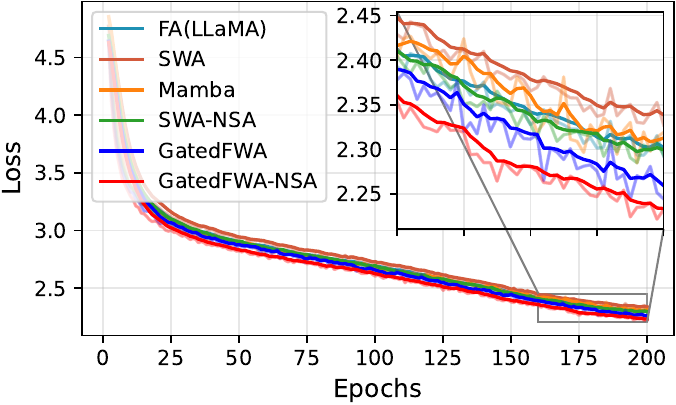}
        \caption{Train losses @ WikiText103}
    \end{subfigure}
    \hfill
    \begin{subfigure}[t]{0.245\linewidth}
        \centering
        \includegraphics[width=\textwidth]{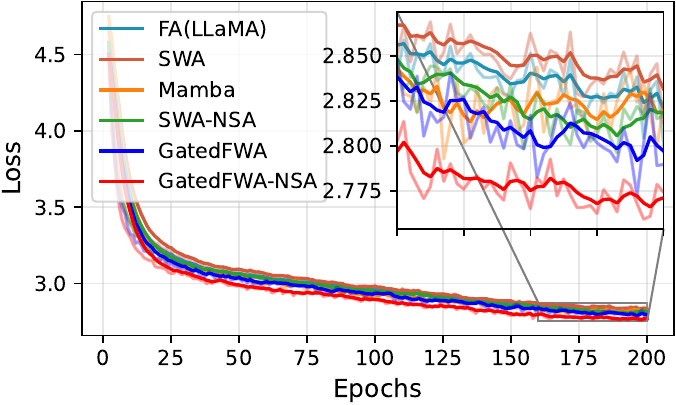}
        \caption{Val losses @ WikiText103}
    \end{subfigure}
    \hfill
    \begin{subfigure}[t]{0.245\linewidth}
        \centering
        \includegraphics[width=\textwidth]{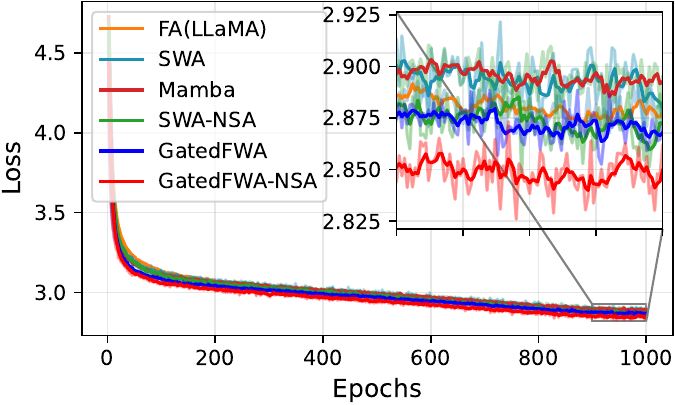}
        \caption{Train losses @ OpenWebText}
    \end{subfigure}
    \hfill
    \begin{subfigure}[t]{0.245\linewidth}
        \centering
        \includegraphics[width=\textwidth]{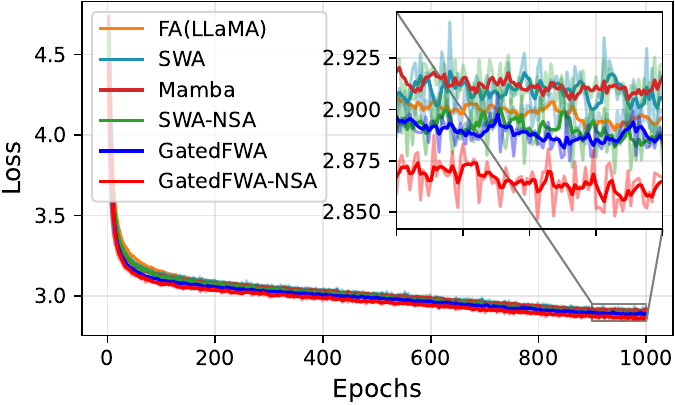}
        \caption{Val losses @ OpenWebText}
    \end{subfigure}
    \caption{Language pretraining loss (curves smoothed via EMA) on WikiText103 and OpenWebText where $N=4096, w=512$.}
    \label{fig:train_val_curves}
\end{figure*}

\begin{figure*}[t]
  \centering
  \vspace{-10pt}
  \captionof{table}{Language modelling scaling law against LLaMA(w/ and w/o SWA), RetNet, RWKV, and Mamba. All models are trained on the OpenWebText dataset. Models vary from $120$-$360$M parameters and $1024$-$4096$ context length.}
  \begin{minipage}[t]{0.615\linewidth}
    \vspace{0pt} 
    \centering
    \label{tab:val-loss}
    \resizebox{\linewidth}{!}{%
      \begin{NiceTabular}{@{}l|cccccc@{}}
\toprule
\multicolumn{1}{c|}{}                                        &                                          & \multicolumn{2}{c}{\textbf{Val. Loss ($\downarrow$)}} &                                          & \multicolumn{2}{c}{\textbf{Val. Loss ($\downarrow$)}} \\ \cmidrule(lr){3-4} \cmidrule(l){6-7} 
\multicolumn{1}{c|}{\multirow{-2}{*}{\textbf{Architecture}}} & \multirow{-2}{*}{\textbf{\# Param. (M)}} & $N=1024$                  & $N=4096$                  & \multirow{-2}{*}{\textbf{\# Param. (M)}} & $N=1024$                  & $N=4096$                  \\ \midrule
\rowcolor[HTML]{FFFFFF} 
RetNet                                                       & 129.1                                    & 3.569                     & 3.492                     & 373.2                                    & 3.362                     & 3.227                     \\
\rowcolor[HTML]{FFFFFF} 
GLA                                                          & 123.8                                    & 3.381                     & 3.364                     & 361.1                                    & 3.018                     & 3.001                     \\
\rowcolor[HTML]{FFFFFF} 
RWKV                                                         & 124.4                                    & 3.291                     & 3.276                     & 354.8                                    & 2.983                     & 2.931                     \\
\rowcolor[HTML]{FFFFFF} 
Mamba                                                        & 129.2                                    & 3.238                     & {\ul 3.231}               & 371.5                                    & 2.902                     & {\ul 2.868}               \\ \midrule
\rowcolor[HTML]{EFEFEF} 
Transformer (LLaMA)                                          & 124.4                                    & 3.247                     & 3.273                     & 357.7                                    & 2.891                     & 2.883                     \\
\rowcolor[HTML]{EFEFEF} 
\textit{$\quad$ + SWA}                                       &    124.4            & 3.248                     & 3.274                     &    357.7            & 2.892                     & 2.887                     \\
\rowcolor[HTML]{EFEFEF} 
\textit{$\quad$ + SWA + NSA}                                 & 125.4                                    & 3.240                     & 3.248                     & 361.8                                    & {\ul 2.870}                     & {\ul 2.868}                     \\ \midrule
\rowcolor[HTML]{DFDFDF} 
\textit{$\quad$ + GatedFWA}                                  & 125.1                                    & {\ul 3.237}               & 3.255                     & 360.7                                    & 2.874               & 2.871                     \\
\rowcolor[HTML]{DFDFDF} 
\textit{$\quad$ + GatedFWA + NSA}                            & 126.1                                    & \textbf{3.215}            & \textbf{3.230}            &    362.7            & \textbf{2.859}            & \textbf{2.842}            \\ \bottomrule
\end{NiceTabular}
    }
  \end{minipage}\hfill
  \begin{minipage}[t]{0.375\linewidth}
    \vspace{0pt}
    \centering
    \includegraphics[width=\textwidth]{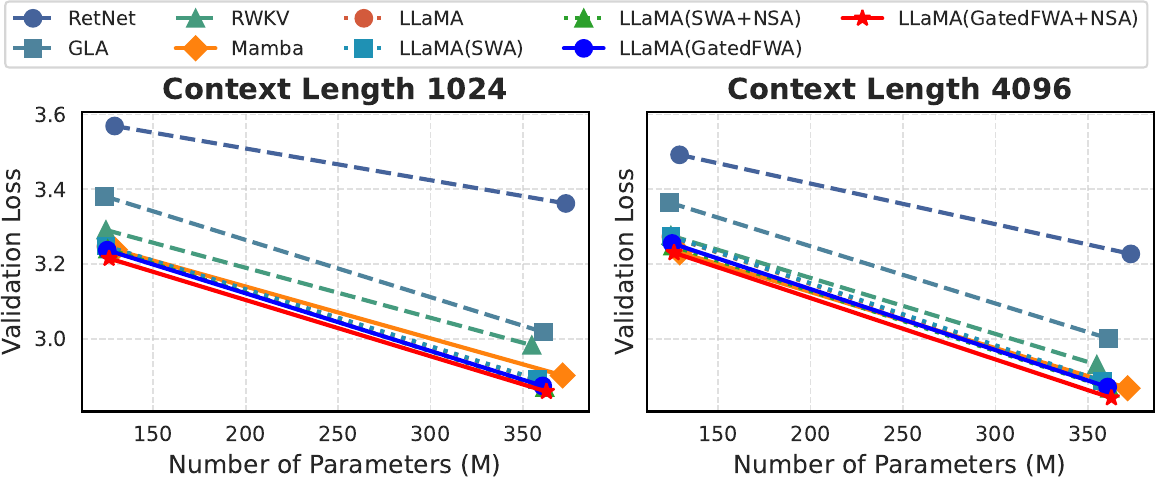}
    \captionof{figure}{Scaling law with $1024$ and $4096$ context length on OpenWebText dataset.}
    \label{fig:side-image}
  \end{minipage}
  \label{tab:scaling_law}
  \vspace{-10pt}
\end{figure*}

\begin{algorithm}[t]
\caption{GatedFWA Attention Fused Kernel}
\label{alg:gated_swfa}
\begin{adjustbox}{max width=1.0\linewidth}
\begin{minipage}{1.25\linewidth}
\small
\begin{algorithmic}[1]
\Require Matrices $\mathbf{Q},\mathbf{K},\mathbf{V}\in\mathbb{R}^{N\times d}$, $\mathbf{U}=\{\mathbf{u}_i\}_{i=1}^N\in \mathbb{R}^{N}$ in HBM, block sizes $B_c$, $B_r$, window $w\ge 1$.
\State Divide $\mathbf{Q}$ into $T_r=\left\lceil \frac{N}{B_r}\right\rceil$ blocks $\mathbf{q}_1,\ldots,\mathbf{q}_{T_r}$ of size $B_r\times d$, and divide $\mathbf{K},\mathbf{V}$ into $T_c=\left\lceil \frac{N}{B_c}\right\rceil$ blocks $\mathbf{k}_1,\ldots,\mathbf{k}_{T_c}$, $\mathbf{v}_1,\ldots,\mathbf{v}_{T_c}$ of size $B_c\times d$ each.
\State Divide $\mathbf{O}$ into $T_r$ blocks $\mathbf{o}_1,\ldots,\mathbf{o}_{T_r}$ of size $B_r\times d$, and $L$ into $L_1,\ldots,L_{T_r}$ of size $B_r$ each.
\State \colorbox{gray!30}{Init $\mathbf{U}^q=\mathbf{U}$, $\mathbf{U}^k=\mathbf{U}$ on HBM. Divide $\mathbf{U}^q$, $\mathbf{U}^k$ into $T_r$, $T_c$ blocks.}
\For{$1 \le i \le T_r$} 
  \State Load $\mathbf{q}_i$, $\mathbf{u}^q_i$ from HBM to on-chip SRAM.
  \State On chip, initialize $\mathbf{o}_i^{(0)}\gets\mathbf{0}_{B_r\times d}$, $\ell_i^{(0)}\gets(0)_{B_r}$, $m_i^{(0)}\gets(-\infty)_{B_r}$.
  \State \colorbox{gray!30}{$r_{\mathrm{start}}\gets (i-1)B_r$, $r_{\mathrm{end}}\gets \min(iB_r,N)-1$.}
  \State \colorbox{gray!30}{$k_{\mathrm{lo}}\gets \max  (0, r_{\mathrm{start}}-w+1 )$, $k_{\mathrm{hi}}\gets r_{\mathrm{end}}+1$.}
  \State \colorbox{gray!30}{$j_{\mathrm{lo}}\gets \left\lfloor \frac{k_{\mathrm{lo}}}{B_c} \right\rfloor + 1$, $j_{\mathrm{hi}}\gets \left\lceil \frac{k_{\mathrm{hi}}}{B_c} \right\rceil$.}
  \For{\colorbox{gray!30}{$j = j_{\mathrm{lo}} \ldots j_{\mathrm{hi}}$}} 
    \State Load $\mathbf{k}_j, \mathbf{v}_j$, $\mathbf{u}^k_j$ from HBM to on-chip SRAM.
    \State \colorbox{gray!30}{On chip, compute $\boldsymbol{\Phi}_i^{(j)} \gets \mathbf{q}_i \mathbf{k}_j^\top
    +
    \mathbf{u}^q_i \mathbf{1}^\top - \mathbf{1}(\mathbf{u}^k_j)^\top
    \in \mathbb{R}^{B_r\times B_c}$.}
    \For{each row $r\in[0,B_r)$ and col $c\in[0,B_c)$ with global indices $q=r_{\mathrm{start}}+r$, $g=(j-1)B_c+c$}  
    	\State  \colorbox{gray!30}{keep $\boldsymbol{\Phi}_i^{(j)}[r,c]$ iff $q-w+1 \le g \le q$; otherwise set to $-\infty$.}
    \EndFor
    \State On chip, compute $m_i^{(j)}\gets\max  (m_i^{(j-1)}, \operatorname{rowmax}(\boldsymbol{\Phi}_i^{(j)}) )$, $\tilde{\mathbf{p}}_i^{(j)}\gets \exp  (\boldsymbol{\Phi}_i^{(j)}-m_i^{(j)} )$, $\tilde \ell_i^{(j)}\gets\operatorname{rowsum}(\tilde{\mathbf{p}}_i^{(j)})$.
    \State On chip, $\mathbf{o}_i^{(j)}\gets\mathrm{diag}  (e^{m_i^{(j-1)}-m_i^{(j)}} ) \mathbf{o}_i^{(j-1)} + \tilde{ \mathbf{p}}_i^{(j)} \mathbf{v}_j$.
    \State On chip, $\ell_i^{(j)}\gets e^{m_i^{(j-1)}-m_i^{(j)}} \ell_i^{(j-1)} + \tilde \ell_i^{(j)}$.
  \EndFor
  \State \colorbox{gray!30}{On chip, compute $\mathbf{o}_i \gets \mathrm{diag}  (\ell_i^{(j_{\mathrm{hi}})})^{-1}   \mathbf{o}_i^{(j_{\mathrm{hi}})}$, $L_i \gets m_i^{(j_{\mathrm{hi}})} + \log  (\ell_i^{(j_{\mathrm{hi}})} )$.}
  \State Write $\mathbf{o}_i$, $L_i$ to HBM as the $i$-th block of $\mathbf{O}$ and $L$.
\EndFor
\State \Return $\mathbf{O}$ and $L$.
\end{algorithmic}
\end{minipage}
\end{adjustbox}
\end{algorithm}

\paragraph{(3) Block Structure and Compatibility.} We describe the computation of a GatedFWA attention block (depicted in Fig.~\ref{fig:overall_framework}) in detailed matrix form: each attention layer takes an input sequence $\mathbf{X}\in\mathbb{R}^{N\times d}$. We first describe the computation for the single head case with head dimension $d_H$. For the $\mathcal{O}(N)$ pre-filling phase, we first compute the normalized query $\mathbf{Q}\in\mathbb{R}^{N\times d}$, normalized key $\mathbf{K}\in\mathbb{R}^{N\times d}$, value $\mathbf{V}\in\mathbb{R}^{N\times d}$, gate parameters $\boldsymbol{\beta}\in\mathbb{R}^{N\times H}$, $\mathbf{h}\in\mathbb{R}^{N\times H}$, then for each head $h\in [1,H]$, the $\texttt{AttnLayer}$ executes:
\begin{figure}[H]
    \centering
    \vspace{-10pt}
    \includegraphics[width=\linewidth]{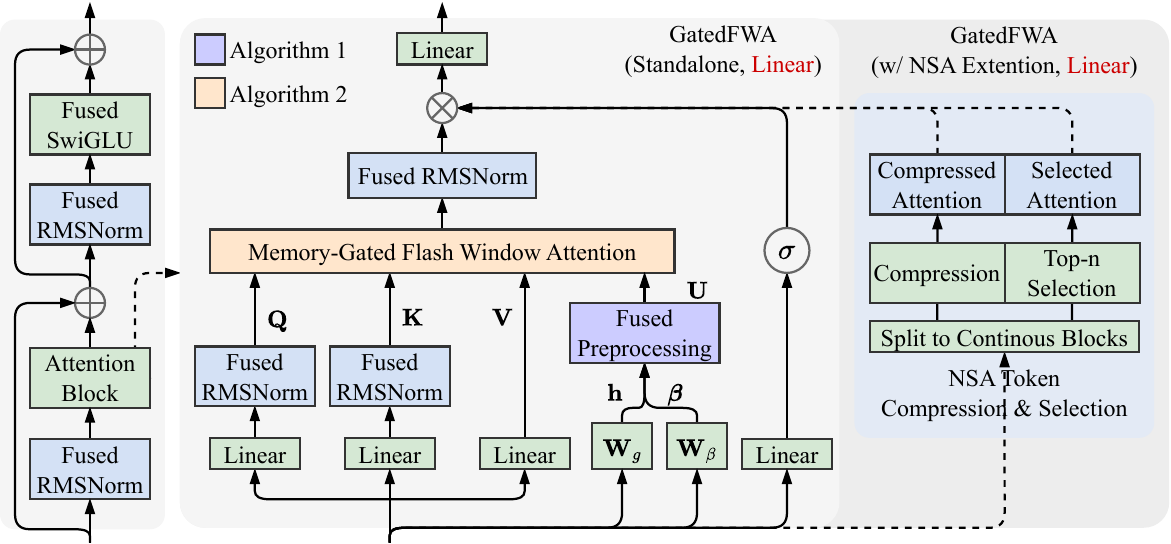}
    \caption{Overall architecture. \textbf{(left)} GatedFWA-Transformer \texttt{block}. \textbf{(middle)} A single standalone GatedFWA \texttt{layer}. \textbf{(right)} A NSA extension that seamlessly connect to GatedFWA w/o increasing complexity. $\sigma$ and $\otimes$ are sigmoid and Hadamard product.}
    \label{fig:overall_framework}
    \vspace{-25pt}
\end{figure}
\begingroup
\allowdisplaybreaks
\begin{align}
    & \mathbf{U}^{(l,h)} = \texttt{preprocess}(\boldsymbol{\beta}^{(l,h)}, \mathbf{h}^{(l,h)})\in \mathbb{R}^{N\times 1}, \nonumber\\
    & \tilde{\mathbf{O}}^{(l,h)} = \texttt{GatedFWA}(\mathbf{Q}^{(l,h)}, \mathbf{K}^{(l,h)}, \mathbf{V}^{(l,h)}, \mathbf{U}^{(l,h)}) \in \mathbb{R}^{N\times d_H}, \nonumber\\
    & \tilde{\mathbf{O}}^{(l)} = \texttt{concat} (\texttt{norm}(\tilde{\mathbf{O}}^{(l,1)}),\dots, \texttt{norm}(\tilde{\mathbf{O}}^{(l,H)})) \in \mathbb{R}^{N\times d}, \nonumber\\
    & \mathbf{G}^{(l)} = \texttt{swish}(\texttt{linear}(\mathbf{X}^{(l)})), \in \mathbb{R}^{N\times d}\nonumber\\
    & \mathbf{O}^{(l)} = (\mathbf{G}^{(l)}\odot \tilde{\mathbf{O}}^{(l)}) \mathbf{W}_O \in \mathbb{R}^{N\times d}. \nonumber
\end{align}
\endgroup
We then build up a Transformer-like model by interleaving multi-head attention layers with gated-FFN (\textit{e.g.} the SwiGLU~\cite{touvron2023llamaopenefficientfoundation}). Concretely, given layer $l$’s contextualized representation $\mathbf{X}^{(l)}$, we obtain $\mathbf{X}^{(l+1)}$ via,
\begin{align}
    & \mathbf{Y}^{(l)} = \texttt{AttnLayer}(\texttt{norm}(\mathbf{X}^{(l)})) + \mathbf{X}^{(l)}, \nonumber\\
    & \mathbf{X}^{(l+1)} = \texttt{SwiGLU}(\texttt{norm}(\mathbf{Y}^{(l)})) + \mathbf{Y}^{(l)}. \nonumber
\end{align}
Finally, we note that our architecture can cleanly replace the sliding attention module of NSA and maintain the overall linear efficiency. We integrate the \emph{Token Compression} and \emph{Token Selection} modules into the GatedFWA to facilitate an \emph{extended version} with global context awareness, as shown in Fig.~\ref{fig:overall_framework} (right). We defer a detailed discussion to Appendix~\ref{app:nsa}.

\section{Experiment}

\begin{figure}[t]
\centering
    \includegraphics[width=\linewidth]{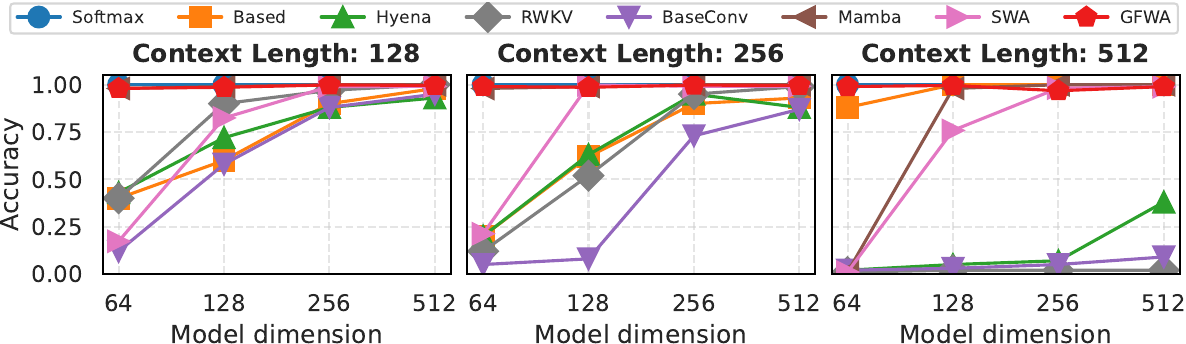}
    \caption{
       Comparison of GatedFWA to Attention and SSM variants on the MQAR benchmark. y-axis is the recall rate.
    } \label{fig:mqar_results}
\end{figure}


\subsection{Recall-Intensive Tasks}
We verify that GatedFWA can \emph{enhance the implicit optimization of associative memory} via Multi-Query Associative Recall (MQAR) \cite{arora2023zoology}: The agent observes a sequence of tokens $\{\mathbf{k}_1, \mathbf{v}_1, \mathbf{k}_2, \mathbf{v}_2, \ldots, \mathbf{k}_r, \mathbf{v}_r\}$, where each consecutive two-tokens become a key-value pair. At test time, the agent is provided with multiple $\mathbf{k} \sim \{\mathbf{k}_1, \dots, \mathbf{k}_r\}$, the goal is to \emph{retrieve} the corresponding values. We consider the sequence length $N \in \{128, 256, 512\}$ and model dimension $d \in \{64, 128, 256, 512\}$. We set $w=N/2$ so all $2$-layer models can capture global context. We compare GatedFWA against Transformer baselines (\emph{e.g.} Softmax, SWA) and various State Space Models (SSMs) \cite{poli2023hyena,peng2023rwkv,gu2023mamba,arora2023zoology,arora2024simple}. Results are summarized in Fig.~\ref{fig:mqar_results}. We observe GatedFWA outperforms existing SSM variants even at $N=512$ and a small $d=64$ whereas the SWA fail at $N=128$ and $d\le 128$. 

\begin{figure}[t]
  \begin{minipage}[c]{0.45\linewidth}
    \includegraphics[width=\textwidth]{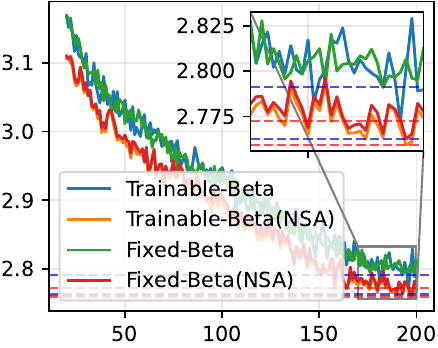}
  \end{minipage}\hfill
  \begin{minipage}[c]{0.5\linewidth}
    \caption{
       Ablation study of the efficacy of learnable amplitude parameter $\boldsymbol{\beta}$ in Eq.~{(\ref{eq:preactivation_and_amplitude})}. For comparison, we set \textcolor{blue}{$\boldsymbol{\beta}=1$ fixed} (with \textbf{best} losses in \textcolor{blue}{dashed-blue}) and \textcolor{red}{learnable} (with \textbf{best} losses in \textcolor{red}{dashed-red}) on both GatedFWA and GatedFWA-NSA variants.
    } \label{fig:with_without_learnable_beta}
  \end{minipage}
  \vspace{-5pt}
\end{figure}

\begin{table*}[!t]
\centering
\vspace{-10pt}
\caption{Models are trained on the subset of SlimPajama dataset with Mistral tokenizer. The model size is $\sim360$M trained for $15$B tokens.}
\resizebox{\linewidth}{!}{\begin{NiceTabular}{@{}c|l|c|c|cccccccccc@{}}
\toprule
\multicolumn{1}{c|}{}                                                               & \multicolumn{1}{l|}{}                                                          & \multicolumn{1}{c|}{}                                 &                                   & \textbf{PIQA}                          & \textbf{Hella}                         & \textbf{Wino}                          & \textbf{ARC-e}                      & \textbf{ARC-c}                      & \textbf{COPA}                          & \textbf{OBQA}                          & \textbf{SciQA}                         & \textbf{BoolQ}                         &                                        \\ \cmidrule(lr){5-13}
\multicolumn{1}{c|}{\multirow{-2}{*}{\textbf{Class}}}                               & \multicolumn{1}{l|}{\multirow{-2}{*}{\textbf{Architecture}}}                   & \multicolumn{1}{c|}{\multirow{-2}{*}{\textbf{Impl.}}} & \multirow{-2}{*}{\textbf{Linear}} & acc $\uparrow$                         & acc\_norm $\uparrow$                   & acc $\uparrow$                         & acc $\uparrow$                      & acc\_norm $\uparrow$                & acc $\uparrow$                         & acc\_norm $\uparrow$                   & acc\_norm $\uparrow$                   & acc $\uparrow$                         & \multirow{-2}{*}{\textbf{Avg.}}        \\ \midrule
\multicolumn{1}{c|}{}                                                               & \multicolumn{1}{l|}{\cellcolor[HTML]{FFFFFF}GLA}                               & \multicolumn{1}{c|}{\cellcolor[HTML]{FFFFFF}Triton}   & \cellcolor[HTML]{FFFFFF}\cmark    & \cellcolor[HTML]{FFFFFF}64.80          & \cellcolor[HTML]{FFFFFF}34.50          & \cellcolor[HTML]{FFFFFF}51.40          & \cellcolor[HTML]{FFFFFF}45.10       & \cellcolor[HTML]{FFFFFF}22.70       & \cellcolor[HTML]{FFFFFF}\textbf{70.00} & \cellcolor[HTML]{FFFFFF}29.20          & \cellcolor[HTML]{FFFFFF}73.20          & \cellcolor[HTML]{FFFFFF}58.70          & \cellcolor[HTML]{FFFFFF}49.95          \\
\multicolumn{1}{c|}{}                                                               & \multicolumn{1}{l|}{\cellcolor[HTML]{FFFFFF}Mamba}                             & \multicolumn{1}{c|}{\cellcolor[HTML]{FFFFFF}CUDA}     & \cellcolor[HTML]{FFFFFF}\cmark    & \cellcolor[HTML]{FFFFFF}\textbf{65.00} & \cellcolor[HTML]{FFFFFF}\textbf{35.40} & \cellcolor[HTML]{FFFFFF}50.10          & \cellcolor[HTML]{FFFFFF}46.30       & \cellcolor[HTML]{FFFFFF}23.60       & \cellcolor[HTML]{FFFFFF}{\ul 69.00}    & \cellcolor[HTML]{FFFFFF}28.00          & \cellcolor[HTML]{FFFFFF}73.70          & \cellcolor[HTML]{FFFFFF}52.60          & \cellcolor[HTML]{FFFFFF}49.30          \\
\multicolumn{1}{c|}{}                                                               & \multicolumn{1}{l|}{\cellcolor[HTML]{FFFFFF}RetNet}                            & \multicolumn{1}{c|}{\cellcolor[HTML]{FFFFFF}CUDA}     & \cellcolor[HTML]{FFFFFF}\cmark    & \cellcolor[HTML]{FFFFFF}63.50          & \cellcolor[HTML]{FFFFFF}33.50          & \cellcolor[HTML]{FFFFFF}\textbf{52.50} & \cellcolor[HTML]{FFFFFF}44.50       & \cellcolor[HTML]{FFFFFF}23.40       & \cellcolor[HTML]{FFFFFF}63.00          & \cellcolor[HTML]{FFFFFF}28.40          & \cellcolor[HTML]{FFFFFF}73.10          & \cellcolor[HTML]{FFFFFF}60.00          & \cellcolor[HTML]{FFFFFF}49.10          \\
\multicolumn{1}{c|}{}                                                               & \multicolumn{1}{l|}{\cellcolor[HTML]{FFFFFF}HGRN2}                             & \multicolumn{1}{c|}{\cellcolor[HTML]{FFFFFF}Triton}   & \cellcolor[HTML]{FFFFFF}\cmark    & \cellcolor[HTML]{FFFFFF}63.49          & \cellcolor[HTML]{FFFFFF}34.94          & \cellcolor[HTML]{FFFFFF}{\ul 51.78}    & \cellcolor[HTML]{FFFFFF}\textbf{50.13}       & \cellcolor[HTML]{FFFFFF}{\ul 25.51}       & \cellcolor[HTML]{FFFFFF}66.00          & \cellcolor[HTML]{FFFFFF}30.00          & \cellcolor[HTML]{FFFFFF}{\ul 75.60}          & \cellcolor[HTML]{FFFFFF}58.41          & \cellcolor[HTML]{FFFFFF}{\ul 50.65}          \\
\multicolumn{1}{c|}{\multirow{-5}{*}{\textbf{\rotatebox[origin=c]{90}{RNN-Like}}}}  & \multicolumn{1}{l|}{\cellcolor[HTML]{FFFFFF}DeltaNet}                          & \multicolumn{1}{c|}{\cellcolor[HTML]{FFFFFF}Triton}   & \cellcolor[HTML]{FFFFFF}\cmark    & \cellcolor[HTML]{FFFFFF}62.73          & \cellcolor[HTML]{FFFFFF}33.28          & \cellcolor[HTML]{FFFFFF}50.28          & \cellcolor[HTML]{FFFFFF}47.39       & \cellcolor[HTML]{FFFFFF}24.32       & \cellcolor[HTML]{FFFFFF}\textbf{70.00} & \cellcolor[HTML]{FFFFFF}29.00          & \cellcolor[HTML]{FFFFFF}74.30          & \cellcolor[HTML]{FFFFFF}54.37          & \cellcolor[HTML]{FFFFFF}49.51          \\ \midrule
\multicolumn{1}{c|}{}                                                               & \multicolumn{1}{l|}{\cellcolor[HTML]{EFEFEF}Transformer (LLaMA)}               & \multicolumn{1}{c|}{\cellcolor[HTML]{EFEFEF}CUDA}     & \cellcolor[HTML]{EFEFEF}\xmark    & \cellcolor[HTML]{EFEFEF}63.22          & \cellcolor[HTML]{EFEFEF}34.20          & \cellcolor[HTML]{EFEFEF}49.49          & \cellcolor[HTML]{EFEFEF}45.98       & \cellcolor[HTML]{EFEFEF}24.49       & \cellcolor[HTML]{EFEFEF}66.00          & \cellcolor[HTML]{EFEFEF}29.40          & \cellcolor[HTML]{EFEFEF}73.90          & \cellcolor[HTML]{EFEFEF}60.09          & \cellcolor[HTML]{EFEFEF}49.96          \\
\multicolumn{1}{c|}{}                                                               & \multicolumn{1}{l|}{\cellcolor[HTML]{EFEFEF}\textit{$\quad$ + SWA}}            & \multicolumn{1}{c|}{\cellcolor[HTML]{EFEFEF}CUDA}     & \cellcolor[HTML]{EFEFEF}\cmark    & \cellcolor[HTML]{EFEFEF}63.10          & \cellcolor[HTML]{EFEFEF}34.10          & \cellcolor[HTML]{EFEFEF}49.44          & \cellcolor[HTML]{EFEFEF}45.60       & \cellcolor[HTML]{EFEFEF}24.40       & \cellcolor[HTML]{EFEFEF}65.93          & \cellcolor[HTML]{EFEFEF}29.22          & \cellcolor[HTML]{EFEFEF}73.79          & \cellcolor[HTML]{EFEFEF}59.96          & \cellcolor[HTML]{EFEFEF}49.50          \\
\multicolumn{1}{c|}{}                                                               & \multicolumn{1}{l|}{\cellcolor[HTML]{EFEFEF}\textit{$\quad$ + SWA + NSA}}      & \multicolumn{1}{c|}{\cellcolor[HTML]{EFEFEF}Triton}   & \cellcolor[HTML]{EFEFEF}\cmark    & \cellcolor[HTML]{EFEFEF}63.97          & \cellcolor[HTML]{EFEFEF}34.70          & \cellcolor[HTML]{EFEFEF}49.92          & \cellcolor[HTML]{EFEFEF}46.24       & \cellcolor[HTML]{EFEFEF}24.93       & \cellcolor[HTML]{EFEFEF}66.92          & \cellcolor[HTML]{EFEFEF}{\ul 30.18}    & \cellcolor[HTML]{EFEFEF}74.75          & \cellcolor[HTML]{EFEFEF}{\ul 60.96}    & \cellcolor[HTML]{EFEFEF}{50.29}    \\ \cmidrule(l){2-14} 
\multicolumn{1}{c|}{}                                                               & \multicolumn{1}{l|}{\cellcolor[HTML]{DFDFDF}\textit{$\quad$ + GatedFWA}}       & \multicolumn{1}{c|}{\cellcolor[HTML]{DFDFDF}Triton}   & \cellcolor[HTML]{DFDFDF}\cmark    & \cellcolor[HTML]{DFDFDF}64.05          & \cellcolor[HTML]{DFDFDF}34.64          & \cellcolor[HTML]{DFDFDF}50.28          & \cellcolor[HTML]{DFDFDF}46.15       & \cellcolor[HTML]{DFDFDF}25.14       & \cellcolor[HTML]{DFDFDF}66.40          & \cellcolor[HTML]{DFDFDF}29.93          & \cellcolor[HTML]{DFDFDF}{74.86}    & \cellcolor[HTML]{DFDFDF}60.58          & \cellcolor[HTML]{DFDFDF}50.23          \\
\multicolumn{1}{c|}{\multirow{-5}{*}{\textbf{\rotatebox[origin=c]{90}{Attention}}}} & \multicolumn{1}{l|}{\cellcolor[HTML]{DFDFDF}\textit{$\quad$ + GatedFWA + NSA}} & \multicolumn{1}{c|}{\cellcolor[HTML]{DFDFDF}Triton}   & \cellcolor[HTML]{DFDFDF}\cmark    & \cellcolor[HTML]{DFDFDF}{\ul 64.86}    & \cellcolor[HTML]{DFDFDF}{\ul 35.10}    & \cellcolor[HTML]{DFDFDF}50.77          & \cellcolor[HTML]{DFDFDF}{\ul 47.20} & \cellcolor[HTML]{DFDFDF}\textbf{25.52} & \cellcolor[HTML]{DFDFDF}67.20          & \cellcolor[HTML]{DFDFDF}\textbf{30.80} & \cellcolor[HTML]{DFDFDF}\textbf{76.20} & \cellcolor[HTML]{DFDFDF}\textbf{61.40} & \cellcolor[HTML]{DFDFDF}\textbf{51.01} \\ \midrule
\multicolumn{4}{c|}{Rel. Improv. to Transformer (LLaMA)}                                                                                                                                                                                                         & \cellcolor[HTML]{F2FFF2}2.59\%         & \cellcolor[HTML]{F2FFF2}2.63\%         & \cellcolor[HTML]{F2FFF2}2.58\%         & \cellcolor[HTML]{F2FFF2}2.65\%      & \cellcolor[HTML]{F2FFF2}4.20\%      & \cellcolor[HTML]{F2FFF2}1.83\%         & \cellcolor[HTML]{F2FFF2}4.76\%         & \cellcolor[HTML]{F2FFF2}3.11\%         & \cellcolor[HTML]{F2FFF2}2.18\%         & \cellcolor[HTML]{F2FFF2}2.49\%         \\ \bottomrule
\end{NiceTabular}}
\label{tab:340m_pretrain}
\end{table*}

\begin{figure*}
\vspace{-5pt}
  \begin{minipage}[c]{0.64\textwidth}
    \includegraphics[width=\textwidth]{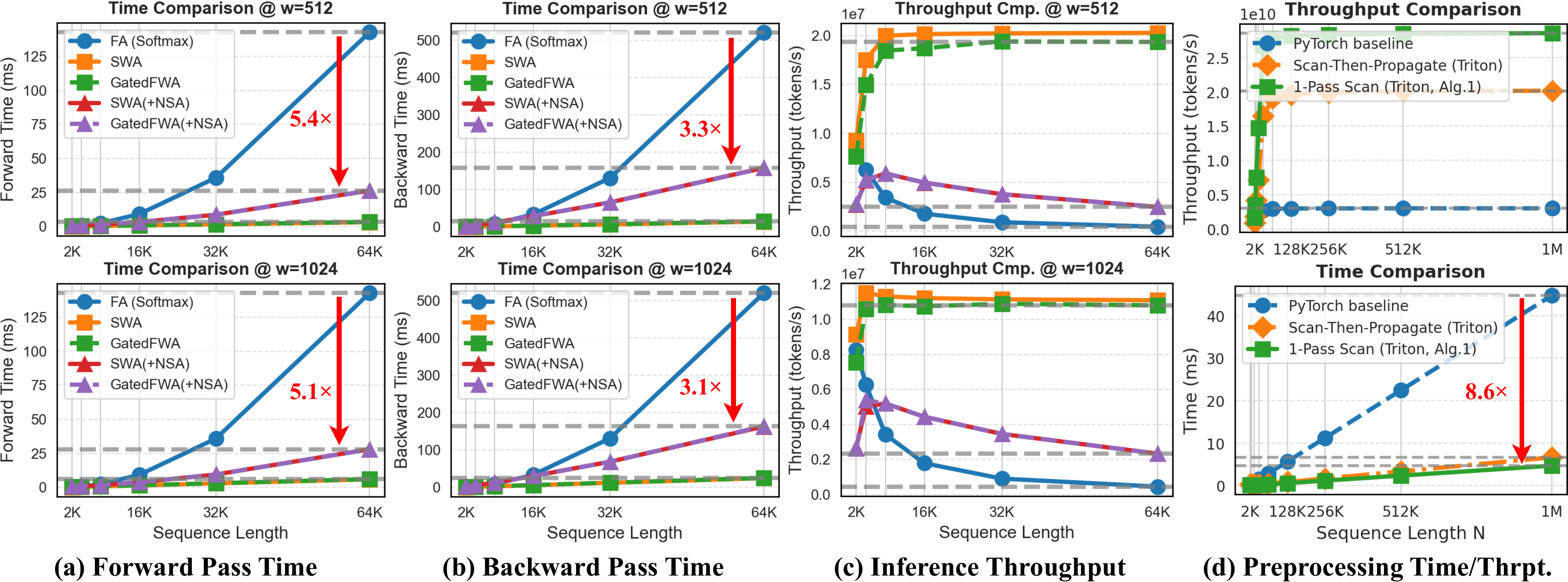}
  \end{minipage}\hfill
  \begin{minipage}[c]{0.34\textwidth}
    \caption{
       \textbf{(a-c)} Time efficiency benchmark for kernels: FA (Softmax FlashAttention) \textit{vs.} SWA (w/ or w/o NSA) \textit{vs.} GatedFWA (w/ or w/o NSA). We evaluate both forward and backward pass with $w=\{512, 1024\}$. We implement NSA compression and selection the same as in \texttt{flash-linear-attention} repository. \textbf{(d)} Benchmark of the \textit{preprocessing algorithms}, compared with the PyTorch baseline and Scan-Then-Propagate implementation (Appendix~\ref{app:hierarchical-scan}), our fused tiled scan kernel (green) achieved negligible computation overhead.
    } \label{fig:time_efficiency_cmp}
  \end{minipage}
  \vspace{-5pt}
\end{figure*}

\begin{figure*}[t]
    \centering
    \includegraphics[width=\linewidth]{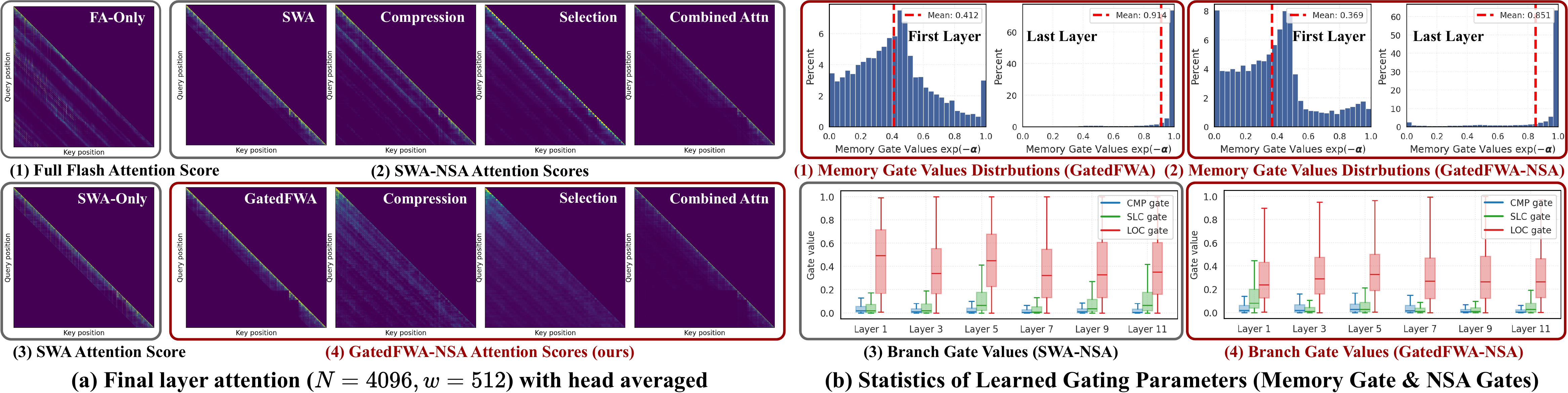}
    \caption{\textbf{Qualitative analysis.} \textbf{(a)} GatedFWA-NSA produces a structurally continuous attention distribution, avoiding the disjointed striding artifacts seen in SWA-NSA. Unlike SWA's unbounded update, GatedFWA applies a learnable contraction $\mathbf{M}_t \leftarrow e^{-\boldsymbol{\alpha}_t} \mathbf{M}_{t-1}$ that selectively down-weights irrelevant history ($\boldsymbol{\alpha}_t \gg 0$) to smooth boundary transitions. \textbf{(b)} Distribution of gate values shows that NSA encourages a stronger memory gating effect, with GatedFWA up-weighting the importance of compression and selection gates.}
    \label{fig:gatedfwa_qualitative_analysis}
    \vspace{-10pt}
\end{figure*}

\subsection{Language Modelling and Scaling Law}

In this section, we consider language modelling tasks on models with $120$M or $360$M parameters with $1024$ to $4096$ context length. Firstly, we begin with the WikiText103 \cite{merity2016pointer} and OpenWebText \cite{Gokaslan2019OpenWeb} dataset as they serve as practically accessible benchmarks for swift evaluation. The details about the parameters/statistics are provided in Appendix~{\ref{app:parameter_statistics}}. We consider the following baseline models: LLaMA \cite{touvron2023llamaopenefficientfoundation}, RetNet \cite{sun2023retentive}, Mamba \cite{gu2023mamba}, RWKV \cite{peng2023rwkv}. Results are summarized in Tab.~\ref{tab:scaling_law} and Fig.~\ref{fig:train_val_curves}. From the figure and table, we can see that GatedFWA variants consistently outperforms baseline Transformers/SSMs up to $360$M and $4096$ context length.

Beyond WikiText103 and OpenWebText, we consider a wide range of downstream tasks covering common-sense reasoning and question-answering as was used in \cite{gu2023mamba}: PiQA \cite{bisk2020piqa}, HellaSwag \cite{zellers2019hellaswag}, WinoGrande \cite{sakaguchi2021winogrande}, ARC-easy (ARC-e) and ARC-challenge (ARC-c) \cite{clark2018think}, Copa \cite{roemmele2011choice}, OpenbookQA \cite{mihaylov2018can}, SciQA \cite{auer2023sciqa}, BoolQA \cite{clark2019boolq}. We report accuracy normalized by length on HellaSwag, ARC-challenge and OpenbookQA,
and accuracy on the other tasks. All evaluations are performed using the LM evaluation harness \cite{eval-harness}. The results are shown in Tab.~\ref{tab:340m_pretrain}. Compared to Transformer architecture without memory gate, the GatedFWA Transformers (and the NSA extended architecture) shows improved results on all tasks, and consistently outperform the RNN-Like models (\emph{e.g.} Linear Transformers/SSMs) with limited memory capacity.

\vspace{-10pt}
\section{Computation Efficiency}

We benchmark the performance of forward and backward pass of the attention kernel and preprocessing kernel. By default, we execute all baselines with $w\in\{512\text{(NSA default)}, 1024\}$, $H = 64$, $d=1024$. All attention have causal mode enabled. We scale $N$ up to $64$K length with a single 80GB A100 GPU. We demonstrate the forward and backward time and forward throughput in Fig.~\ref{fig:time_efficiency_cmp}. The full Softmax-FlashAttention, despite its high I/O efficiency, still scales quadratically on long sequence ($>64$K), which is incapable of very long sequence modelling. The SWA and GatedFWA perform very similarly, achieving $\sim 30\times$ of forward/backward efficiency on sequence length $N\ge64$K compared to the FA counterpart due to their receptive field constraints. The NSA compression and selection kernel expands the theoretical receptive field: adding NSA techniques (\textit{i.e.} Fig.~\ref{fig:overall_framework} (right)) maintains the linear complexity nature, while still outperform the FA (roughly $5\times$ less computation time on $N=64K$). For the preprocessing step that is unique to our GatedFWA, under $N=64$K, our 1-pass fused kernel achieved $0.3$ms forward time compared to the PyTorch baseline $2.9$ms. $0.3$ms is negligible compared to the $6.1$ms forward time of GatedFWA kernel.


\vspace{-10pt}
\section{Conclusion}
We introduced \emph{GatedFWA}, a linear-time attention mechanism that employs learnable memory gates to stabilize memory recurrence, effectively resolving the gradient vanishing and gradient instability of Softmax and SWA memory update. Implemented via I/O-efficient fused kernels, GatedFWA (integrates seamlessly with token compression and selection) achieves competitive performance and throughput among both linear \& quadratic baselines on sequence modelling benchmarks. 

\clearpage
\section*{Limitation and Future Work}
As we will detail in Appendix~\ref{app:circuit_complexity}, GatedFWA is limited to $\mathsf{TC}^{0}$ circuit complexity, restricted to parallelizable updates similar to standard Transformers. Future research will explore \emph{read-write} memory mechanisms, such as the \emph{Delta Rule}, to elevate expressivity to the $\mathsf{NC}^{1}$ class, enabling the modelling of complex, non-commutative state transitions.

\section*{Acknowledgement}
We thank Hermann L. F. He for valuable comments and suggestions on the first version of the manuscript. 
This research was partially funded by Horizon Europe Programme (Grant 101178362), project HAMLET. 
We thank the GPU cluster support from Computer Lab of Paris 6 (Lip6), Sorbonne University, GENCI–IDRIS (Grant 2025-AD011014447), and TACPS Lab, the University of Liverpool. 


\bibliographystyle{named}
\bibliography{ijcai25}

@article{choromanski2020rethinking,
  title={Rethinking attention with performers},
  author={Choromanski, Krzysztof and Likhosherstov, Valerii and Dohan, David and Song, Xingyou and Gane, Andreea and Sarlos, Tamas and Hawkins, Peter and Davis, Jared and Mohiuddin, Afroz and Kaiser, Lukasz and others},
  journal={arXiv preprint arXiv:2009.14794},
  year={2020}
}

@article{child2019generating,
  title={Generating long sequences with sparse transformers},
  author={Child, Rewon and Gray, Scott and Radford, Alec and Sutskever, Ilya},
  journal={arXiv preprint arXiv:1904.10509},
  year={2019}
}

@article{vaswani2017attention,
  title={Attention is all you need},
  author={Vaswani, Ashish and Shazeer, Noam and Parmar, Niki and Uszkoreit, Jakob and Jones, Llion and Gomez, Aidan N and Kaiser, {\L}ukasz and Polosukhin, Illia},
  journal={Advances in neural information processing systems},
  volume={30},
  year={2017}
}

@misc{Gokaslan2019OpenWeb,
    title={OpenWebText Corpus},
    author={Gokaslan, Aaron and Cohen, Vanya and Pavlick, Ellie and Tellex, Stefanie},
    howpublished={\url{http://Skylion007.github.io/OpenWebTextCorpus}},
    year={2019}
}

@misc{touvron2023llamaopenefficientfoundation,
      title={LLaMA: Open and Efficient Foundation Language Models}, 
      author={Hugo Touvron and Thibaut Lavril and Gautier Izacard and Xavier Martinet and Marie-Anne Lachaux and Timothée Lacroix and Baptiste Rozière and Naman Goyal and Eric Hambro and Faisal Azhar and Aurelien Rodriguez and Armand Joulin and Edouard Grave and Guillaume Lample},
      year={2023},
      eprint={2302.13971},
      archivePrefix={arXiv},
      primaryClass={cs.CL},
      url={https://arxiv.org/abs/2302.13971}, 
}

@article{zaheer2020big,
  title={Big bird: Transformers for longer sequences},
  author={Zaheer, Manzil and Guruganesh, Guru and Dubey, Kumar Avinava and Ainslie, Joshua and Alberti, Chris and Ontanon, Santiago and Pham, Philip and Ravula, Anirudh and Wang, Qifan and Yang, Li and others},
  journal={Advances in neural information processing systems},
  volume={33},
  pages={17283--17297},
  year={2020}
}

@article{beltagy2020longformer,
  title={Longformer: The long-document transformer},
  author={Beltagy, Iz and Peters, Matthew E and Cohan, Arman},
  journal={arXiv preprint arXiv:2004.05150},
  year={2020}
}

@article{ainslie2020etc,
  title={ETC: Encoding long and structured inputs in transformers},
  author={Ainslie, Joshua and Ontanon, Santiago and Alberti, Chris and Cvicek, Vaclav and Fisher, Zachary and Pham, Philip and Ravula, Anirudh and Sanghai, Sumit and Wang, Qifan and Yang, Li},
  journal={arXiv preprint arXiv:2004.08483},
  year={2020}
}

@article{kitaev2020reformer,
  title={Reformer: The efficient transformer},
  author={Kitaev, Nikita and Kaiser, {\L}ukasz and Levskaya, Anselm},
  journal={arXiv preprint arXiv:2001.04451},
  year={2020}
}

@inproceedings{katharopoulos2020transformers,
  title={Transformers are rnns: Fast autoregressive transformers with linear attention},
  author={Katharopoulos, Angelos and Vyas, Apoorv and Pappas, Nikolaos and Fleuret, Fran{\c{c}}ois},
  booktitle={International conference on machine learning},
  pages={5156--5165},
  year={2020},
  organization={PMLR}
}

@article{gu2021efficiently,
  title={Efficiently modeling long sequences with structured state spaces},
  author={Gu, Albert and Goel, Karan and R{\'e}, Christopher},
  journal={arXiv preprint arXiv:2111.00396},
  year={2021}
}

@article{ho2019axial,
  title={Axial attention in multidimensional transformers},
  author={Ho, Jonathan and Kalchbrenner, Nal and Weissenborn, Dirk and Salimans, Tim},
  journal={arXiv preprint arXiv:1912.12180},
  year={2019}
}

@article{roy2021efficient,
  title={Efficient content-based sparse attention with routing transformers},
  author={Roy, Aurko and Saffar, Mohammad and Vaswani, Ashish and Grangier, David},
  journal={Transactions of the Association for Computational Linguistics},
  volume={9},
  pages={53--68},
  year={2021},
  publisher={MIT Press One Rogers Street, Cambridge, MA 02142-1209, USA journals-info~…}
}

@article{yuan2025native,
  title={Native Sparse Attention: Hardware-Aligned and Natively Trainable Sparse Attention},
  author={Yuan, Jingyang and Gao, Huazuo and Dai, Damai and Luo, Junyu and Zhao, Liang and Zhang, Zhengyan and Xie, Zhenda and Wei, YX and Wang, Lean and Xiao, Zhiping and others},
  journal={arXiv preprint arXiv:2502.11089},
  year={2025}
}

@article{merity2016pointer,
  title={Pointer sentinel mixture models},
  author={Merity, Stephen and Xiong, Caiming and Bradbury, James and Socher, Richard},
  journal={arXiv preprint arXiv:1609.07843},
  year={2016}
}

@article{gu2023mamba,
  title={Mamba: Linear-time sequence modeling with selective state spaces},
  author={Gu, Albert and Dao, Tri},
  journal={arXiv preprint arXiv:2312.00752},
  year={2023}
}

@inproceedings{poli2023hyena,
  title={Hyena hierarchy: Towards larger convolutional language models},
  author={Poli, Michael and Massaroli, Stefano and Nguyen, Eric and Fu, Daniel Y and Dao, Tri and Baccus, Stephen and Bengio, Yoshua and Ermon, Stefano and R{\'e}, Christopher},
  booktitle={International Conference on Machine Learning},
  pages={28043--28078},
  year={2023},
  organization={PMLR}
}

@article{dao2022flashattention,
  title={Flashattention: Fast and memory-efficient exact attention with io-awareness},
  author={Dao, Tri and Fu, Dan and Ermon, Stefano and Rudra, Atri and R{\'e}, Christopher},
  journal={Advances in neural information processing systems},
  volume={35},
  pages={16344--16359},
  year={2022}
}

@article{agarwal2025gpt,
  title={gpt-oss-120b \& gpt-oss-20b model card},
  author={Agarwal, Sandhini and Ahmad, Lama and Ai, Jason and Altman, Sam and Applebaum, Andy and Arbus, Edwin and Arora, Rahul K and Bai, Yu and Baker, Bowen and Bao, Haiming and others},
  journal={arXiv preprint arXiv:2508.10925},
  year={2025}
}

@article{minaee2024large,
  title={Large language models: A survey},
  author={Minaee, Shervin and Mikolov, Tomas and Nikzad, Narjes and Chenaghlu, Meysam and Socher, Richard and Amatriain, Xavier and Gao, Jianfeng},
  journal={arXiv preprint arXiv:2402.06196},
  year={2024}
}

@article{chu2024qwen2,
  title={Qwen2-audio technical report},
  author={Chu, Yunfei and Xu, Jin and Yang, Qian and Wei, Haojie and Wei, Xipin and Guo, Zhifang and Leng, Yichong and Lv, Yuanjun and He, Jinzheng and Lin, Junyang and others},
  journal={arXiv preprint arXiv:2407.10759},
  year={2024}
}

@article{roziere2023code,
  title={Code llama: Open foundation models for code},
  author={Roziere, Baptiste and Gehring, Jonas and Gloeckle, Fabian and Sootla, Sten and Gat, Itai and Tan, Xiaoqing Ellen and Adi, Yossi and Liu, Jingyu and Sauvestre, Romain and Remez, Tal and others},
  journal={arXiv preprint arXiv:2308.12950},
  year={2023}
}

@article{bordes2024introduction,
  title={An introduction to vision-language modeling},
  author={Bordes, Florian and Pang, Richard Yuanzhe and Ajay, Anurag and Li, Alexander C and Bardes, Adrien and Petryk, Suzanne and Ma{\~n}as, Oscar and Lin, Zhiqiu and Mahmoud, Anas and Jayaraman, Bargav and others},
  journal={arXiv preprint arXiv:2405.17247},
  year={2024}
}

@article{zhong2025survey,
  title={A Survey on Vision-Language-Action Models: An Action Tokenization Perspective},
  author={Zhong, Yifan and Bai, Fengshuo and Cai, Shaofei and Huang, Xuchuan and Chen, Zhang and Zhang, Xiaowei and Wang, Yuanfei and Guo, Shaoyang and Guan, Tianrui and Lui, Ka Nam and others},
  journal={arXiv preprint arXiv:2507.01925},
  year={2025}
}

@article{hopfield1982neural,
  title={Neural networks and physical systems with emergent collective computational abilities.},
  author={Hopfield, John J},
  journal={Proceedings of the national academy of sciences},
  volume={79},
  number={8},
  pages={2554--2558},
  year={1982}
}

@inproceedings{schlag2021linear,
  title={Linear transformers are secretly fast weight programmers},
  author={Schlag, Imanol and Irie, Kazuki and Schmidhuber, J{\"u}rgen},
  booktitle={International conference on machine learning},
  pages={9355--9366},
  year={2021},
  organization={PMLR}
}

@article{arora2023zoology,
  title={Zoology: Measuring and improving recall in efficient language models},
  author={Arora, Simran and Eyuboglu, Sabri and Timalsina, Aman and Johnson, Isys and Poli, Michael and Zou, James and Rudra, Atri and R{\'e}, Christopher},
  journal={arXiv preprint arXiv:2312.04927},
  year={2023}
}

@article{peng2023rwkv,
  title={Rwkv: Reinventing rnns for the transformer era},
  author={Peng, Bo and Alcaide, Eric and Anthony, Quentin and Albalak, Alon and Arcadinho, Samuel and Biderman, Stella and Cao, Huanqi and Cheng, Xin and Chung, Michael and Grella, Matteo and others},
  journal={arXiv preprint arXiv:2305.13048},
  year={2023}
}

@article{arora2024simple,
  title={Simple linear attention language models balance the recall-throughput tradeoff},
  author={Arora, Simran and Eyuboglu, Sabri and Zhang, Michael and Timalsina, Aman and Alberti, Silas and Zinsley, Dylan and Zou, James and Rudra, Atri and R{\'e}, Christopher},
  journal={arXiv preprint arXiv:2402.18668},
  year={2024}
}

@article{sun2023retentive,
  title={Retentive network: A successor to transformer for large language models},
  author={Sun, Yutao and Dong, Li and Huang, Shaohan and Ma, Shuming and Xia, Yuqing and Xue, Jilong and Wang, Jianyong and Wei, Furu},
  journal={arXiv preprint arXiv:2307.08621},
  year={2023}
}

@article{yang2023gated,
  title={Gated linear attention transformers with hardware-efficient training},
  author={Yang, Songlin and Wang, Bailin and Shen, Yikang and Panda, Rameswar and Kim, Yoon},
  journal={arXiv preprint arXiv:2312.06635},
  year={2023}
}

@inproceedings{bisk2020piqa,
  title={Piqa: Reasoning about physical commonsense in natural language},
  author={Bisk, Yonatan and Zellers, Rowan and Gao, Jianfeng and Choi, Yejin and others},
  booktitle={Proceedings of the AAAI conference on artificial intelligence},
  volume={34},
  number={05},
  pages={7432--7439},
  year={2020}
}

@article{zellers2019hellaswag,
  title={Hellaswag: Can a machine really finish your sentence?},
  author={Zellers, Rowan and Holtzman, Ari and Bisk, Yonatan and Farhadi, Ali and Choi, Yejin},
  journal={arXiv preprint arXiv:1905.07830},
  year={2019}
}

@article{sakaguchi2021winogrande,
  title={Winogrande: An adversarial winograd schema challenge at scale},
  author={Sakaguchi, Keisuke and Bras, Ronan Le and Bhagavatula, Chandra and Choi, Yejin},
  journal={Communications of the ACM},
  volume={64},
  number={9},
  pages={99--106},
  year={2021},
  publisher={ACM New York, NY, USA}
}

@article{clark2018think,
  title={Think you have solved question answering? try arc, the ai2 reasoning challenge},
  author={Clark, Peter and Cowhey, Isaac and Etzioni, Oren and Khot, Tushar and Sabharwal, Ashish and Schoenick, Carissa and Tafjord, Oyvind},
  journal={arXiv preprint arXiv:1803.05457},
  year={2018}
}

@inproceedings{roemmele2011choice,
  title={Choice of Plausible Alternatives: An Evaluation of Commonsense Causal Reasoning.},
  author={Roemmele, Melissa and Bejan, Cosmin Adrian and Gordon, Andrew S},
  booktitle={AAAI spring symposium: logical formalizations of commonsense reasoning},
  pages={90--95},
  year={2011}
}

@article{auer2023sciqa,
  title={The sciqa scientific question answering benchmark for scholarly knowledge},
  author={Auer, S{\"o}ren and Barone, Dante AC and Bartz, Cassiano and Cortes, Eduardo G and Jaradeh, Mohamad Yaser and Karras, Oliver and Koubarakis, Manolis and Mouromtsev, Dmitry and Pliukhin, Dmitrii and Radyush, Daniil and others},
  journal={Scientific Reports},
  volume={13},
  number={1},
  pages={7240},
  year={2023},
  publisher={Nature Publishing Group UK London}
}

@article{mihaylov2018can,
  title={Can a suit of armor conduct electricity? a new dataset for open book question answering},
  author={Mihaylov, Todor and Clark, Peter and Khot, Tushar and Sabharwal, Ashish},
  journal={arXiv preprint arXiv:1809.02789},
  year={2018}
}

@misc{eval-harness,
  author       = {Gao, Leo and Tow, Jonathan and Abbasi, Baber and Biderman, Stella and Black, Sid and DiPofi, Anthony and Foster, Charles and Golding, Laurence and Hsu, Jeffrey and Le Noac'h, Alain and Li, Haonan and McDonell, Kyle and Muennighoff, Niklas and Ociepa, Chris and Phang, Jason and Reynolds, Laria and Schoelkopf, Hailey and Skowron, Aviya and Sutawika, Lintang and Tang, Eric and Thite, Anish and Wang, Ben and Wang, Kevin and Zou, Andy},
  title        = {The Language Model Evaluation Harness},
  month        = 07,
  year         = 2024,
  publisher    = {Zenodo},
  version      = {v0.4.3},
  doi          = {10.5281/zenodo.12608602},
  url          = {https://zenodo.org/records/12608602}
}

@article{clark2019boolq,
  title={Boolq: Exploring the surprising difficulty of natural yes/no questions},
  author={Clark, Christopher and Lee, Kenton and Chang, Ming-Wei and Kwiatkowski, Tom and Collins, Michael and Toutanova, Kristina},
  journal={arXiv preprint arXiv:1905.10044},
  year={2019}
}

@article{fu2022hungry,
  title={Hungry hungry hippos: Towards language modeling with state space models},
  author={Fu, Daniel Y and Dao, Tri and Saab, Khaled K and Thomas, Armin W and Rudra, Atri and R{\'e}, Christopher},
  journal={arXiv preprint arXiv:2212.14052},
  year={2022}
}

@article{gu2021combining,
  title={Combining recurrent, convolutional, and continuous-time models with linear state space layers},
  author={Gu, Albert and Johnson, Isys and Goel, Karan and Saab, Khaled and Dao, Tri and Rudra, Atri and R{\'e}, Christopher},
  journal={Advances in neural information processing systems},
  volume={34},
  pages={572--585},
  year={2021}
}

@article{li2024snapkv,
  title={Snapkv: Llm knows what you are looking for before generation},
  author={Li, Yuhong and Huang, Yingbing and Yang, Bowen and Venkitesh, Bharat and Locatelli, Acyr and Ye, Hanchen and Cai, Tianle and Lewis, Patrick and Chen, Deming},
  journal={Advances in Neural Information Processing Systems},
  volume={37},
  pages={22947--22970},
  year={2024}
}

@article{rae2019compressive,
  title={Compressive transformers for long-range sequence modelling},
  author={Rae, Jack W and Potapenko, Anna and Jayakumar, Siddhant M and Lillicrap, Timothy P},
  journal={arXiv preprint arXiv:1911.05507},
  year={2019}
}

@article{liu2025toward,
  title={Toward Linearly Regularizing the Geometric Bottleneck of Linear Generalized Attention},
  author={Liu, Jiaxu and Yi, Xinping and Yin, Xiangyu and Song, Yuhang and Jin, Gaojie and Huang, Xiaowei},
  journal={Transactions on Machine Learning Research},
  year={2025}
}

@article{munkhdalai2024leave,
  title={Leave no context behind: Efficient infinite context transformers with infini-attention},
  author={Munkhdalai, Tsendsuren and Faruqui, Manaal and Gopal, Siddharth},
  journal={arXiv preprint arXiv:2404.07143},
  volume={101},
  year={2024}
}

@article{rodkin2024associative,
  title={Associative recurrent memory transformer},
  author={Rodkin, Ivan and Kuratov, Yuri and Bulatov, Aydar and Burtsev, Mikhail},
  journal={arXiv preprint arXiv:2407.04841},
  year={2024}
}

@article{ramsauer2020hopfield,
  title={Hopfield networks is all you need},
  author={Ramsauer, Hubert and Sch{\"a}fl, Bernhard and Lehner, Johannes and Seidl, Philipp and Widrich, Michael and Adler, Thomas and Gruber, Lukas and Holzleitner, Markus and Pavlovi{\'c}, Milena and Sandve, Geir Kjetil and others},
  journal={arXiv preprint arXiv:2008.02217},
  year={2020}
}

@inproceedings{tillet2019triton,
  title={Triton: an intermediate language and compiler for tiled neural network computations},
  author={Tillet, Philippe and Kung, Hsiang-Tsung and Cox, David},
  booktitle={Proceedings of the 3rd ACM SIGPLAN International Workshop on Machine Learning and Programming Languages},
  pages={10--19},
  year={2019}
}

@article{penedo2024fineweb,
  title={The fineweb datasets: Decanting the web for the finest text data at scale},
  author={Penedo, Guilherme and Kydl{\'\i}{\v{c}}ek, Hynek and Lozhkov, Anton and Mitchell, Margaret and Raffel, Colin A and Von Werra, Leandro and Wolf, Thomas and others},
  journal={Advances in Neural Information Processing Systems},
  volume={37},
  pages={30811--30849},
  year={2024}
}

@article{allal2025smollm2,
  title={SmolLM2: When Smol Goes Big--Data-Centric Training of a Small Language Model},
  author={Allal, Loubna Ben and Lozhkov, Anton and Bakouch, Elie and Bl{\'a}zquez, Gabriel Mart{\'\i}n and Penedo, Guilherme and Tunstall, Lewis and Marafioti, Andr{\'e}s and Kydl{\'\i}{\v{c}}ek, Hynek and Lajar{\'\i}n, Agust{\'\i}n Piqueres and Srivastav, Vaibhav and others},
  journal={arXiv preprint arXiv:2502.02737},
  year={2025}
}

\clearpage
\appendix
\onecolumn
\section*{Appendix}
\section{Related Works}
\label{sec:related}

\begin{table*}[h]
\centering
\caption{Comparison of memory recurrence and optimization objective among related associative-memory models.}
\resizebox{0.96\textwidth}{!}{
\begin{NiceTabular}{l|l|l}
\toprule
\textbf{Architecture} & \textbf{Memory Recurrence} & \textbf{Optimization Objective $\mathcal{L}_t(\mathbf{M}_{t-1})$}\\
\midrule
\textbf{Linear Attn} & 
$\mathbf{M}_t = \mathbf{M}_{t-1} + \mathbf{k}_t^{\top}\mathbf{v}_t $ & 
$-\langle \mathbf{M}_{t-1} \mathbf{k}_t, \mathbf{v}_t \rangle$ \\[2pt]

\textbf{Mamba2} &
$\mathbf{M}_t = \lambda_t\mathbf{M}_{t-1} - \beta_t \mathbf{k}_t^\top\mathbf{v}_t$ &
$\frac{1}{2}\|(\sqrt{1-\lambda_t})\mathbf{M}_{t-1}\|_F^2- \beta_t\langle \mathbf{M}_{t-1}\mathbf{k}_t, \mathbf{v}_t\rangle$ \\[2pt]

\textbf{GLA} &
$\mathbf{M}_t = (\boldsymbol{\lambda}_t \mathbf{I}_k)\mathbf{M}_{t-1} + \mathbf{k}_t^\top \mathbf{v}_t $ &
$\frac{1}{2}\|(\sqrt{1-\boldsymbol{\lambda}_t} \mathbf{I}_k)\mathbf{M}_{t-1}\|_F^2-\langle \mathbf{M}_{t-1}\mathbf{k}_t, \mathbf{v}_t\rangle$ \\[2pt]

\textbf{HGRN2} &
$\mathbf{M}_t = (\boldsymbol{\lambda}_t \mathbf{I}_k)\mathbf{M}_{t-1} + (1-\boldsymbol{\lambda}_t)^\top \mathbf{v}_t $ &
$\frac{1}{2}\|(\sqrt{(1-\boldsymbol{\lambda}_t)} \mathbf{I}_k)\mathbf{M}_{t-1}\|_F^2-\langle \mathbf{M}_{t-1}(1-\boldsymbol{\lambda}_t), \mathbf{v}_t\rangle$ \\[2pt]

\textbf{Softmax Attn} &
$\mathbf{M}_t = \mathbf{M}_{t-1} + \frac{1}{w} (\phi(\mathbf{k}_t)^{\top} \mathbf{v}_t  - \phi(\mathbf{k}_{t-w})^{\top} \mathbf{v}_{t-w} )$ &
$\frac{1}{2t}\|\mathbf{M}_{t-1}\|_F^2-\frac{1}{t}\langle \phi(\mathbf{k}_t)\mathbf{M}_{t-1}, \mathbf{v}_t \rangle$ \\[2pt]

\textbf{SWA} &
$\mathbf{M}_t = \sum_{i=t-w+1}^t \mathbf{v}_i \mathbf{k}_i^\top$ &
$\frac{1}{w}\langle\mathbf{M}_{t-1}, \phi(\mathbf{k}_{t-w})^\top\mathbf{v}_{t-w} - \phi(\mathbf{k}_{t})^\top\mathbf{v}_{t}\rangle$ \\[2pt]

\textbf{GatedFWA (This Work)} &
$\mathbf{M}_t = (\boldsymbol{\lambda}_t\mathbf{I}_k)\mathbf{M}_{t-1} + \frac{1}{w} (\phi(\mathbf{k}_t)^\top \mathbf{v}_t 
    - 
    (\boldsymbol{\lambda}_t'\mathbf{I}_k)
    \phi(\mathbf{k}_{t-w})^\top \mathbf{v}_{t-w})$ &
$\frac{1}{2}\| ( \sqrt{1-\boldsymbol{\lambda}_t}  \mathbf{I}_k ) \mathbf{M}_{t-1}\|_F^2 - 
    \frac{1}{w} \langle
    \mathbf{M}_{t-1}
    ,
    \boldsymbol{\lambda}'_t\mathbf{I}_k
    \Delta_t
    + (1-\boldsymbol{\lambda}'_t)\mathbf{I}_k 
    \phi(\mathbf{k}_t)\mathbf{v}_t^\top
    \rangle$ \\
\bottomrule
\end{NiceTabular}
}
\label{tab:memory_update}
\end{table*}

\paragraph{Efficient Long-context Transformers.}
A large body of work reduces the quadratic cost of full Softmax attention in autoregressive models. Sparse-pattern designs constrain attention to local, dilated, axial, or block-sparse structures, often with a small set of global tokens, yielding subquadratic complexity while preserving parallel training \cite{child2019generating,beltagy2020longformer,ainslie2020etc,zaheer2020big,roy2021efficient}.
Sliding-window attention (SWA) is a particularly practical instance used widely in large-scale systems due to its simplicity, strong performance, and compatibility with production kernels \cite{beltagy2020longformer,dao2022flashattention}.
These approaches, however, are largely combinatorial: they do not directly address the stability of the induced memory update or the depth-wise gradient flow in deep stacks issues we target explicitly.

\paragraph{Linear Attention and Gated Variants.}
Linear attention replaces the exponential kernel with a feature map (more details on mapping $\phi$ is elaborated in Appendix~\ref{app:kernel_mapping}), enabling recurrent or prefix-scan style computation in linear time \cite{katharopoulos2020transformers,choromanski2020rethinking,schlag2021linear}.
While elegant and asymptotically appealing, vanilla linear attention often underperforms strong Softmax baselines on language modelling and can suffer from attention dilution and limited effective memory \cite{schlag2021linear}. Conceptually, these works treat attention as a recurrent memory with explicit control, but they operate in the global linear-attention regime, focusing on feature maps and decay schedules. By contrast, GatedFWA retains the industry-standard explicit sliding-window pattern and views gating as a learnable contraction on the SWA recurrence, directly targeting gradient pathologies of both Softmax and SWA while preserving the practical deployment footprint.

\paragraph{State-space Models and Recurrent Alternatives.}
State-Space Models (SSMs) and linear RNN-style architectures capture long-range dependencies with structured state transitions and parallelizable training \cite{gu2021combining,gu2021efficiently,fu2022hungry,gu2023mamba}.
These models trade the flexible content-based addressing of attention for parameterized dynamics, and typically rely on specialized kernels for efficiency.
While competitive on long-context and streaming tasks, capacity is tied to state dimensionality, and integrating them into attention-centric LLM stacks may require substantial architectural changes.

\paragraph{Token Selection, Compression, and Sparse Retrieval.}
Orthogonal to designing better local or linear operators, token selection and compression methods expand effective context by preserving only salient states \cite{rae2019compressive,li2024snapkv,munkhdalai2024leave,liu2025toward}.
Native Sparse Attention (NSA) \cite{yuan2025native} exemplifies this strategy by coupling lightweight token scoring and top-k blockwise selection with SWA to realize natively sparse decoding and efficient long-context pretraining, achieving speedups without heavily depending on hand-crafted fixed patterns. GatedFWA is complementary: it can replace the local SWA component in NSA-style pipelines, improving stability and controllability of the underlying memory update.


\paragraph{Associative Memory and Fast Weights.}
Our analysis is grounded in the view of attention as a fast-weight associative memory in which updates and retrieval define an explicit recurrence \cite{hopfield1982neural,schlag2021linear,rodkin2024associative,ramsauer2020hopfield}.
Classical results connect update rules to memory capacity and stability, we show a brief comparison of associative-memory interpretation of different sequence models in Tab.~\ref{tab:memory_update}. Situating SWA within this framework, we show that its difference-style recurrence induces an effectively unbounded objective, and propose a non-negative gate that yields a bounded, controllable contraction on the carried memory. This positions GatedFWA at the intersection of efficient attention, gated recurrent models, and associative-memory theory.

\section{GatedFWA with NSA Extension (Compression and Selection Branch)}
\label{app:nsa}

\begin{figure*}[t]
    \centering
    \includegraphics[width=0.7\linewidth]{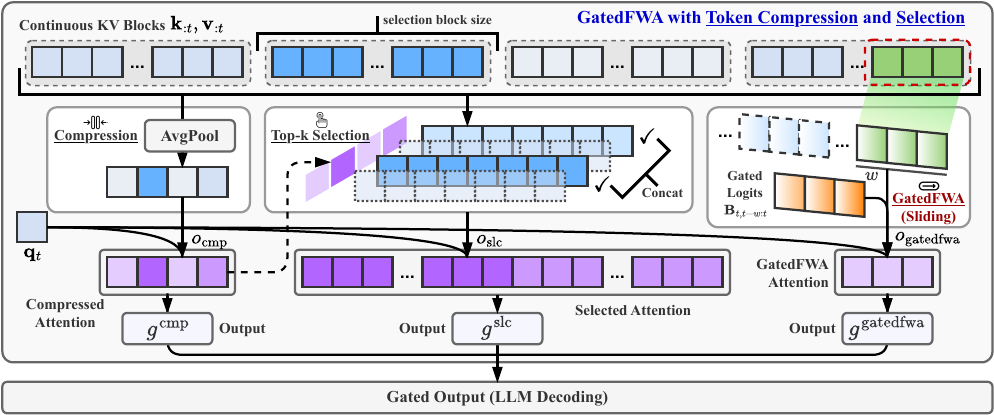}
    \caption{GatedFWA with token compression and selection. For each query $\mathbf{q}_t$, the continuous KV blocks $\mathbf{K}_{1:t}, \mathbf{V}_{1:t}$ are processed by three parallel branches: \textbf{(left)} a compression branch that averages blockwise tokens into compressed KV blocks and produces compressed attention $\mathbf{o}_{\mathrm{cmp}}$ gated by $g_{\mathrm{cmp}}$; \textbf{(middle)} a selection branch that scores blocks, performs top-$k$ selection, concatenates the surviving blocks, and yields selected attention $\mathbf{o}_{\mathrm{slc}}$ with gate $g_{\mathrm{slc}}$; and \textbf{(right)} a local branch where the last $w$ tokens are attended with GatedFWA using precomputed gated logits $\mathbf{B}_{t, t-w+1:t}$, producing $\mathbf{o}_{\mathrm{gatedfwa}}$ and gate $g_{\mathrm{gatedfwa}}$. The three gated outputs are combined into a single gated output as final representation for decoding.}
    \label{fig:gatedfwa_nsa}
\end{figure*}

\paragraph{Brief Review of NSA.}
Native Sparse Attention (NSA)~\cite{yuan2025native} is a learnable sparse attention scheme that constructs, for each query position $t$, a compact, information-dense subset of the KV cache instead of attending to all past tokens. Concretely, NSA maintains three complementary branches operating on different granularities of context:
\emph{\textbf{(i)}} a \underline{\emph{compression}} branch that aggregates nearby tokens into coarse block-level summaries,
\emph{\textbf{(ii)}} a \underline{\emph{selection}} branch that keeps only a small number of high-importance blocks, and
\emph{\textbf{(iii)}} a \underline{\emph{local}} sliding branch that preserves fine-grained recent context.
Given a query $\mathbf{q}_t \in \mathbb{R}^{1 \times d}$ and a history of keys/values $\{\mathbf{K}_{1:t}, \mathbf{V}_{1:t}\}$, NSA forms three branch-specific KV subsets.

Formally, let $\mathcal{C} = \{\mathrm{cmp}, \mathrm{slc}, \mathrm{loc}\}$ index the compression, selection, and local branches. For each branch $c \in \mathcal{C}$, NSA applies learnable remapping functions
\begin{equation}
\tilde{\mathbf{K}}_t^{c} = f_K^{c}(\mathbf{K}_{1:t}), \quad
\tilde{\mathbf{V}}_t^{c} = f_V^{c}(\mathbf{V}_{1:t}),
\end{equation}
where $\tilde{\mathbf{K}}_t^{c} \in \mathbb{R}^{N_t^{c} \times d}$ and $\tilde{\mathbf{V}}_t^{c} \in \mathbb{R}^{N_t^{c} \times d}$ denote the branch-specific key and value subsets, and $N_t^{c} \ll t$ is the number of effective tokens in that branch. The three branches are summarized as:

\begin{align}
\textbf{Compression (cmp)}:\quad
&\tilde{\mathbf{K}}_t^{\mathrm{cmp}} = \{\phi(\mathbf{K}_{i s_{\mathrm{cmp}}+1:i s_{\mathrm{cmp}}+b_{\mathrm{cmp}}}) \mid 0 \le i < \lfloor (t - b_{\mathrm{cmp}})/s_{\mathrm{cmp}} \rfloor\}, \label{eq:nsa-cmp-K}\\
&\tilde{\mathbf{V}}_t^{\mathrm{cmp}} = \{\phi(\mathbf{V}_{i s_{\mathrm{cmp}}+1:i s_{\mathrm{cmp}}+b_{\mathrm{cmp}}}) \mid 0 \le i < \lfloor (t - b_{\mathrm{cmp}})/s_{\mathrm{cmp}} \rfloor\}, \nonumber
\end{align}
where $b_{\mathrm{cmp}}$ is the compression block length, $s_{\mathrm{cmp}}$ is the compression stride, and $\phi$ is a learnable MLP that maps each contiguous block into a single compressed representation.

\begin{align}
\textbf{Selection (slc)}:\quad
& I_t = \{ i \mid \operatorname{rank}(p_t^{\mathrm{slc}'}[i]) \le k_{\mathrm{slc}} \}, \\
&\tilde{\mathbf{K}}_t^{\mathrm{slc}} = \operatorname{Concat}\{\mathbf{K}_{i b_{\mathrm{slc}}+1:(i+1)b_{\mathrm{slc}}} \mid i \in I_t\}, \\
&\tilde{\mathbf{V}}_t^{\mathrm{slc}} = \operatorname{Concat}\{\mathbf{V}_{i b_{\mathrm{slc}}+1:(i+1)b_{\mathrm{slc}}} \mid i \in I_t\},
\end{align}
where $p_t^{\mathrm{slc}'}$ denotes block-level importance scores induced from the compression attention, $k_{\mathrm{slc}}$ is the number of blocks to retain, and $b_{\mathrm{slc}}$ is the selection block size.

\begin{align}
\textbf{Local (loc)}:\quad
&\tilde{\mathbf{K}}_t^{\mathrm{loc}} = \mathbf{K}_{t-w+1:t}, \\
&\tilde{\mathbf{V}}_t^{\mathrm{loc}} = \mathbf{V}_{t-w+1:t},
\end{align}
where $w$ is the local window size and the indices are clipped when $t < w$. For each branch, the standard attention output is
\begin{equation}
\mathbf{o}_t^{c} = \operatorname{AttnKernel}(\mathbf{q}_t, \tilde{\mathbf{K}}_t^{c}, \tilde{\mathbf{V}}_t^{c}), \quad c \in \mathcal{C},
\end{equation}
with $\operatorname{AttnKernel}$ denoting the usual Softmax attention implemented in a fused kernel.

NSA then aggregates these branch outputs using a learnable gate. Let $g_t^{c} \in [0, 1]$ be scalar gate values (typically produced by a small MLP on $\mathbf{q}_t$ or the current hidden state), and $g_t = \{g_t^{\mathrm{cmp}}, g_t^{\mathrm{slc}}, g_t^{\mathrm{loc}}\}$. The final NSA output is
\begin{equation}
\mathbf{o}_t^{\mathrm{NSA}} = \sum_{c \in \mathcal{C}} g_t^{c} \mathbf{o}_t^{c}.
\end{equation}
This yields a hardware-aligned sparse operator in which each query attends only to a compact set of compressed, selected, and local tokens while keeping training fully differentiable.

\paragraph{GatedFWA with NSA Extension.}
The proposed GatedFWA can replace the local sliding branch in NSA, leaving the compression and selection branches unchanged. Let $\mathbf{U}=\{\mathbf{u}_i\}_{i=1}^N \in \mathbb{R}^{N\times H}$ so that $\mathbf{U}_{1:t} \in \mathbb{R}^{t\times H}$ denotes the cumulative gate vector defined in Sec.~3.2 (per head, we maintain a one-dimensional prefix-sum gate along the sequence). For the sliding branch, we restrict this gate to the current window and feed it into GatedFWA:
\begin{equation}
\mathbf{o}_t^{\mathrm{gatedfwa}} = \operatorname{GatedFWAKernel}(\mathbf{q}_t, \tilde{\mathbf{K}}_t^{\mathrm{loc}}, \tilde{\mathbf{V}}_t^{\mathrm{loc}}, \mathbf{U}_{t-w+1:t}),
\end{equation}
where $\operatorname{GatedFWAKernel}(\cdot)$ is our GatedFWA kernel with local window and gated logits introduced in Sec.~3.2, and $\mathbf{U}_{t-w+1:t}$ denotes the slice of the gate aligned with the local window. The compression and selection branches keep their original Softmax attention form,
\begin{equation}
\mathbf{o}_t^{\mathrm{cmp}} = \operatorname{AttnKernel}(\mathbf{q}_t, \tilde{\mathbf{K}}_t^{\mathrm{cmp}}, \tilde{\mathbf{V}}_t^{\mathrm{cmp}}), \quad
\mathbf{o}_t^{\mathrm{slc}} = \operatorname{AttnKernel}(\mathbf{q}_t, \tilde{\mathbf{K}}_t^{\mathrm{slc}}, \tilde{\mathbf{V}}_t^{\mathrm{slc}}),
\end{equation}
and the overall hybrid output becomes
\begin{equation}
\mathbf{o}_t^{\mathrm{GatedFWA}^\star} = g_t^{\mathrm{cmp}} \mathbf{o}_t^{\mathrm{cmp}} + g_t^{\mathrm{slc}} \mathbf{o}_t^{\mathrm{slc}} + g_t^{\mathrm{loc}} \mathbf{o}_t^{\mathrm{gatedfwa}}.
\end{equation}
This design preserves the native sparse structure and blockwise kernels of NSA for compression and selection, while the sliding branch benefits from the gated memory recurrence of GatedFWA. In particular, the local branch now operates with a learnable contraction on its associative memory state (Sec.~3.2), stabilizing the update within each window without changing the asymptotic linear complexity of NSA: the number of attended tokens per query remains $N_t^{\mathrm{cmp}} + N_t^{\mathrm{slc}} + w$, and the underlying implementations reuse NSA’s block-sparse kernels and our linear-time GatedFWA kernel as in Fig.~\ref{fig:gatedfwa_nsa}.

\section{Supplemental Results}
\subsection{Pretraining-Finetuning with \texttt{nanochat}}
\begin{table*}[t]
\centering
\caption{Benchmarking results of standard Transformer, Transformer(SWA) and Transformer(GatedFWA) models on CORE tasks within the \texttt{nanochat} pipeline (Pretrained on FineWeb-Edu dataset).}
\resizebox{1\textwidth}{!}{
\begin{NiceTabular}{@{}l|ccccccccccccc@{}}
\toprule
\textbf{Architecture}                    & \textbf{\begin{tabular}[c]{@{}c@{}}HellaSwag\\ 0-Shot\end{tabular}} & \textbf{\begin{tabular}[c]{@{}c@{}}HellaSwag\\ 10-Shot\end{tabular}} & \textbf{Jeopardy} & \textbf{Winograd} & \textbf{Winogrande} & \textbf{\begin{tabular}[c]{@{}c@{}}Lambada\\ OpenAI\end{tabular}} & \textbf{ARC-c}  & \textbf{ARC-e}  & \textbf{CoQA}   & \textbf{BoolQ}  & \textbf{\begin{tabular}[c]{@{}c@{}}Bigbench\\ Dyck\end{tabular}} & \textbf{\begin{tabular}[c]{@{}c@{}}Bigbench\\ CS\end{tabular}} & \textbf{\begin{tabular}[c]{@{}c@{}}Bigbench\\ Lang ID\end{tabular}} \\ \midrule
\textit{nanochat}          & \textbf{0.4482}                                                     & 0.4498                                                               & 0.0746            & 0.6374            & 0.5233              & 0.3732                                                            & 0.3404          & 0.6549          & \textbf{0.2054} & 0.5636          & 0.1150                                                           & 0.3977                                                         & 0.2564                                                              \\
\textit{nanochat(SWA)}      & 0.4440                                                              & 0.4440                                                               & \textbf{0.1420}   & 0.6374            & 0.5230              & 0.3700                                                            & 0.3680          & 0.6520          & 0.1540          & 0.5460          & 0.1000                                                           & 0.4280                                                         & 0.2540                                                              \\
\textit{nanochat(GatedFWA)} & 0.4440                                                              & \textbf{0.4600}                                                      & 0.1300            & \textbf{0.6447}   & \textbf{0.5360}     & \textbf{0.3880}                                                   & \textbf{0.3700} & \textbf{0.6580} & 0.1940          & \textbf{0.5820} & \textbf{0.1220}                                                  & \textbf{0.4360}                                                & \textbf{0.2820}                                                     \\ \bottomrule
\end{NiceTabular}
}
\label{tab:benchmark_nanochat_core}
\end{table*}
\begin{table*}[t]
\centering
\caption{Benchmarking results of standard Transformer, Transformer(SWA) and Transformer(GatedFWA) models within the \texttt{nanochat} pipeline (Pretrain$\rightarrow$Midtrain$\rightarrow$SFT). Performance is evaluated across ARC-e, ARC-c and MMLU datasets.}
\resizebox{0.7\textwidth}{!}{
\begin{NiceTabular}{@{}l|cc|cc|cc@{}}
\toprule
\multirow{2}{*}{\textbf{Metric}} & \multicolumn{2}{c|}{\textit{\text{nanochat}}} & \multicolumn{2}{c|}{\textit{\text{nanochat(SWA)}}} & \multicolumn{2}{c}{\textit{\text{nanochat(GatedFWA)}}} \\ \cmidrule(l){2-7} 
                                 & \textbf{Mid-Train}               & \textbf{SFT}               & \textbf{Mid-Train}                 & \textbf{SFT}                 & \textbf{Mid-Train}                  & \textbf{SFT}                    \\ \midrule
\textbf{ARC-Challenge (HF, chat-format)}           & 0.2875                           & 0.2807                     & 0.2818                             & 0.2758                       & \textbf{0.3319}                     & 0.3208                          \\
\textbf{ARC-Easy (HF, chat-format)}                & 0.3561                           & 0.3876                     & 0.3697                             & 0.3731                       & 0.4297                              & \textbf{0.4381}                 \\
\textbf{MMLU (HF, chat-format)}                    & 0.2973                           & 0.3081                     & 0.2960                             & 0.3034                       & 0.3183                              & \textbf{0.3249}                 \\ \bottomrule
\end{NiceTabular}
}
\label{tab:benchmark_nanochat_post}
\end{table*}
\begin{wrapfigure}{l}{0.45\linewidth}
  \vspace{-10pt}
  \begin{center}
    \begin{subfigure}[t]{0.498\linewidth}
        \centering
        \includegraphics[width=\textwidth]{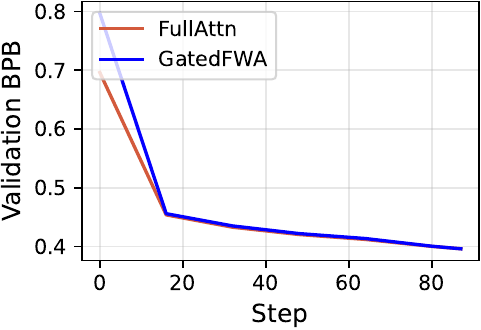}
        \caption{Mid-training Val. BPB}
    \end{subfigure}%
    \hfill
    \begin{subfigure}[t]{0.498\linewidth}
        \centering
        \includegraphics[width=\textwidth]{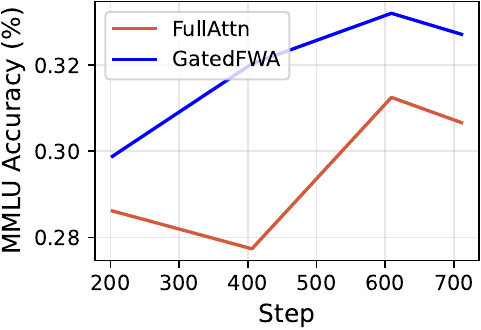}
        \caption{SFT Acc. on MMLU}
    \end{subfigure}
  \end{center}
  \caption{Efficacy of Mid-training and SFT on various attention modes ($N=2048$ and $w=512$).}
  \vspace{-10pt}
  \label{fig:mid_sft}
\end{wrapfigure}In this section, we verify the efficacy of GatedFWA within the \texttt{nanochat}\footnote{https://github.com/karpathy/nanochat} framework by comparing it against standard Transformer and Sliding Window Attention (SWA) baselines. We evaluate all architectures across the full training pipeline (\textbf{Pre-Training}$\rightarrow$\textbf{Mid-Training}$\rightarrow$\textbf{SFT}): \textbf{\textit{(i)}} starting with \textit{Base} pretraining on a large web-text corpus prepared following the \textbf{fineweb-edu-100B} recipe \cite{penedo2024fineweb} to establish fundamental language capabilities; \textbf{\textit{(ii)}} proceeding to \textit{Mid-training} on a task mixture that combines SmolTalk conversational data \cite{allal2025smollm2} with reasoning benchmarks (\textit{e.g.} MMLU), synthetic spelling/counting tasks, and synthetic identity conversations that collectively teach conversation special tokens, tool use, and multiple-choice reasoning; \textbf{\textit{(iii)}} and finally \textit{Supervised Fine-Tuning} (SFT) on a broader instruction/chat mixture including ARC (Easy/Challenge), truncated SmolTalk, and identity-focused conversational data to sharpen interactive chat behaviour. The model and training parameter specifications are detailed in Tab.~\ref{tab:details_params_nanochat}. 

We show the benchmarking results of pre-training and post-training and in Tab.~\ref{tab:benchmark_nanochat_core} and Tab.~\ref{tab:benchmark_nanochat_post}, and the efficacy on mid-training and SFT on various attention modes in Fig.~\ref{fig:mid_sft}. Across all three stages, GatedFWA exhibits particular strength on reasoning-intensive benchmarks, achieving competitive performance among the compared models. Specifically, it records the highest scores on ARC-Challenge (0.3319 at Midtrain) and MMLU (0.3249 at SFT), surpassing the full-attention Transformer baseline which achieved 0.2875 and 0.3081 respectively. Moreover, the transition from Midtrain to SFT consistently improves GatedFWA's performance on ARC-Easy (rising from 0.4297 to 0.4381) and MMLU, highlighting the model's robustness and adaptability during the instruction tuning phase.

\subsection{Attention Patterns}
\begin{figure*}[t]
    \centering
    \begin{subfigure}[t]{0.498\textwidth}
        \centering
        \includegraphics[width=\textwidth]{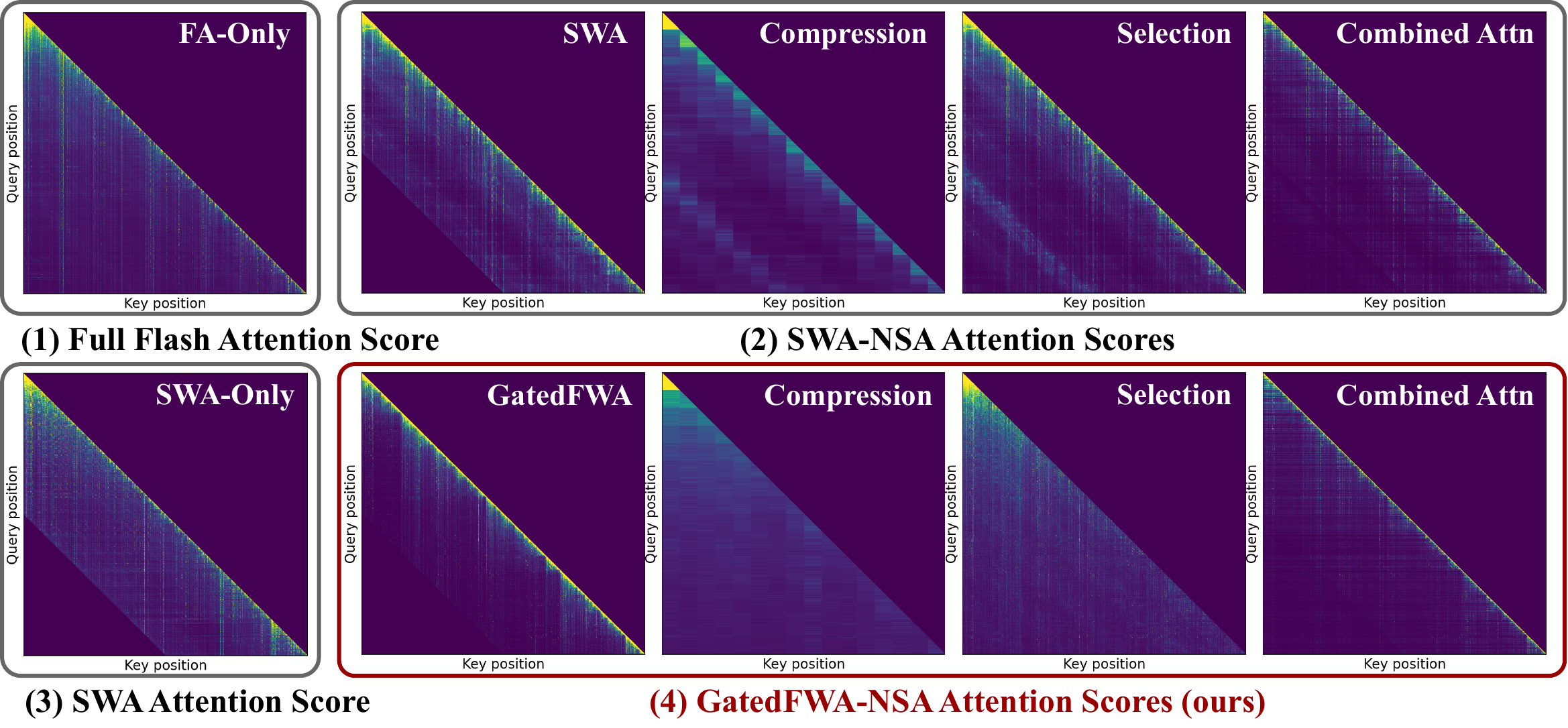}
        \caption{Final layer attention ($N=1024,w=512$) with head averaged}
    \end{subfigure}%
    \hfill
    \begin{subfigure}[t]{0.498\textwidth}
        \centering
        \includegraphics[width=\textwidth]{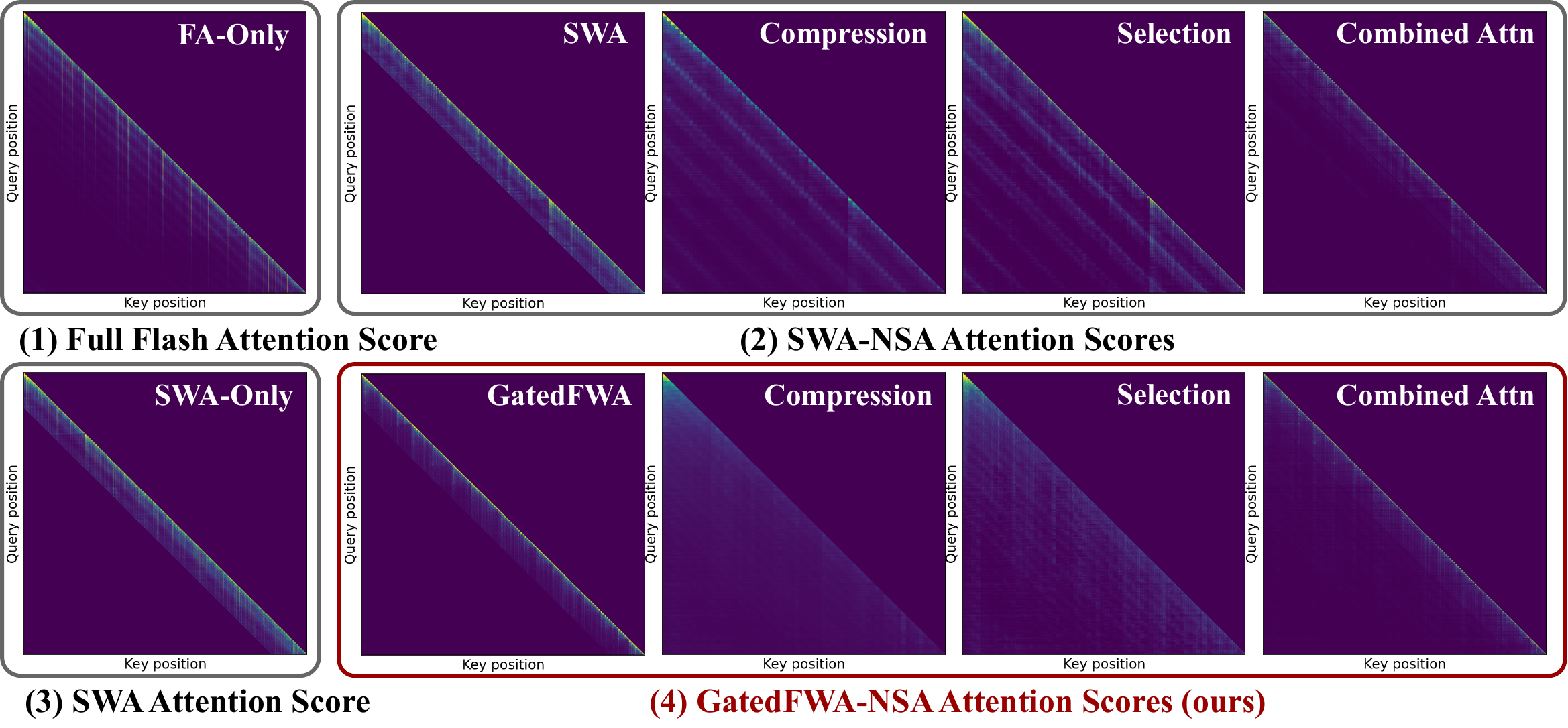}
        \caption{Final layer attention ($N=4096,w=512$) with head averaged}
    \end{subfigure}
    \caption{Attention score visualization (pretrained on WikiText) with various setups. 
    }
    \label{fig:visualization_attention_scores_wiki}
\end{figure*}

\begin{figure*}[t]
    \centering
    \begin{subfigure}[t]{0.498\textwidth}
        \centering
        \includegraphics[width=\textwidth]{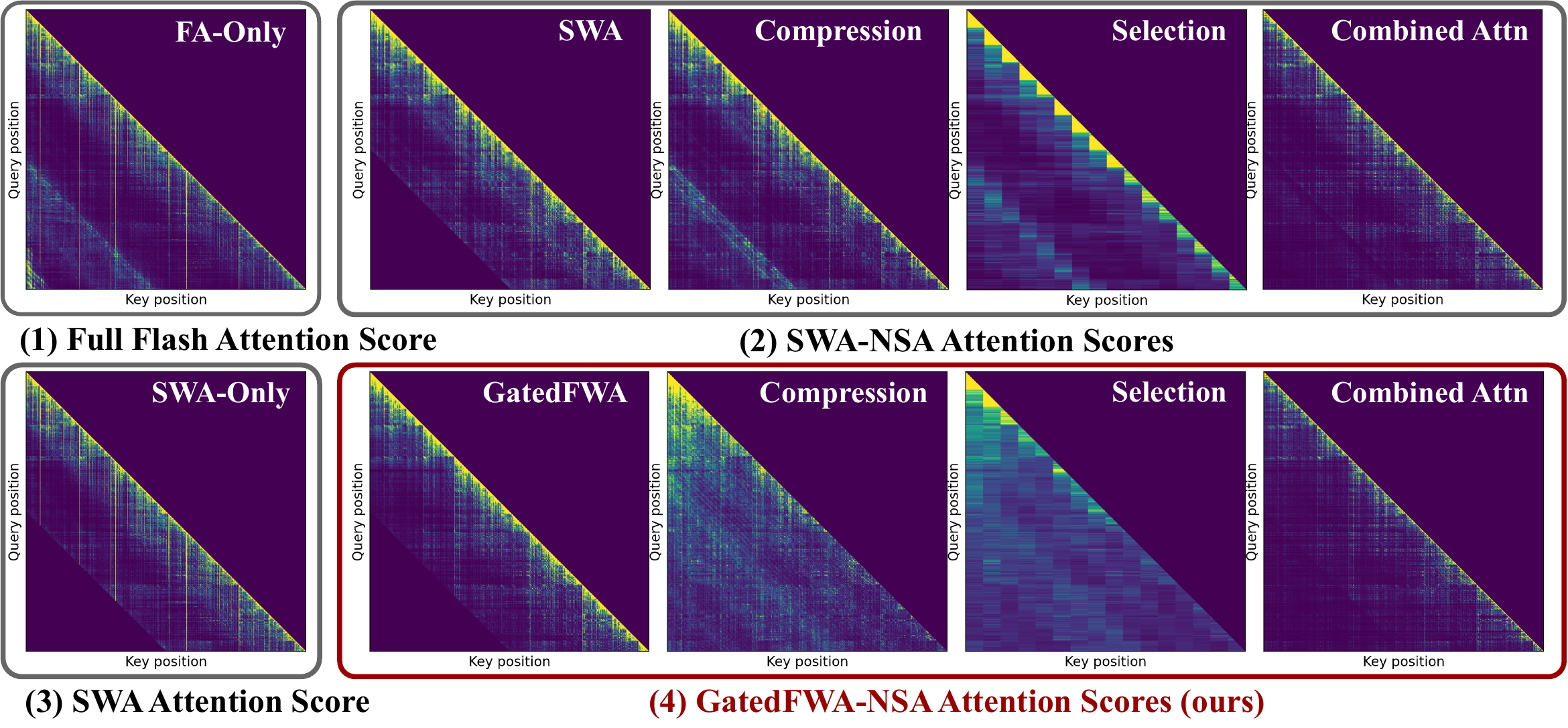}
        \caption{Final layer attention ($N=1024,w=512$) with head averaged}
    \end{subfigure}%
    \hfill
    \begin{subfigure}[t]{0.498\textwidth}
        \centering
        \includegraphics[width=\textwidth]{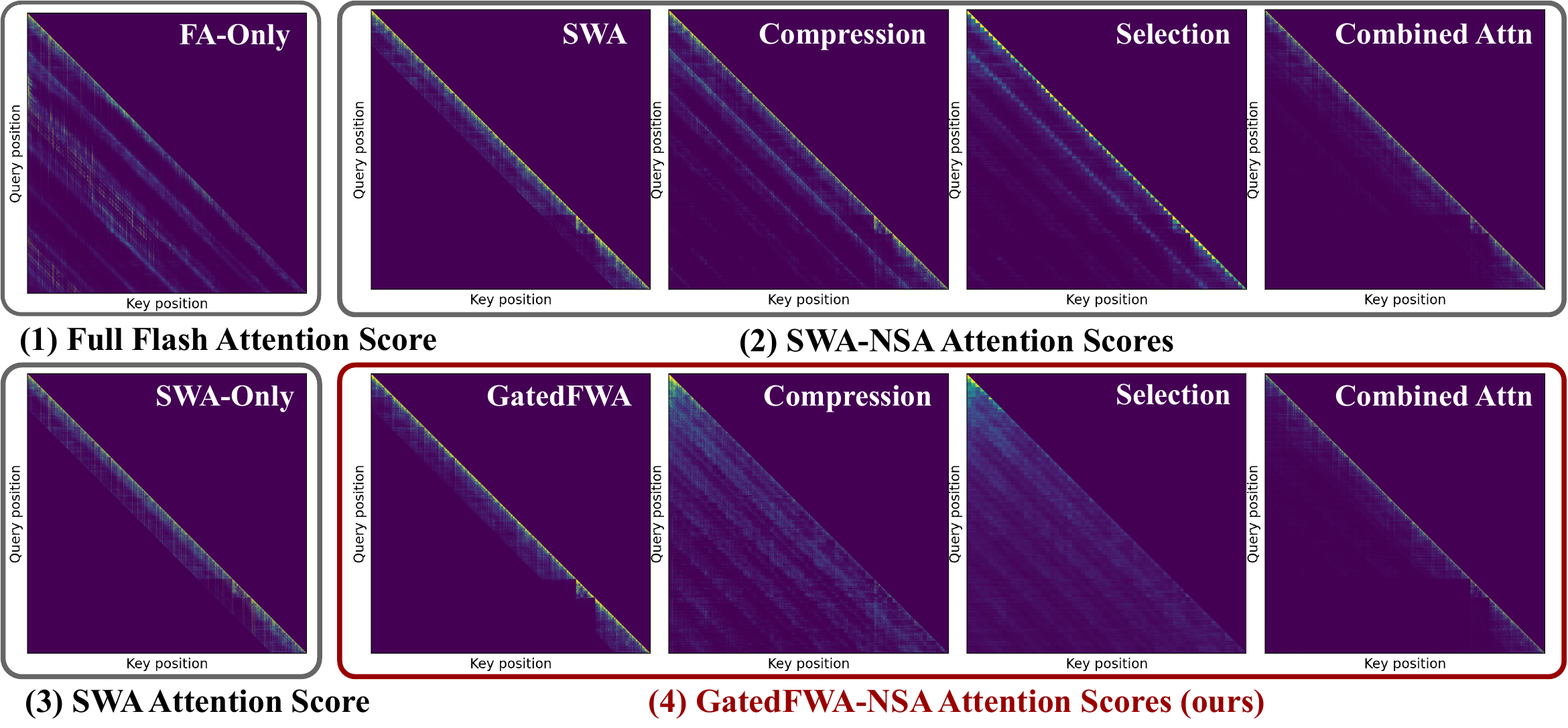}
        \caption{Final layer attention ($N=4096,w=512$) with head averaged}
    \end{subfigure}
    \caption{Attention score visualization (pretrained on OpenWebText) with various setups. 
    }
    \label{fig:visualization_attention_scores_owt}
\end{figure*}

We demonstrate in Fig.~\ref{fig:visualization_attention_scores_wiki} and Fig.~\ref{fig:visualization_attention_scores_owt} that GatedFWA-NSA produces a structurally continuous attention distribution, whereas the SWA-NSA baseline exhibits disjointed striding artifacts. These discontinuities result from SWA's hard window constraints and unbounded difference-style update ($\Delta \mathbf{M} \propto \phi_t \bf \mathbf{v}_t - \phi_{t-w} \bf \mathbf{v}_{t-w}$), which lacks a damping factor for high-magnitude tokens. In contrast, GatedFWA integrates a data-dependent gate $\alpha_t$ that applies a learnable contraction ($\mathbf{M}_t \leftarrow e^{-\boldsymbol{\alpha}_t} \mathbf{M}_{t-1}$) to the memory state. This mechanism allows the model to selectively down-weight irrelevant history (via $\boldsymbol{\alpha}_t \gg 0$), mitigating boundary artifacts and ensuring local attention aligns numerically with the global sparse blocks selected by NSA.


\begin{figure*}[t]
    \centering
    \begin{subfigure}[t]{0.162\linewidth}
        \centering
        \includegraphics[width=\textwidth]{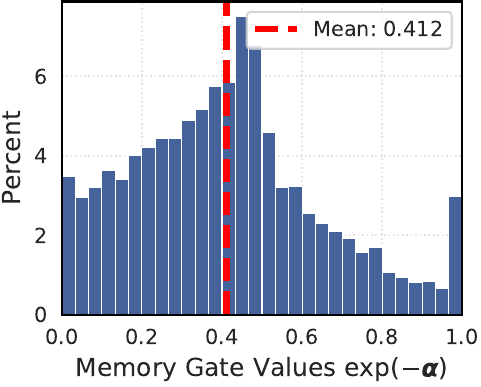}
        \caption{Layer 1}
    \end{subfigure}
    \hfill
    \begin{subfigure}[t]{0.162\linewidth}
        \centering
        \includegraphics[width=\textwidth]{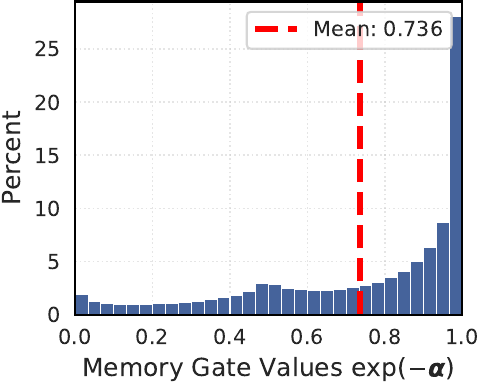}
        \caption{Layer 3}
    \end{subfigure}
    \hfill
    \begin{subfigure}[t]{0.162\linewidth}
        \centering
        \includegraphics[width=\textwidth]{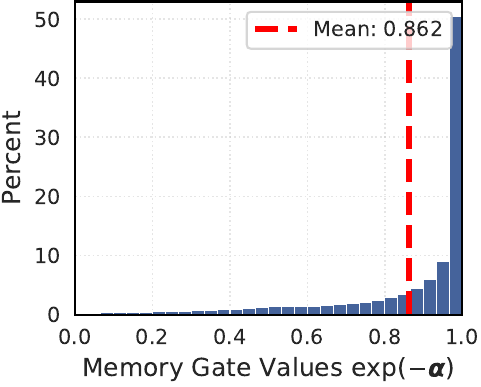}
        \caption{Layer 5}
    \end{subfigure}
    \hfill
    \begin{subfigure}[t]{0.162\linewidth}
        \centering
        \includegraphics[width=\textwidth]{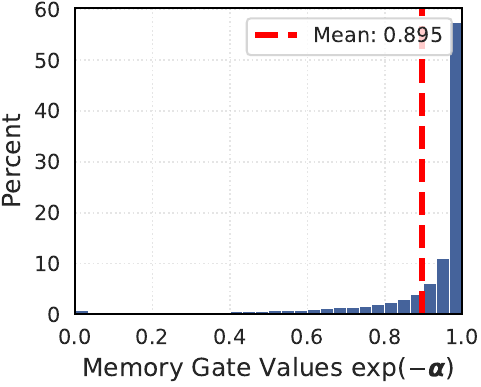}
        \caption{Layer 7}
    \end{subfigure}
    \hfill
    \begin{subfigure}[t]{0.162\linewidth}
        \centering
        \includegraphics[width=\textwidth]{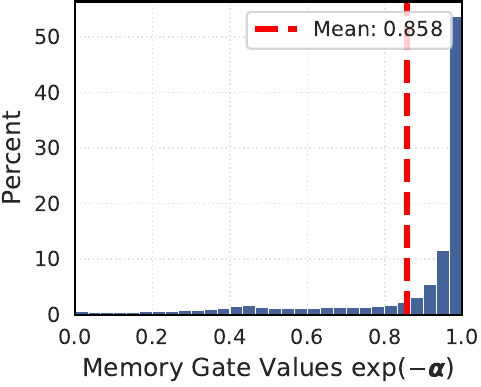}
        \caption{Layer 9}
    \end{subfigure}
    \hfill
    \begin{subfigure}[t]{0.162\linewidth}
        \centering
        \includegraphics[width=\textwidth]{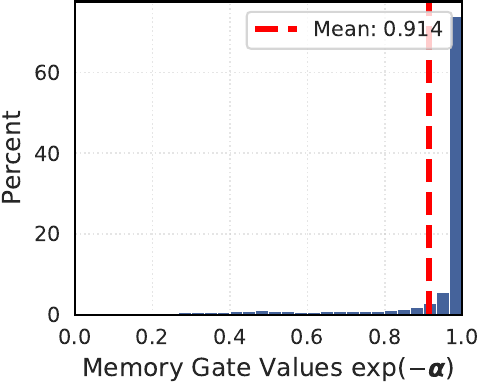}
        \caption{Layer 11}
    \end{subfigure}
    \caption{Visualization of distributions of Memory Gate Values $\exp (-\boldsymbol{\alpha}_t)$. We plot the values of layers $1\sim 12$ on the model with GatedFWA only trained on language modelling tasks.}
    \label{fig:per_layer_hists_owt_sw_fg}
\end{figure*}
\begin{figure*}[t]
    \centering
    \begin{subfigure}[t]{0.162\linewidth}
        \centering
        \includegraphics[width=\textwidth]{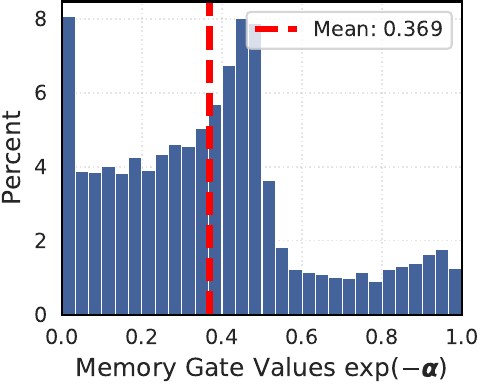}
        \caption{Layer 1}
    \end{subfigure}
    \hfill
    \begin{subfigure}[t]{0.162\linewidth}
        \centering
        \includegraphics[width=\textwidth]{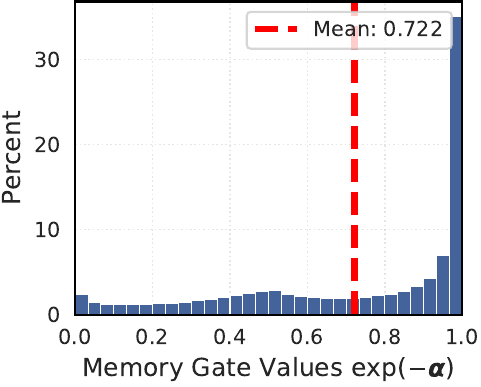}
        \caption{Layer 3}
    \end{subfigure}
    \hfill
    \begin{subfigure}[t]{0.162\linewidth}
        \centering
        \includegraphics[width=\textwidth]{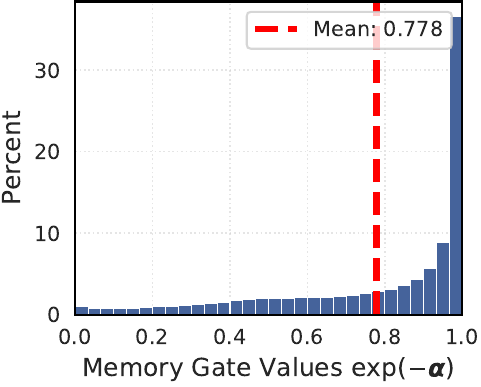}
        \caption{Layer 5}
    \end{subfigure}
    \hfill
    \begin{subfigure}[t]{0.162\linewidth}
        \centering
        \includegraphics[width=\textwidth]{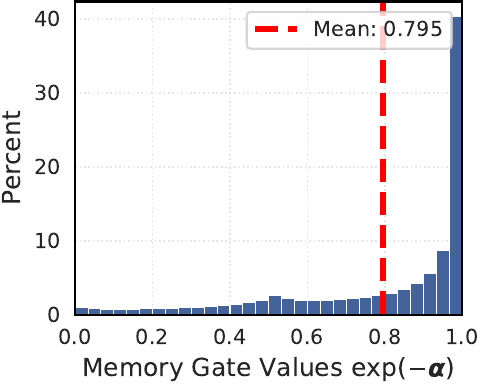}
        \caption{Layer 7}
    \end{subfigure}
    \hfill
    \begin{subfigure}[t]{0.162\linewidth}
        \centering
        \includegraphics[width=\textwidth]{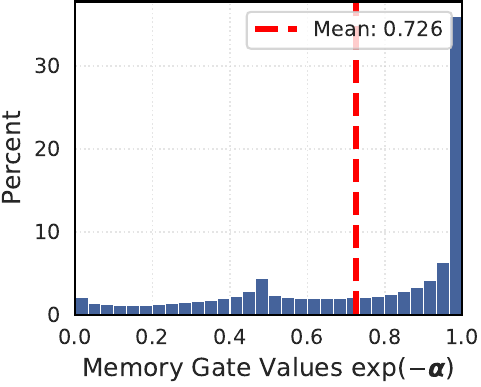}
        \caption{Layer 9}
    \end{subfigure}
    \hfill
    \begin{subfigure}[t]{0.162\linewidth}
        \centering
        \includegraphics[width=\textwidth]{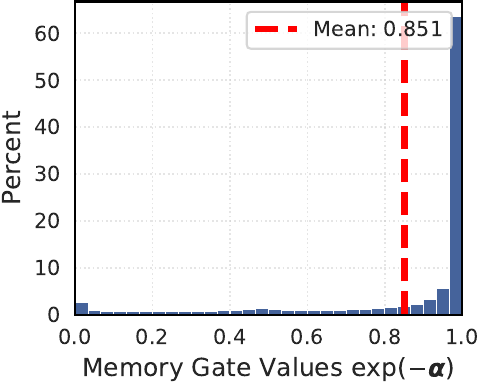}
        \caption{Layer 11}
    \end{subfigure}
    \caption{Visualization of distributions of Memory Gate Values $\exp (-\boldsymbol{\alpha}_t)$. We plot the values of layers $1\sim 12$ on the model with GatedFWA with Token Compression and Selection (NSA) extension trained on language modelling tasks.}
    \label{fig:per_layer_hists_owt_sw_fg_nsa}
\end{figure*}

\subsection{The Behaviour of Memory Gate}

We present the distributions of memory-gate values $\exp(-\alpha_t)$ across layers in Fig.~13 (standalone GatedFWA) and Fig.~14 (GatedFWA-NSA), both trained on autoregressive language modelling.  The histograms show a clear depth-wise pattern. In the lower layers, gate values are broadly spread with substantial mass in the low–to–mid range, meaning these layers frequently apply non-trivial contraction/forgetting to the carried associative memory (e.g., Layer 1 exhibits a wide distribution rather than a single sharp mode).  As depth increases, the distributions progressively shift toward values very close to 1.0 and become sharply peaked, indicating that higher layers usually set ($\alpha_t \approx 0$) and thus preserve memory with minimal decay.

This trend aligns with GatedFWA’s associative-memory interpretation. Because ($\exp(-\alpha_t)$) directly controls the multiplicative sensitivity of the recurrence and backprop path, shallow layers learn to “reset” or damp noisy/local history, while deeper layers keep gates open to sustain stable long-range credit assignment within each window.  Importantly, the same qualitative behavior appears when GatedFWA is used as the local branch inside NSA: compression/selection reshapes what is visible, but the local gated recurrence still learns a depth hierarchy where early layers filter and later layers retain, yielding a controllable, non-vanishing gradient route without the unbounded amplification seen in vanilla SWA.

\section{Proof}
\subsection{Proof of Thm.~\ref{thm:recurrent_forms}}
\begin{proof}
    For Softmax attention in the recurrent output form,
\begin{align}
    \mathbf{o}_t &= \sum_{i=1}^t \mathbf{M}_{ti} \mathbf{v}_i\\
    &= \sum_{i=1}^t \frac{\exp(\mathbf{q}_t \mathbf{k}_i^\top / \sqrt{d_h})}{
    \sum_{k=1}^t \exp(\mathbf{q}_t \mathbf{k}_k^\top / \sqrt{d_h})
    } \mathbf{v}_i\\
    &= \sum_{i=1}^t \frac{
    \phi(\mathbf{q}_t) \phi(\mathbf{k}_i)^\top 
    }{
    \sum_{k=1}^t \exp(\mathbf{q}_t \mathbf{k}_k^\top / \sqrt{d_h})
    } \mathbf{v}_i,
\end{align}
where $\phi$ is the feature map as described in Thm.~\ref{thm:recurrent_forms} for near perfect kernel approximation. For big sample size $t$, we have $\frac{1}{t}\mathbb{E}[\sum_{k=1}^t \exp(\mathbf{q}_t \mathbf{k}_k^\top / \sqrt{d_h})]  = c(\mathbf{q}_t)$ according to law of large numbers, where $c(\mathbf{q}_t)$ is a query-specific constant. Thus we can reformulate the output by
\begin{align}
    \mathbf{o}_t = \frac{1}{c(\mathbf{q}_t)} \sum_{i=1}^t \frac{\phi(\mathbf{q}_t) \phi(\mathbf{k}_i)^\top }{t} \mathbf{v}_i.
\end{align}
Let the Softmax attention associative memory be
\begin{align}
    \mathbf{M}_t = \sum_{i=1}^t \frac{\phi(\mathbf{k}_i)^\top \mathbf{v}_i}{t} \in \mathbb{R}^{\mathrm{dim}(\phi) \times d_v},
\end{align}
we have the memory recurrence
\begin{align}
    \mathbf{M}_t 
    &= \frac{t-1}{t} (\sum_{i=1}^{t-1} \frac{\phi(\mathbf{k}_i)^\top \mathbf{v}_i}{t-1} + \frac{\phi(\mathbf{k}_t)^{\top} \mathbf{v}_t}{t-1}  )\\
    &= \frac{t-1}{t}\mathbf{M}_{t-1} + \frac{1}{t} \phi(\mathbf{k}_t)^{\top} \mathbf{v}_t.
\end{align}
This gives the form as Eq.~\eqref{eq:recurrent_form_softmax}. 

Similarly for SWA, we have the recurrent output form
\begin{align}
    \mathbf{o}_t = \sum_{i=t-w+1}^t \frac{
    \phi(\mathbf{q}_t) \phi(\mathbf{k}_i)^\top 
    }{
    \sum_{k=t-w+1}^{t} \exp(\mathbf{q}_t \mathbf{k}_k^\top / \sqrt{d_h})
    } \mathbf{v}_i .
\end{align}
Assume similarly that $\frac{1}{w}\mathbb{E}[\sum_{k=t-w+1}^{t} \exp(\mathbf{q}_t \mathbf{k}_k^\top / \sqrt{d_h})] = c(\mathbf{q}_t)$, then we can reformulate the recurrence as
\begin{align}
    \mathbf{o}_t = \frac{1}{c(\mathbf{q}_t)} \sum_{i=t-w+1}^t \frac{\phi(\mathbf{q}_t) \phi(\mathbf{k}_i)^\top }{t} \mathbf{v}_i.
\end{align}
Similarly with the definition of SWA associative memory $\mathbf{M}_t = \sum_{i=t-w+1}^t \frac{\phi(\mathbf{k}_i)^\top \mathbf{v}_i}{w} \in \mathbb{R}^{\mathrm{dim}(\phi) \times d_v}$. Then for memory recurrent form we have
\begin{align}
    \mathbf{M}_t 
    &= \sum_{i=t-w}^{t-1} \frac{\phi(\mathbf{k}_i)^\top \mathbf{v}_i}{w}  + \frac{\phi(\mathbf{k}_t)^{\top} \mathbf{v}_t}{w} - \frac{\phi(\mathbf{k}_{t-w})^{\top} \mathbf{v}_{t-w}}{w}  \nonumber\\
    &= \mathbf{M}_{t-1} + \frac{1}{w}(\phi(\mathbf{k}_i)^\top \mathbf{v}_i - \phi(\mathbf{k}_{t-w})^{\top} \mathbf{v}_{t-w})
\end{align}
This gives the form as Eq.~\eqref{eq:recurrent_form_swa}. 
\end{proof}

\subsection{Proof of Prop.~\ref{prop:recurrence_gatedfwa}}
\begin{proof}
Given the GatedFWA 
\begin{align}
    \mathbf{o}_t &= \sum_{i=t-w+1}^t \tilde{\mathbf{M}}_{ti} \mathbf{v}_i\\
    & = \sum_{i=t-w+1}^t \frac{
    \exp (\mathbf{q}_t \mathbf{k}_i^\top /\sqrt{d_h} + \mathbf{B}_{ti})
    }{
    \sum_{k=t-w+1}^t  \exp (\mathbf{q}_t \mathbf{k}_k^\top /\sqrt{d_h} + \mathbf{B}_{tk})
    } \mathbf{v}_i \nonumber\\
    & = \sum_{i=t-w+1}^t \frac{
    \exp (\mathbf{B}_{ti}) \phi(\mathbf{q}_t) \phi(\mathbf{k}_i)^\top
    }{
    \sum_{k=t-w+1}^t  \exp(\mathbf{B}_{tk}) \exp (\mathbf{q}_t \mathbf{k}_k^\top /\sqrt{d_h} )
    } \mathbf{v}_i \nonumber
\end{align}
As $\mathbf{B}_{tk}$ are some input dependent constant, we can assume similarly as Thm.~\ref{thm:recurrent_forms} that 
\begin{align}
    \frac{1}{w}\mathbb{E}[\sum_{k=t-w+1}^{t} \exp(\mathbf{B}_{tk}) \exp(\mathbf{q}_t \mathbf{k}_k^\top / \sqrt{d_h})] = c(\mathbf{q}_t),
\end{align}
thus we can reformulate the output by
\begin{align}
    \mathbf{o}_t &= \frac{1}{c(\mathbf{q}_t)} \sum_{i=t-w+1}^t\frac{\exp (\mathbf{B}_{ti}) \phi(\mathbf{q}_t) \phi(\mathbf{k}_i)^\top}{w} \mathbf{v}_i\\
    &= \frac{1}{c(\mathbf{q}_t)}  \frac{
    \phi(\mathbf{q}_t) 
    (\sum_{i=t-w+1}^t
    (\exp (\mathbf{B}_{ti}) \mathbf{I}_k)
    \phi(\mathbf{k}_i)^\top \mathbf{v}_i )
    }{w} 
\end{align}
Let the GatedFWA associative memory be
\begin{align}
    \mathbf{M}_t = \sum_{i=t-w+1}^t\frac{ 
    (\exp (\mathbf{B}_{ti}) \mathbf{I}_k)
    \phi(\mathbf{k}_i)^\top \mathbf{v}_i 
    }{w},
\end{align}
we have the gated memory recurrence
\begin{align}
    \mathbf{M}_t &= \sum_{i=t-w+1}^t\frac{ 
    (\exp (\mathbf{B}_{ti}) \mathbf{I}_k)
    \phi(\mathbf{k}_i)^\top \mathbf{v}_i 
    }{w}\\
     &= \frac{1}{w}(\sum_{i=t-w}^{t-1} (\exp (\mathbf{B}_{ti}) \mathbf{I}_k) \phi(\mathbf{k}_i)^\top\mathbf{v}_i + (\exp (\mathbf{B}_{tt}) \mathbf{I}_k)\phi(\mathbf{k}_t)^\top\mathbf{v}_t 
      -
     (\exp (\mathbf{B}_{t-1, t-w}) \mathbf{I}_k) \phi(\mathbf{k}_{t-w})^\top \mathbf{v}_{t-w}
     )  \nonumber\\
    &= \frac{1}{w}(
    \sum_{i=t-w}^{t-1} (\exp(-\sum_{j=i+1}^t \boldsymbol{\alpha}_j)\mathbf{I}_k) \phi(\mathbf{k}_i)^\top\mathbf{v}_i + \phi(\mathbf{k}_t)^\top\mathbf{v}_t) 
     - 
    (\exp(-\sum_{j=t-w+1}^{t-1} \boldsymbol{\alpha}_j)\mathbf{I}_k)
    \phi(\mathbf{k}_{t-w})^\top \mathbf{v}_{t-w}
    ) \nonumber\\
    &= \frac{1}{w}(
    \sum_{i=t-w}^{t-1} (\exp(-\boldsymbol{\alpha}_t)\exp(-\sum_{j=i+1}^{t-1} \boldsymbol{\alpha}_j) \mathbf{I}_k) \phi(\mathbf{k}_i)^\top\mathbf{v}_i + \phi(\mathbf{k}_t)^\top\mathbf{v}_t) 
     - 
    (\prod_{j=t-w+1}^{t-1}\exp(- \boldsymbol{\alpha}_j) \mathbf{I}_k)
    \phi(\mathbf{k}_{t-w})^\top \mathbf{v}_{t-w}
    ) \nonumber\\
    &= \frac{1}{w} (
    (\exp(-\boldsymbol{\alpha}_t)\mathbf{I}_k) 
    \sum_{i=t-w}^{t-1} 
    (\exp(-\sum_{j=i+1}^{t-1} \boldsymbol{\alpha}_j) \mathbf{I}_k)
    \phi(\mathbf{k}_i)^\top\mathbf{v}_i + \phi(\mathbf{k}_t)^\top\mathbf{v}_t) 
    -
    (\prod_{j=t-w+1}^{t-1}\exp(- \boldsymbol{\alpha}_j) \mathbf{I}_k)
    \phi(\mathbf{k}_{t-w})^\top \mathbf{v}_{t-w}
    ) \nonumber\\
    &= (\exp(-\boldsymbol{\alpha}_t)\mathbf{I}_k)\mathbf{M}_{t-1} + \frac{1}{w} (\phi(\mathbf{k}_t)^\top \mathbf{v}_t 
    - 
    (\prod_{j=t-w+1}^{t-1}\exp(- \boldsymbol{\alpha}_j)\mathbf{I}_k)
    \phi(\mathbf{k}_{t-w})^\top \mathbf{v}_{t-w}). \nonumber
\end{align}
For simplicity, denote $\mathbf{c}_t = \prod_{j=t-w+1}^{t-1}\exp(- \boldsymbol{\alpha}_j) \in (0,1)$ and let
\begin{align}
    \mathbf{D}_t = \frac{1}{w}(
    \phi(\mathbf{k}_t)^\top \mathbf{v}_t
    -
    (\prod_{j=t-w+1}^{t-1}\exp(- \boldsymbol{\alpha}_j)\mathbf{I}_k)
    \phi(\mathbf{k}_{t-w})^\top \mathbf{v}_{t-w}) 
    =
    \phi(\mathbf{k}_t)^\top \mathbf{v}_t
    -
    (\mathbf{c}_t\mathbf{I}_k)
    \phi(\mathbf{k}_{t-w})^\top \mathbf{v}_{t-w})
    ,
\end{align}
we can derive the objective for the associative memory recurrence as
\begin{align}
    \mathcal{L}_t(\mathbf{M}_{t-1}) &= \frac{1}{2} \mathrm{tr}(\mathbf{M}_{t-1} \mathbf{M}_{t-1}^\top) - \frac{1}{2}\mathrm{tr}(
    (\exp(-\boldsymbol{\alpha}_t)\mathbf{I}_k)
    \mathbf{M}_{t-1} \mathbf{M}_{t-1}^\top
    )
    -
    \mathrm{tr}(\mathbf{M}_{t-1}\mathbf{D}_{t}^\top)
     \\
    &= \frac{1}{2}\|  ( \sqrt{1-\exp(\boldsymbol{-\alpha}_t)}  \mathbf{I}_k ) \mathbf{M}_{t-1}\|_F^2 - 
    \frac{1}{w} (
    \mathbf{c}_t\mathbf{I}_k
    \phi(\mathbf{k}_{t-w})\mathbf{M}_{t-1}\mathbf{v}_{t-1}^\top 
    - 
    \phi(\mathbf{k}_t)\mathbf{M}_{t-1}\mathbf{v}_t^\top
    ) \nonumber \\
    &= \frac{1}{2}\|  ( \sqrt{1-\exp(\boldsymbol{-\alpha}_t)}  \mathbf{I}_k ) \mathbf{M}_{t-1}\|_F^2 - 
    \frac{1}{w} \langle
    \mathbf{M}_{t-1}
    ,
    \mathbf{c}_t\mathbf{I}_k
    \phi(\mathbf{k}_{t-w})\mathbf{v}_{t-1}^\top 
    - 
    \phi(\mathbf{k}_t)\mathbf{v}_t^\top
    \rangle \nonumber\\
    &= \frac{1}{2}\| ( \sqrt{1-\exp(\boldsymbol{-\alpha}_t)}  \mathbf{I}_k ) \mathbf{M}_{t-1}\|_F^2 - 
    \frac{1}{w} \langle
    \mathbf{M}_{t-1}
    ,
    \mathbf{c}_t\mathbf{I}_k
    \Delta_t
    + (1-\mathbf{c}_t)\mathbf{I}_k 
    \phi(\mathbf{k}_t)\mathbf{v}_t^\top
    \rangle, \nonumber
\end{align}
where $\Delta_t = \phi(\mathbf{k}_{t-w})^\top\mathbf{v}_{t-w} - \phi(\mathbf{k}_{t})^\top\mathbf{v}_{t}$. This concludes the proof.
\end{proof}

\section{Supplemental Kernel Implementation Details}  
\subsection{Preprocessing via Scan-Then-Propagate}
\label{app:hierarchical-scan}
\begin{algorithm}[t]
\caption{Gated Processing (Scan-Then-Propagate) Kernel}
\label{alg:hierarchical-scan}
\begin{algorithmic}[1]
\Require Matrices $\mathbf{H}, \boldsymbol{\beta}$ (as defined in Alg.~\ref{alg:fused-gated-scan}), chunk size $B_t$, small $\varepsilon>0$.
\State Divide $\mathbf{H}$, $\boldsymbol{\beta}$ into $T_t=\left\lceil \frac{N}{B_t}\right\rceil$ blocks $\mathbf{h}_1,\ldots,\mathbf{h}_{T_t}$ and $\boldsymbol{\beta}_1,\ldots,\boldsymbol{\beta}_{T_t}$ of size $B_t\times H$.
\State Allocate workspace matrices $\mathbf{S}, \mathbf{O} \in \mathbb{R}^{T_t \times H}$ in HBM.

\Statex \textbf{Phase 1: Parallel Block Reduction}
\For{$1 \le i \le T_t$ in parallel}
    \State Load chunk $\mathbf{h}_i$, $\boldsymbol{\beta}_i$ from HBM to SRAM.
    \State On chip, compute $\mathbf{z}_i\gets \boldsymbol{\beta}_i \odot \mathbf{h}_i$, $\boldsymbol{\nu}_i\gets \max(\mathbf{z}_i,0)$.
    \State On chip, compute $\mathrm{softplus}(\mathbf{z}_i)\gets \boldsymbol{\nu}_i + \log (e^{\mathbf{z}_i-\boldsymbol{\nu}_i} + e^{-\boldsymbol{\nu}_i})$.
    \State On chip, compute $\boldsymbol{\alpha}_i\gets \mathrm{softplus}(\mathbf{z}_i) \odot (\boldsymbol{\beta}_i+\varepsilon)^{-1}$.
    \State On chip, compute block sum $\mathbf{s}_i \gets \sum -\boldsymbol{\alpha}_i$ (sum over time dimension).
    \State Write $\mathbf{s}_i$ to HBM as $i$-th row of $\mathbf{S}$.
\EndFor

\Statex \textbf{Phase 2: Global Scan on Aggregates}
\State On chip (or via separate kernel), compute $\mathbf{O} \gets \mathrm{cumsum}(\mathbf{S})$ along time dimension.
\State Write $\mathbf{O}$ to HBM.

\Statex \textbf{Phase 3: Re-compute \& Distribute}
\For{$1 \le i \le T_t$ in parallel}
    \State Load chunk $\mathbf{h}_i$, $\boldsymbol{\beta}_i$ and global offset $\mathbf{o}_{i-1}$ (from $\mathbf{O}$, if $i>1$) from HBM to SRAM.
    \State On chip, set $\textsc{carry} \gets \mathbf{o}_{i-1}$ (if $i=1$ set $\mathbf{0}_H$).
    \State On chip, recompute $\mathbf{z}_i, \boldsymbol{\nu}_i, \boldsymbol{\alpha}_i$ (same as Phase 1).
    \State On chip, compute $\mathbf{p}_i\gets \mathrm{cumsum}(-\boldsymbol{\alpha}_i) + \textsc{carry}$.
    \State Write $\mathbf{u}_i \gets \mathbf{p}_i$ to HBM.
\EndFor
\State \Return $\mathbf{U}$.
\end{algorithmic}
\end{algorithm}

We additionally provide the Fused \emph{Scan-Then-Propagate algorithm} for gate preprocessing in Alg.~\ref{alg:hierarchical-scan}. Theoretically, this decouples the sequential prefix scan dependency across blocks, reducing the time complexity from $\mathcal{O}(N)$ to $\mathcal{O}(\log N)$ and parallelizing carry propagation to maximize GPU SM utilization on long sequences.

However, benchmarking results in Fig.~\ref{fig:time_efficiency_cmp}(d) demonstrate that the 1-Pass (Fused Tiled Scan) algorithm consistently outperforms the Scan-Then-Propagate variant ($\sim 28.5$ vs. $\sim 20.1$ billion tokens/s). This gap exists because the operation is memory-bandwidth bound. The 1-Pass algorithm achieves optimal I/O efficiency by reading inputs $\mathbf{H}$ and $\boldsymbol{\beta}$ exactly once. In contrast, Scan-Then-Propagate effectively doubles the memory traffic by re-reading inputs during Phase 3 to recompute $\boldsymbol{\alpha}$ on-the-fly, avoiding the costlier storage of large intermediate matrices. Furthermore, any theoretical latency advantage is nullified by device saturation. With $H=128$, the 1-Pass algorithm already provides sufficient work to fully saturate a modern GPU's SMs. The fine-grained parallelism of the 3-Phase approach adds kernel launch and synchronization overheads without unlocking idle resources. Consequently, the 1-Pass Fused Scan remains the optimal choice for this workload.

\subsection{GatedFWA Backward Pass Kernel}
\begin{algorithm}[t]
\caption{GatedFWA Backward Kernel}
\label{alg:gated_swfa_backward}
\small
\begin{algorithmic}[1]
\Require Matrices $\mathbf{Q},\mathbf{K},\mathbf{V},\mathbf{O},\mathbf{d}\mathbf{O}\in\mathbb{R}^{N\times d}$ in HBM, 
vector $\mathbf{U}\in\mathbb{R}^{N}$ in HBM, vector $L\in\mathbb{R}^N$ in HBM, 
block sizes $B_c,B_r$, window size $w$.
\Ensure $\mathbf{d}\mathbf{Q},\mathbf{d}\mathbf{K},\mathbf{d}\mathbf{V}\in\mathbb{R}^{N\times d}$, $\mathbf{d}\mathbf{U}\in\mathbb{R}^{N}$.

\State Divide $\mathbf{Q}$ into $T_r= \lceil \frac{N}{B_r} \rceil$ blocks $\mathbf{q}_1,\ldots,\mathbf{q}_{T_r}$ of size $B_r\times d$, 
and divide $\mathbf{K},\mathbf{V}$ into $T_c= \lceil \frac{N}{B_c} \rceil$ blocks $\mathbf{k}_1,\ldots,\mathbf{k}_{T_c}$ and $\mathbf{v}_1,\ldots,\mathbf{v}_{T_c}$ of size $B_c\times d$.
\State Divide $\mathbf{O}$ into $T_r$ blocks $\mathbf{o}_1,\ldots,\mathbf{o}_{T_r}$, and divide $\mathbf{d}\mathbf{O}$ into $T_r$ blocks $\mathbf{do}_1,\ldots,\mathbf{do}_{T_r}$.
\State Divide $L$ into $T_r$ blocks $L_1,\ldots,L_{T_r}$ of size $B_r$.
\State \colorbox{gray!30}{Let $\mathbf{U}^q=\mathbf{U}$ and $\mathbf{U}^k=\mathbf{U}$. Divide $\mathbf{U}^q$ into $T_r$ blocks $\mathbf{u}^q_1,\ldots,\mathbf{u}^q_{T_r}$ of size $B_r$, and divide $\mathbf{U}^k$ into $T_c$ blocks $\mathbf{u}^k_1,\ldots,\mathbf{u}^k_{T_c}$ of size $B_c$.}
\State Initialize $\mathbf{d}\mathbf{Q},\mathbf{d}\mathbf{K},\mathbf{d}\mathbf{V}\gets (0)_{N\times d}$ in HBM. 
\colorbox{gray!30}{Initialize $\mathbf{d}\mathbf{U}^q,\mathbf{d}\mathbf{U}^k\gets (0)_N$ in HBM.}

\State Compute $D=\operatorname{rowsum}(\mathbf{O}\odot \mathbf{d}\mathbf{O})\in\mathbb{R}^{N}$ (pointwise multiply), write $\mathbf{d}$ to HBM and divide into $T_r$ blocks $D_1,\ldots,D_{T_r}$ of size $B_r$.

\For{$1 \le j \le T_c$}
  \State Load $\mathbf{k}_j,\mathbf{v}_j,\mathbf{u}^k_j$ from HBM to on-chip SRAM.
  \State On chip, initialize $\mathbf{dk}_j,\mathbf{dv}_j\gets (0)_{B_c\times d}$, \colorbox{gray!30}{$\mathbf{du}^k_j\gets (0)_{B_c}$.}
  \State \colorbox{gray!30}{Let $g_{\mathrm{start}}=(j-1)B_c$, $g_{\mathrm{end}}=\min(jB_c,N)-1$.}
  \State \colorbox{gray!30}{$q_{\mathrm{lo}}\gets g_{\mathrm{start}}$, $q_{\mathrm{hi}}\gets \min(N, g_{\mathrm{end}}+w)$ (exclusive).}
  \State \colorbox{gray!30}{$i_{\mathrm{lo}}\gets  \lfloor \frac{q_{\mathrm{lo}}}{B_r} \rfloor + 1$, $i_{\mathrm{hi}}\gets  \lceil \frac{q_{\mathrm{hi}}}{B_r} \rceil$.}

  \For{\colorbox{gray!30}{$i=i_{\mathrm{lo}}\ldots i_{\mathrm{hi}}$}}
    \State Load $\mathbf{q}_i,\mathbf{o}_i,\mathbf{do}_i,L_i,\mathbf{d}_i,\mathbf{u}^q_i$ from HBM to on-chip SRAM.
    \State On chip, initialize $\mathbf{dq}_i\gets (0)_{B_r\times d}$, \colorbox{gray!30}{$\mathbf{du}^q_i\gets (0)_{B_r}$.}

    \State On chip, compute $\mathbf{S}_i^{(j)}\gets \texttt{sm\_scale} \cdot \mathbf{q}_i\mathbf{k}_j^\top \in\mathbb{R}^{B_r\times B_c}$.
    \State \colorbox{gray!30}{On chip, compute gate bias $\mathbf{B}_i^{(j)}\gets \mathbf{u}^q_i\mathbf{1}^\top-\mathbf{1}(\mathbf{u}^k_j)^\top$, and set $\mathbf{S}_i^{(j)}\gets \mathbf{S}_i^{(j)}+\mathbf{B}_i^{(j)}$.}

    \For{each row $r\in[0,B_r)$ and col $c\in[0,B_c)$ with global indices $q=(i-1)B_r+r$, $g=g_{\mathrm{start}}+c$}
      \State \colorbox{gray!30}{keep $\mathbf{S}_i^{(j)}[r,c]$ iff $q-w+1\le g\le q$; otherwise set to $-\infty$.}
    \EndFor

    \State On chip, compute $\mathbf{p}_i^{(j)}\gets \exp(\mathbf{S}_i^{(j)}-L_i\mathbf{1}^\top)\in\mathbb{R}^{B_r\times B_c}$.
    \State On chip, compute $\mathbf{dp}_i^{(j)}\gets \mathbf{do}_i\mathbf{v}_j^\top\in\mathbb{R}^{B_r\times B_c}$.
    \State On chip, compute $\mathbf{ds}_i^{(j)}\gets \mathbf{p}_i^{(j)}\odot(\mathbf{dp}_i^{(j)}-D_i \mathbf{1}^\top)\in\mathbb{R}^{B_r\times B_c}$.
    
    \State On chip, update $\mathbf{dv}_j\gets \mathbf{dv}_j + (\mathbf{p}_i^{(j)})^\top\mathbf{do}_i\in\mathbb{R}^{B_c\times d}$.
    \State On chip, update $\mathbf{dq}_i\gets \mathbf{dq}_i + \mathbf{ds}_i^{(j)}\mathbf{k}_j\in\mathbb{R}^{B_r\times d}$.
    \State \colorbox{gray!30}{On chip, update $\mathbf{du}^q_i\gets \mathbf{du}^q_i + \operatorname{rowsum}(\mathbf{ds}_i^{(j)})\in\mathbb{R}^{B_r}$.}

    \State Load $\mathbf{dq}_i$ from SRAM to HBM, then on chip update $\mathbf{d}\mathbf{Q}_i \gets \mathbf{d}\mathbf{Q}_i + \mathbf{dq}_i$, and write back to HBM.
    \State \colorbox{gray!30}{Load $\mathbf{du}^q_i$ from SRAM to HBM, then on chip update $\mathbf{d}\mathbf{U}^q_i \gets \mathbf{d}\mathbf{U}^q_i + \mathbf{du}^q_i$, and write back to HBM.}
    
    \State On chip, update $\mathbf{dk}_j\gets \mathbf{dk}_j + (\mathbf{ds}_i^{(j)})^\top\mathbf{q}_i\in\mathbb{R}^{B_c\times d}$.
    \State \colorbox{gray!30}{On chip, update $\mathbf{du}^k_j\gets \mathbf{du}^k_j - \operatorname{rowsum}(\mathbf{ds}_i^{(j)})\in\mathbb{R}^{B_c}$.}
  \EndFor

  \State Write $\mathbf{dk}_j,\mathbf{dv}_j$ to HBM as blocks of $\mathbf{d}\mathbf{K},\mathbf{d}\mathbf{V}$.
  \State \colorbox{gray!30}{Write $\mathbf{du}^k_j$ to HBM as the $j$-th block of $\mathbf{d}\mathbf{U}^k$.}
\EndFor

\State \colorbox{gray!30}{Compute $\mathbf{d}\mathbf{U}\gets \mathbf{d}\mathbf{U}^q + \mathbf{d}\mathbf{U}^k$.}
\State \Return $\mathbf{d}\mathbf{Q},\mathbf{d}\mathbf{K},\mathbf{d}\mathbf{V},\mathbf{d}\mathbf{U}$.
\end{algorithmic}
\end{algorithm}

In Alg.~\ref{alg:gated_swfa_backward}, we present a hardware-aware backward pass for GatedFWA that mirrors the FlashAttention-2 style tiling while incorporating the gated sliding-window bias. The sequence is partitioned into row blocks of queries and column blocks of keys/values, and the gate prefix-sums are simply reused as $\mathbf{U}^q$ and $\mathbf{U}^k$ and block-partitioned accordingly. The backward begins with a preprocessing step that computes the row-wise dot product $D=\operatorname{rowsum}(\mathbf{O}\odot \mathbf{d}\mathbf{O})$, which is later used to form the stable softmax gradient. For each KV block $\mathbf{k}_j,\mathbf{v}_j$, we iterate only over the query blocks that can attend to it under the causal sliding window, recompute the gated logits $\mathbf{S}_i^{(j)}=\texttt{sm\_scale}\cdot \mathbf{q}_i\mathbf{k}_j^\top+\mathbf{u}^q_i\mathbf{1}^\top-\mathbf{1}(\mathbf{u}^k_j)^\top$, apply the same window mask as in forward, and reconstruct probabilities via the saved log-normalizer $L_i$. Using these, we accumulate $\mathbf{d}\mathbf{V}$, $\mathbf{d}\mathbf{K}$, and the gate gradients $\mathbf{d}\mathbf{U}^k$ in a streaming fashion, while also producing per-tile contributions to $\mathbf{d}\mathbf{Q}$ and $\mathbf{d}\mathbf{U}^q$ that are immediately written back. For ease of presentation we describe all gradient updates in a single algorithm, but in practice the Triton implementation computes $(\mathbf{d}\mathbf{K},\mathbf{d}\mathbf{V},\mathbf{d}\mathbf{U}^k)$ and $(\mathbf{d}\mathbf{Q},\mathbf{d}\mathbf{U}^q)$ in two separate kernels, matching the forward-compatible recomputation and masking used by GatedFWA.

\section{Final Remarks}
\subsection{Kernel Mapping}
\label{app:kernel_mapping}

\paragraph{Exponential Kernel and Infinite-Dimensional Feature Space.}
In Sec.~\ref{sec:motivation}, we interpret the Softmax attention as an associative memory mechanism governed by an exponential kernel $\mathcal{K}(\mathbf{q}, \mathbf{k}) = \exp(\frac{\mathbf{q}\mathbf{k}^\top}{\sqrt{d}})$. Here, we provide the explicit construction of the feature map $\phi(\cdot)$ that linearizes this kernel, justifying the memory recurrence formulation.

Given query vector $\mathbf{q} \in \mathbb{R}^{1 \times d}$ and key vector $\mathbf{k} \in \mathbb{R}^{1 \times d}$, utilizing the Taylor series expansion of the exponential function, we can decompose the kernel as
\begin{align}
\mathcal{K}(\mathbf{q}, \mathbf{k}) &= \exp (\frac{\mathbf{q}\mathbf{k}^\top}{\sqrt{d}} ) = \sum_{n=0}^{\infty} \frac{1}{n!}  (\frac{\mathbf{q}\mathbf{k}^\top}{\sqrt{d}} )^n = \sum_{n=0}^{\infty} \frac{1}{n! (\sqrt{d})^n} (\mathbf{q}\mathbf{k}^\top)^n.
\end{align}
Noting that $(\mathbf{q}\mathbf{k}^\top)^n = \langle \mathbf{q}^{\otimes n}, \mathbf{k}^{\otimes n} \rangle$, where $\otimes$ denotes the tensor product and the vectors are flattened, we can define the feature map $\phi: \mathbb{R}^{1 \times d} \to \mathcal{H}$ mapping input vectors to an infinite-dimensional Hilbert space:
\begin{align}
    \phi(\mathbf{x}) =  [ 1, \frac{1}{\sqrt{1!\sqrt{d}}} \mathbf{x}, \frac{1}{\sqrt{2!(\sqrt{d})^2}} \mathbf{x}^{\otimes 2}, \dots, \frac{1}{\sqrt{n!(\sqrt{d})^n}} \mathbf{x}^{\otimes n}, \dots  ]^\top
\end{align}
With this construction, the inner product in the feature space recovers the exponential kernel exactly:
\begin{align}
    \langle \phi(\mathbf{q}), \phi(\mathbf{k}) \rangle = \sum_{n=0}^{\infty} \frac{1}{n!}  (\frac{\mathbf{q}\mathbf{k}^\top}{\sqrt{d}} )^n = \exp (\frac{\mathbf{q}\mathbf{k}^\top}{\sqrt{d}} )
\end{align}
This formalizes the assumption in Thm.~\ref{thm:recurrent_forms} that $\langle \phi(\mathbf{q}), \phi(\mathbf{k}) \rangle \approx \exp(\mathbf{q}\mathbf{k}^\top / \sqrt{d})$. Consequently, the Softmax attention operation can be strictly viewed as retrieving from a memory $\mathbf{M}_t = \sum_{i=1}^t \phi(\mathbf{k}_i)^\top \mathbf{v}_i$ via the normalized associative map: $\mathbf{o}_t = \frac{\phi(\mathbf{q}_t) \mathbf{M}_t}{\sum_{j=1}^t \langle \phi(\mathbf{q}_t), \phi(\mathbf{k}_j) \rangle}$. This kernel perspective reveals the origin of the gradient vanishing issue discussed in Prop.~\ref{prop:optimization_objs}: the normalization term grows linearly with $t$, effectively scaling the memory update contribution by a factor of $1/t$ as the sequence length increases.

\subsection{Future Direction: GatedFWA Beyond \texorpdfstring{$\mathsf{TC}^0$}{TC0} Circuit Complexity}
\label{app:circuit_complexity}

In this section, we analyze the theoretical expressivity of GatedFWA through the lens of \textit{\textbf{Circuit Complexity}}. We demonstrate that despite the introduction of a data-dependent decay mechanism, GatedFWA belongs to the complexity class $\mathsf{TC}^0$, sharing the same fundamental limitations as standard Transformers and diagonal State Space Models (SSMs). We further discuss why solving $\mathsf{NC}^1$-complete problems requires architectural modifications reserved for future work.

\paragraph{Memory Recurrence and Parallelizability.}
The complexity class $\mathsf{TC}^0$ contains problems solvable by constant-depth circuits with unbounded fan-in threshold gates. Standard Transformers are known to fall within $\mathsf{TC}^0$ because their attention mechanism computes a weighted sum over the entire context, an operation that can be fully parallelized. Conversely, the class $\mathsf{NC}^1$ (logarithmic-depth circuits) includes inherently sequential problems, such as tracking a state through non-commutative updates (\emph{e.g.}, permutation composition), which standard attention cannot solve.

To determine the class of GatedFWA, we examine its memory recurrence. Let $\mathbf{M}_t \in \mathbb{R}^{d \times d}$ denote the memory state and $\mathbf{k}_t, \mathbf{v}_t \in \mathbb{R}^{1 \times d}$ denote the key and value vectors at step $t$. The GatedFWA update rule is formulated as:
\begin{equation}
    \mathbf{M}_t = (\boldsymbol{\lambda}_t   \mathbf{I}) \mathbf{M}_{t-1} + \frac{1}{w}  ( \mathbf{v}_t^\top \mathbf{k}_t - (\mathbf{c}_t   \mathbf{I}) \mathbf{v}_{t-w}^\top \mathbf{k}_{t-w}  ),
\end{equation}
where $\boldsymbol{\lambda}_t = \exp(-\boldsymbol{\alpha}_t) \in (0, 1)^d$ represents the diagonal decay gate derived from the input, and $\mathbf{c}_t$ represents the accumulated decay for the windowed term. Crucially, this recurrence is linear, and the transition coefficient $\boldsymbol{\lambda}_t$ acts as a scalar (or diagonal) contraction on the state $\mathbf{M}_{t-1}$. By recursively substituting $\mathbf{M}_{t-1}$ into the equation, we can expand the state at time $t$ as follows:
\begin{align}
    \mathbf{M}_t &= (\boldsymbol{\lambda}_t   \mathbf{I}) \mathbf{M}_{t-1} + \mathbf{U}_t \\
                 &= (\boldsymbol{\lambda}_t   \mathbf{I})  ( (\boldsymbol{\lambda}_{t-1}   \mathbf{I}) \mathbf{M}_{t-2} + \mathbf{U}_{t-1}  ) + \mathbf{U}_t \\
                 &= (\boldsymbol{\lambda}_t   \boldsymbol{\lambda}_{t-1}   \mathbf{I}) \mathbf{M}_{t-2} + (\boldsymbol{\lambda}_t   \mathbf{I}) \mathbf{U}_{t-1} + \mathbf{U}_t \\
                 &\dots \nonumber \\
                 &= \sum_{i=1}^{t}  ( \prod_{j=i+1}^{t} \boldsymbol{\lambda}_j  )   \mathbf{U}_i,
\end{align}
where $\mathbf{U}_i = \frac{1}{w} (\mathbf{v}_i^\top \mathbf{k}_i - (\mathbf{c}_i   \mathbf{I}) \mathbf{v}_{i-w}^\top \mathbf{k}_{i-w})$ represents the sliding window update term at step $i$. This derivation confirms that the state $\mathbf{M}_t$ is a direct summation of past updates, weighted by the cumulative product of decay gates, which can be computed in parallel.

\paragraph{Equivalence to SSMs and Transformers.}
The structure of the solution in Eq.~\eqref{eq:associative_mem_update} reveals that $\mathbf{M}_t$ can be computed using a parallel associative scan (prefix sum) or a direct convolution. Since generalized matrix multiplication and parallel prefix sums are computable in $\mathsf{TC}^0$, GatedFWA does not require sequential depth proportional to the input length $N$. This places GatedFWA in the same complexity class as:
\begin{itemize}
    \item \textbf{Standard Transformers:} Which compute global weighted sums via Softmax.
    \item \textbf{Diagonal SSMs (\emph{e.g.}, Mamba/S4):} Which utilize diagonal state transitions that can be parallelized via convolution or associative scans.
\end{itemize}
Unlike general Recurrent Neural Networks (RNNs) which use non-linear or non-diagonal matrix transitions ($\mathbf{h}_t = \sigma(\mathbf{W}\mathbf{h}_{t-1} + \mathbf{x}_t)$), GatedFWA lacks the capacity for arbitrary state manipulation. Specifically, the gating mechanism $\boldsymbol{\lambda}_t$ enables the model to \textit{forget} history or \textit{ignore} new inputs, but it does not allow for the selective \textit{modification} or \textit{erasure} of specific vector components based on state-content interaction. Therefore, it cannot solve $\mathsf{NC}^1$-complete problems such as dynamic graph connectivity or the $S_5$ permutation tracking problem.

\paragraph{Example: The \texorpdfstring{$S_5$}{S5} Permutation Problem.}
To illustrate the limitation concretely, consider the problem of tracking a state evolved by a sequence of permutations from the symmetric group $S_5$. Let the state at time $t$ be a permutation $\pi_t \in S_5$, updated via $\pi_t = \pi_{t-1} \circ \sigma_t$. Because permutation composition is non-commutative ($\pi \circ \sigma \neq \sigma \circ \pi$), the final state depends on the specific order of operations, a problem known to be $\mathsf{NC}^1$-complete. GatedFWA, however, employs a memory recurrence governed by data-dependent decay gates $\mathbf{\Lambda}_t = \text{diag}(\exp(-\boldsymbol{\alpha}_t))$. Crucially, diagonal matrices commute ($\mathbf{\Lambda}_i \mathbf{\Lambda}_j = \mathbf{\Lambda}_j \mathbf{\Lambda}_i$), which reduces the state evolution to a commutative parallel scan (or weighted sum). Consequently, the model falls into the complexity class $\mathsf{TC}^0$, rendering it structurally incapable of solving inherently sequential, non-commutative problems like $S_5$ without a depth that grows with sequence length.

\paragraph{Towards \texorpdfstring{$\mathsf{NC}^1$}{NC1} Expressivity.}
To elevate the expressivity of the model to $\mathsf{NC}^1$, the memory update mechanism must support non-commutative state transitions, effectively moving from a \emph{write-only} memory to a \emph{read-write} memory. A promising direction is the incorporation of the \textit{Delta Rule}, which modifies the recurrence to:
\begin{equation}
    \mathbf{M}_t = \mathbf{M}_{t-1} + (\mathbf{v}_t - \mathbf{k}_t \mathbf{M}_{t-1})^\top \mathbf{k}_t.
\end{equation}
Rearranging the terms reveals the implicit state transition: $\mathbf{M}_t \approx \mathbf{M}_{t-1}(\mathbf{I} - \mathbf{k}_t^\top \mathbf{k}_t) + \mathbf{v}_t^\top \mathbf{k}_t$. Here, the term $(\mathbf{I} - \mathbf{k}_t^\top \mathbf{k}_t)$ acts as a rank-one, non-diagonal transition matrix. Unlike the diagonal gates in GatedFWA, these matrices do not generally commute, enabling the model to represent complex, order-dependent state evolutions characteristic of $\mathsf{NC}^1$ problems. However, this expressivity comes at a computational cost: the recurrence can no longer be computed via simple commutative parallel scans, potentially requiring specialized chunk-wise algorithms or logarithmic-depth parallelization for efficient training. Integrating such mechanisms into the GatedFWA framework remains a subject for future optimization.

\section{Additional Experiment Details}
\label{app:parameter_statistics}
We provide the architecture details for conducting the scaling law experiments on OpenWebText in Tab.~\ref{tab:details_owt}. The architecture configs follow exactly from the GLA paper \cite{yang2023gated}. We also provide architecture details for pretraining-finetuning with \texttt{nanochat} framework in Tab.~\ref{tab:details_params_nanochat}. The architecture configs are tuned based on \texttt{nanochat} repository.
\begin{table*}[h]
\centering
\caption{Training details on OpenWebText.}
\resizebox{0.99\textwidth}{!}{
\begin{NiceTabular}{@{}c|cccccc|ccc@{}}
\toprule
\textbf{Params} & \textbf{n\_layers} & \textbf{d\_model} & \textbf{n\_heads / d\_head} & \textbf{LR} & \textbf{BS} & \textbf{Tokens} & \textbf{nsa\_block\_size} & \textbf{nsa\_block\_counts} & \textbf{heads\_per\_gqa\_group} \\ \midrule
120M            & 12                 & 768               & 12 / 64                     & 2e-3                        & 0.5M tokens         & 2.5B            & 64                        & 16                          & 4                               \\
360M            & 24                 & 1024              & 16 / 64                     & 2e-3                        & 0.5M tokens         & 7B              & 64                        & 16                          & 4                               \\ \bottomrule
\end{NiceTabular}
}
\label{tab:details_owt}
\end{table*}

\begin{table*}[h]
\centering
\caption{Training details with \texttt{nanochat}.}
\resizebox{0.99\textwidth}{!}{
\begin{NiceTabular}{@{}c|ccclllcccc@{}}
\toprule
\textbf{Model}        & \textbf{n\_layers} & \textbf{d\_model} & \textbf{n\_heads / d\_head} & \textbf{BS(Pre)} & \textbf{BS(Mid)} & \textbf{BS(Sft)} & \textbf{LR (emb)} & \textbf{LR(2D Matrices)} & \textbf{LR(Non-2D Matrices \textit{e.g.} bias/norm)} & \textbf{LR(gate)} \\ \midrule
Transformer           & 20                 & 1280              & 10 / 128                    & 256              & 256              & 32               & AdamW:0.2         & Muon:0.004               & AdamW:0.004                                          & -                 \\
Transformer(GatedFWA) & 20                 & 1280              & 10 / 128                    & 256              & 256              & 32               & AdamW:0.2         & Muon:0.004               & AdamW:0.004                                          & AdamW:0.004       \\ \bottomrule
\end{NiceTabular}

}
\label{tab:details_params_nanochat}
\end{table*}

\end{document}